\documentclass[review]{elsarticle}

\makeatletter
\def\ps@pprintTitle{%
  \let\@oddhead\@empty
  \let\@evenhead\@empty
  \let\@oddfoot\@empty
  \let\@evenfoot\@oddfoot
}
\makeatother
\usepackage{lineno,hyperref}
\modulolinenumbers[5]

\journal{Elsevier}
\usepackage[english]{babel}

\usepackage{graphicx}        

\usepackage{amsmath,bm,amsfonts,amssymb}

\usepackage{commath}      
\usepackage{color,xcolor,ucs}
\usepackage{subfig}
\usepackage[font=small,labelfont=bf]{caption}
\usepackage{float}
\usepackage{amsthm}

\usepackage{upgreek}
\usepackage[euler]{textgreek}
\usepackage{wrapfig}
\usepackage{algpseudocode}
\usepackage[ruled,vlined,linesnumbered]{algorithm2e}

\usepackage{moreverb,url}
\usepackage{multirow}

\usepackage{textcomp} 
\SetKwInOut{Input}{input}
\SetKwInOut{Output}{output}

\newtheorem{theorem}{Theorem}
\newtheorem{definition}{Definition}
\newtheorem{lemma}{Lemma}
\newtheorem{remark}{Remark}

\DeclareMathOperator*{\argmax}{arg\,max}
\DeclareMathOperator*{\argmin}{arg\,min}
\DeclareMathOperator{\EX}{\mathbb{E}} 

\makeatletter
\newcommand{\clearsubcaptcounter}{\setcounter{sub\@captype}{0}}
\makeatother    

\newcommand{\B}{\textbf}

\definecolor{revisioncolor}{rgb}{0,0,0}
\newcommand{\revtext}[1]{{\color{revisioncolor} #1}}

\begin{document}

\begin{frontmatter}

\title{Safe Motion Planning with Environment Uncertainty}

\author{Antony Thomas, Fulvio Mastrogiovanni, Marco Baglietto}
\address{Department of Informatics, Bioengineering, Robotics, and Systems Engineering, University of Genoa, Via All'Opera Pia 13, 16145 Genoa, Italy. \\\  \textsl{\{antony.thomas@dibris.unige.it, fulvio.mastrogiovanni@unige.it, marco.baglietto@unige.it\}}}




\begin{abstract}
We present an approach for safe motion planning under robot state and environment (obstacle and landmark location) uncertainties. To this end, we first develop an approach that accounts for the landmark uncertainties during robot localization. Existing planning approaches assume that the landmark locations are well known or are known with little uncertainty. However, this might not be true in practice. Noisy sensors and imperfect motions compound to the errors originating from the estimate of environment features. Moreover, possible occlusions and dynamic objects in the environment render imperfect landmark estimation. Consequently, not considering this uncertainty can wrongly localize the robot, leading to inefficient plans. Our approach thus incorporates the landmark uncertainty within the Bayes filter estimation framework. We also analyse the effect of considering this uncertainty and delineate the conditions under which it can be ignored. Second, we extend the state-of-the-art by computing an exact expression for the collision probability under Gaussian distributed robot motion, perception and obstacle location uncertainties. We formulate the collision probability process as a quadratic form in random variables. Under Gaussian distribution assumptions, an exact expression for collision probability is thus obtained which is computable in real-time. In contrast, existing approaches approximate the collision probability using upper-bounds that can lead to overly conservative estimate and thereby suboptimal plans.  We demonstrate and evaluate our approach using a theoretical example and simulations. We also present a comparison of our approach to different state-of-the-art methods.
\end{abstract}


\end{frontmatter}


\section{Introduction}
\label{sec:intro}
Robots have become more pervasive and are being increasingly used in close proximity to humans and other objects (both static and dynamic) in
factories, living spaces, elderly care, and robotic surgery. Planning for collision free trajectories in real-time is imperative for robots to operate safely and efficiently in such realistic conditions. However, uncertainties often arise due to insufficient knowledge about the environment, imperfect sensing or inexact robot motions. In these situations, it is indispensable to employ approaches that perform safe motion planning under motion and sensing uncertainties. Planning is therefore performed in the \textit{belief} space, which corresponds to the set of all probability distributions over possible robot states and other variables of interest. However, at the planning time, future observations are yet to be obtained. Thus, for efficient planning and decision making, it is required to reason about future belief distributions due to possible actions and the corresponding expected future observations. The corresponding problem, known as \textit{Belief Space Planning} (BSP), falls under the category of \textit{Partially Observable Markov Decision Processes} (POMDPs)~\cite{kaelbling1998AI}. 

Uncertain environments are such that they often preclude the existence of collision free trajectories~\cite{aoude2013AR}. In the presence of noisy sensors, both the robot and the environment state cannot be estimated precisely and one can only reason in terms of the corresponding belief states. Moreover, in case of dynamic obstacles, their future states have to be predicted and they are not known exactly due to the lack of perfect knowledge of their motions. As such, providing safety guarantees is difficult and for safe navigation, both the robot state uncertainty and the uncertainty in obstacle estimates need to be considered while computing collision probabilities.  

\begin{figure}[t]
	\centering
		\includegraphics[scale=0.25]{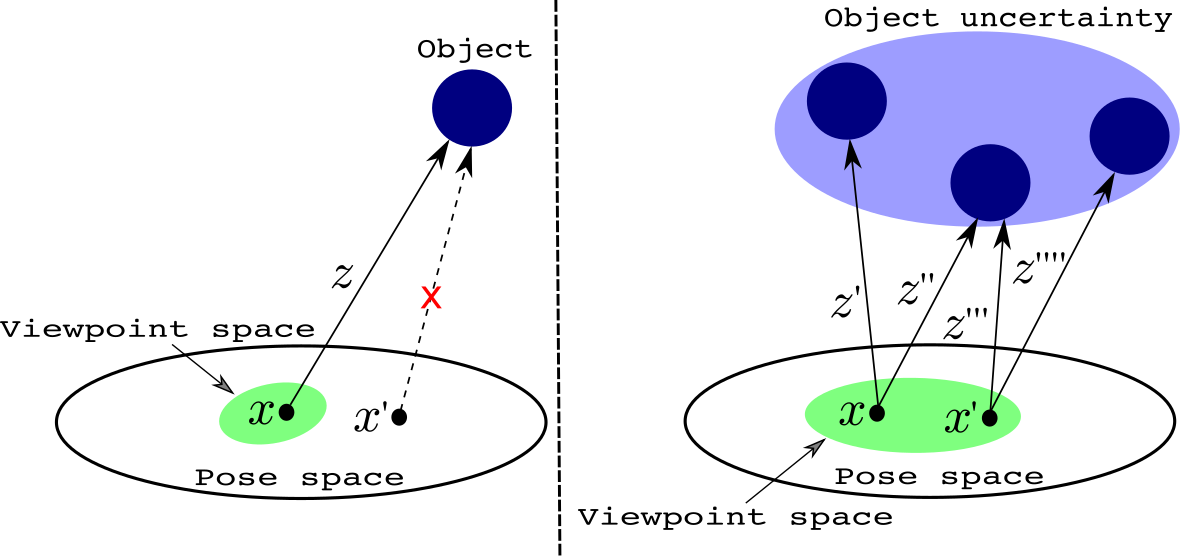}
		\caption{The blue blob denotes an object in the environment. The green region is the viewpoint space corresponding to the set of poses from which the object can be observed. Robot pose $x$ produces an observation $z$ however $x'$ does not produce an observation. On the right hand side, the light-blue shaded region denotes the uncertainty in object location. In practice the object can be anywhere within the uncertainty region. As a result, depending on where the object really is, it can be observed from either $x$ or $x'$ or both the poses.}
	\label{fig:concept}
\end{figure}

Most robotic tasks require the knowledge of where the robot is with respect to the environment. Localization is therefore one of the most fundamental problem in robotics and significantly impacts planning and decision making. As such, localization is therefore a key aspect for safe and efficient navigation. However, existing approaches assume that the landmark locations are known precisely or with little uncertainty. For example, given the map of the environment, while planning for future actions the standard Markov localization does not take into account the map uncertainty (that is, landmark locations are assumed to be perfect). This means that given the map and the sensing range, there exists a region from which the landmark can be observed. This however, might not be true in practice. For example, let us consider a \textit{Simultaneous Localization and Mapping} (SLAM) session. Wrong data association or dynamics objects preventing loop closures could lead to wrongly estimated landmark locations and thereby the corresponding map. Thus landmark estimates arising out of such a SLAM session might not be known precisely. It is noteworthy that due to this landmark location uncertainty the regions from which the landmark can be observed are also uncertain. This is visualized in Fig.~\ref{fig:concept}. We define the \textit{pose space} as the set of all possible poses the robot can assume. The blue blob denotes the object which when viewed from a pose $x$ produces an observation $z$. Different observations are produced when the object is viewed from distinct poses such that the object falls within the sensing range and there are no occlusions. The set of all such poses is a subset of the pose space and is defined to be the \textit{viewpoint space} (green region in the figure). We note that the viewpoint space is sensor-dependent and is determined by the sensing range and other aspects such as occluding objects. Intuitively, ~\revtext{the} viewpoint space is object depended and this set changes for each object. However, a pose $x'$ that falls outside the viewpoint space does not produce an observation. Consequently, as seen on the left hand side of the figure, when the object location is known precisely, there exists a subset of the pose space from which the object can be observed. On the right hand side of the figure, the light-blue shaded region denotes the uncertainty in object location. Thus in practice the object can be anywhere within the uncertainty region. As a result, given a pose, it cannot be said with certainty that the object can be observed. Subsequently, the cardinality of the subset of the pose space from which the object can be observed is increased, that is, the viewpoint space has increased (green region in the figure). As seen in the figure, depending on the object really is, it can be observed from either $x$ or $x'$ or both the poses. As a result, one can only reason in terms of the probability of observing the object from the considered pose or the viewpoint. Therefore, a probability distribution function for the viewpoint space is obtained where the mean viewpoint corresponds to observing the object with highest probability. Not accounting for this uncertainty can cause localization errors, leading to inefficient plans. In this paper, we will use the term \textit{object uncertainty} to refer to this notion of uncertainty in landmark location. 

\subsection{Related Work}
Much research activities about planning under robot state uncertainty have been carried out in the past few years, with applications spanning a variety of areas~\cite{prentice2009IJRR, van_den_berg2012IJRR, kaelbling2013IJRR, agha_mohammadi2014IJRR, Kurniawati2016ISRR, pathak2018IJRR, thomas2019ISRR,garrett2019arxiv}. Yet,  most approaches assume that the landmarks are fairly well known or are known with little uncertainty. However, in practice, the environment is seldom known with high certainty and hence providing formal guarantees for safe robotic tasks under environment uncertainty is of vital importance. \revtext{In this work we explicitly consider uncertainties in the landmark locations and  derive the resulting Bayes filter.}

\revtext{Several methods exist in the literature to compute collision probability and they differ from one another in different aspects such as (1) the formulation of the collision constraint, (2) assumptions in the shape of the robot and the obstacle, (3) modelling the associated uncertainties. The approach of truncating Gaussian distributions is leveraged to compute risk-aware and asymptotically
optimal trajectories by~\cite{liu2014ICRA}.}~\cite{patil2012ICRA} truncate~\cite{johnson1994truncatedGaussian} the estimated \textit{a priori} Gaussian state distributions to consider only the collision free samples. Thus propagating these truncated distributions enable them to compute collision free trajectories.~\revtext{However, it must hold that the propagated distributions are Gaussians and therefore the truncated distributions are approximated to be Gaussian distributions. Though we model uncertainties using Gaussian distribution, we do not truncate them.}~\revtext{Bounding volume approaches enlarge the robot by their 3-$\sigma$ uncertainties. Rectangular bounding boxes for robot and the obstacles are used in~\cite{hardy2013TRO} to compute an approximate upper bound. In}~\cite{bry2011ICRA} the future state distributions are predicted and the uncertainties are used to compute bounded collision probabilities.~\cite{lee2013IROS} use sigma hulls for~\revtext{bounding} robot links and compute the signed distance of these hulls to the obstacles to formulate the collision avoidance constraints.~\revtext{These approaches typically overestimate collision probabilities and the computed values tend to be larger than the actual values.}   

The joint distribution between the robot and the obstacle is used to derive the collision constraint in~\cite{dutoit2011IEEE} and~\cite{park2018IEEE}~\revtext{, giving bounded collision probabilities}.~\revtext{The probabilities are obtained by marginalizing the joint distribution.} However, since there is no closed form solution to this formulation, an approximate formulation is computed~\revtext{in~\cite{dutoit2011IEEE},~\cite{park2018IEEE}}. Assuming that the robot radius is negligible the joint distribution can be approximated as the product of the volume occupied by the robot and the conditional
distribution of the obstacle evaluated at the robot location.~\revtext{However, as identified by~\cite{zhu2019RAL}, both the approaches produce tight bounds only when the sizes of objects are relatively very small compared with their position uncertainties. In contrast, we do not assume any approximation and derive an exact expression for collision probability computatation.}~\cite{lambert2008ICCARV}, an approximation is computed using \textit{Monte Carlo Integration} (MCI), which is nonetheless computationally intensive. Another related work that uses a Monte Carlo approach and is real-time compatible is \textit{Monte Carlo Motion Planning} (MCMP)~\cite{janson2018ISRR}. They first solve a deterministic motion planning problem with an inflated obstacle and later adjust the inflation to compute the desired safe path.

\revtext{Chance-constrained\footnote{A chance-constrained approach finds the optimal sequence of control inputs subject to the constraint that the collision probability must be below a user-specified threshold. This constraint is known as a chance constraint.} approaches compute approximate upper bounds by linearizing the collision conditions. In contrast, we compute the exact collsion probability value.~\cite{blackmore2011TRO} employs chance-constraints to ensure that the probability of collision is below a specified threshold.} This approach is leveraged to compute bounded collision-free trajectories with dynamic obstacles by~\cite{zhu2019RAL}, wherein the dynamic obstacles follow a constant velocity model with Gaussian noise. In~\cite{aoude2013AR}, a \textit{Gaussian Process} (GP) based approach is used to learn motion patterns (a mapping from states to trajectory derivatives) to identify possible future obstacles trajectories.~\cite{axelrod2018IJRR} focus exclusively on obstacle uncertainty. They formalize a notion of \textit{shadows}, which are the geometric equivalent of confidence intervals for uncertain obstacles. The shadows fundamentally give rise to loose bounds but the computational complexity of bounding the collision probability is greatly reduced. Uncertain obstacles are modelled as polytopes with Gaussian-distributed faces by~\cite{shimanuki2018WAFR}. Planning a collision-free path in the presence of \textit{risk zones} is considered by~\cite{salzman2017ICAPS} by penalizing the time spent in these zones. Risk contours map, which take into account the risk
information (uncertainties in location, size and geometry of obstacles) in uncertain environments are used by~\cite{jasour2019RSS} to obtain safe paths with bounded risks. A related approach for randomly moving obstacles is presented by~\cite{hakobyan2019RAL}. Formal verification methods have also been used to construct safe plans~\cite{ding2013ICRA, sadigh2016RSS}. 

Most of the approaches discussed above leverage Boole's inequality to compute the collision probability along a path by summing or multiplying the probabilities along different waypoints in the path. However, the additive approach assumes that the probabilities along the waypoints are mutually exclusive and the multiplicative approach treats them as independent. Such approaches tend to be overly conservative and rather than computing bounded collision probabilities along a path, the bound should be checked for each configuration along the path itself.~\revtext{In this work we perform this check for each configuration along the path.} Moreover, in most approaches, the collision probability computed along each waypoint is an approximation of the true value. For example, the MCI approach of~\cite{lambert2008ICCARV} approximates the resulting double summation expression for collision probability to a single summation.~\cite{zhu2019RAL} compute an approximate upper bound for collision probability by linearizing the collision condition.~\cite{park2018IEEE} and~\cite{dutoit2011IEEE} assume the volume occupied by the robot to be negligible. On the one hand, such approximations can overly penalize paths and could gauge all plans to be infeasible. On the other hand some approximations can be lower\footnote{For example, the approach of~\cite{dutoit2011IEEE} computes a lower value than the actual when the robot state covariance is small.} than the true collision probability values and can lead to synthesizing unsafe plans.

\subsection{Contributions}
This paper makes three main theoretical contributions. First, we incorporate object uncertainties in BSP and derive the resulting Bayes filter in terms of the prediction and measurement updates of the \textit{Extended Kalman Filter} (EKF). We also analyse the effect of incorporating object uncertainty while computing the posterior robot belief state. 

The second is the computation of the collision probability under robot state uncertainty and the uncertainties in estimated obstacle locations. We formulate the collision avoidance constraint as a quadratic form in random variables under the assumption of spherical geometries for robot and obstacles. Unlike previous approaches that compute an upper bound or derive conservative estimates for the probability of collision, we derive an exact expression for computing it. The convergence of the obtained expression is proved and an upper bound for the truncation error is also derived. We also formalize a notion of safety in order to compute configurations that satisfy the required collision probability bounds. Moreover, we employ a Bayesian framework to predict future states of dynamic obstacles when their motion model is unknown. The current state of dynamic obstacles is estimated given the measurements, and the estimated states are used to predict future trajectories. The approach is not limited to a single obstacle and can be used to estimate the states of all obstacles detected by the robot.

The third is the derivation of collision constraints for convex shaped polygonal objects. Note that we derive the collision avoidance constraint by assuming robot and obstacles to be spherical objects. This is a reasonable assumption for most practical purposes since the robot and obstacles may be enclosed by minimum volume spheres. However, in the case of 2D mobile robot collision avoidance planning, most often it is enough to consider the robot footprint. Due to the minimum volume spheres considered, the footprints are thus circular. Yet, this assumption can lead to overly conservative estimates and while deriving the collision constraint and therefore we go a step beyond previous approaches by considering the exact convex footprints of the robot and the obstacles. 

This paper is an extension of the preliminary work presented by~\cite{thomas2021ISR}. Compared with~\cite{thomas2021ISR}, a more rigorous treatment of object uncertainty is presented by introducing viewpoint and pose spaces. We also provide the derivations for the mean and covariance of the posterior belief when incorporating object uncertainty and further discuss the impact of object uncertainty under varying object location uncertainties. We further derive the collision constraint for non-circular geometries and provide a rigorous comparison of our approach with other state-of-the-art methods. Collision computation is extended to dynamic obstacles by estimating the state of all the obstacles perceived by the robot. Further, the offline planning approach is extended to real-time online planning. Finally, we discuss the limitations of our approach and delineate suitable extensions to overcome them. We also present new results with static and dynamic obstacles in both single-robot and multi-robot settings.   

\subsection{Organization}
The rest of the paper is organized as follows. We introduce the notations, define the considered problem and the assumptions in Section~\ref{sec:notations}. In Section~\ref{sec:obj_uncertianty}, we derive the posterior belief parameters incorporating object uncertainty. We formulate the collision constraint and derive an expression for the collision probability in Section~\ref{sec:coll_prob}. In this section, we also provide a rigorous comparison of our approach to other approaches. State estimation for dynamic obstacles is then discussed in Section~\ref{sec:obs}. In Section~\ref{experiments} we evaluate our approach in simulation under different mobile robot scenarios. In Section~\ref{sec:discussion} we outline the limitations and discuss possible extensions. Conclusion are drawn in Section~\ref{sec:conclusion}. 

\section{Notations and Problem Definition}
\label{sec:notations}
We shall denote vectors by bold lower case letters, that is $\B{x}$ and its components by lower case letters. The transpose of $\textbf{x}$ will be denoted by $\textbf{x}^T$ and its Euclidean norm by $\norm{\textbf{x}} = \sqrt{\textbf{x}^T\textbf{x}}$. The the expected value of a random vector\footnote{By a random vector we refer to a vector random variable.} $\textbf{x}$ will be denoted by $\EX(\B{x})$. Matrices will be denoted by capital letters, that is $M$. The trace of a square matrix $M$ will be denoted by $tr(M)$ and its determinant by $det(M)$. The identity matrix will be denoted by $I$ or $I_n$ when the dimension needs to be stressed. A diagonal matrix with diagonal elements $\lambda_1, \ldots, \lambda_n$ will be denoted by $diag(\lambda_1, \ldots, \lambda_n)$. Sets will be denoted using mathcal fonts, that is $\mathcal{S}$. Unless otherwise mentioned, subscripts on vectors/matrices will be used to denote time indexes and (whenever necessary) superscripts will be used to indicate the robot or the object that it refers to. For example, $\textbf{x}_k^i$ represents the state of robot $i$ at time $k$. The notation $P(\cdot)$ will be used to denote the probability of an event and the probability density function (pdf) will be denoted by $p(\cdot)$.

We now formally define the problem that we tackle in this paper. Let us consider a mobile robot operating in a partially-observable environment. The map of the environment is either known \textit{a priori} or it is built using a standard \textit{Simultaneous Localization and Mapping} (SLAM) algorithm. At any time $k$, we denote the robot pose (or configuration) by $\B{x}_k\doteq(x_k, y_k,\theta_k)$,  the acquired measurement from objects is denoted by $\B{z}_k$ and the applied control action is denoted as $\B{u}_k$. It is noteworthy that by \textit{objects} we refer to both the landmarks and the obstacles in the environment. We also make the following assumptions: (1) the uncertainties are modelled using Gaussian distributions, (2) the robot and obstacles are assumed to be non-deformable objects.  

To describe the dynamics of the robot, we consider a standard motion model with Gaussian noise
\begin{equation}
\B{x}_{k+1} = f(\B{x}_k,\B{u}_k) + w_{k}\  ,  \ w_{k} \sim \mathcal{N}(0,R_{k})
\label{eq:odometry_model}
\end{equation}
\noindent where $w_k$ is the random unobservable noise, modeled as a zero mean Gaussian. Objects are detected through the robot's sensors and assuming known data association, the observation model can be written as  
\begin{equation}
\B{z}_k = h(\B{x}_k,O_k^i) + v_k \  ,  \ v_k \sim \mathcal{N}(0,Q_k)
\label{eq:measurement_model}
\end{equation}
\noindent where $O_k^i$ is the $i-$th detected object and $v_k$ is the zero mean Gaussian noise. We note here that the motion (\ref{eq:odometry_model}) and observation (\ref{eq:measurement_model}) models can be written probabilistically as 
$p(\B{x}_{k+1}|\B{x}_k, \B{u}_k)$ and $p(\B{z}_k|\B{x}_k,O_k^i)$, respectively.

Given the models in (\ref{eq:odometry_model}) and (\ref{eq:measurement_model}), in this paper we compute \textit{safe} plans, wherein the probability of collision of the robot with any obstacle is guaranteed to be less than a specified bound while navigating to the goal.  To this end, we consider the object
uncertainties while localizing the robot. Further, we employ a Bayesian framework to estimate the current state of dynamic obstacles and use the estimated states to predict future obstacle states. The estimated states are then used for non-myopic collision avoidance planning. Given the robot and obstacle locations, we compute the exact probability of collision under motion, sensing uncertainty, and the uncertainty in obstacle location.

\section{\revtext{Object Uncertainty}}
\label{sec:obj_uncertianty}
As discussed in Section~\ref{sec:intro}, most localization approaches assume that the landmark locations are known precisely or with little uncertainty. This however, might not be true in practice due to noisy measurements and/or imperfect sensors. Thus, it is pertinent that landmark uncertainties are considered within the localization and planning framework. Below, we delineate the incorporation of object uncertainty within the Bayes filter.
\subsection{\revtext{Object Uncertainty Model for BSP}}
We define the collection of all objects in the environment to be the \textit{object space} $\mathcal{O} = \{O^i| \text{$O^i$ is an object, and} \ 1 \leq i \leq |\mathcal{O}|\}$. Let us posit that at time $k$ the robot received a measurement $\B{z}_k$ which was originated by observing the object $O^i_k$. Given an initial distribution $p(\B{x}_0)$, and the motion and observation models in (\ref{eq:odometry_model}) and (\ref{eq:measurement_model}), the posterior probability distribution at time $k$ is the \textit{belief} $b[\B{x}_k]$ and can be written as
\begin{equation}
  b[\B{x}_k] = p(\B{x}_k|\B{z}_k, O_k^i, \B{z}_{0:k-1},\B{u}_{0:k-1})
\end{equation}  
\noindent where $O^i_k$ is the object observed at time $k$, $\B{z}_{0:k-1} \doteq \{\B{z}_0,...,\B{z}_{k-1}\}$ and $\B{u}_{0:k-1} \doteq \{\B{u}_0,...,\B{u}_{k-1}\}$. This posterior is thus a multivariate Gaussian, that is, $b[\B{x}_k] \sim \mathcal{N}\left(\bm{\mu}_k,\Sigma_k \right)$, where $\bm{\mu}_k$, $\Sigma_k$ are the mean and covariance of $\B{x}_k$, respectively. We note that this posterior belief is computed after incorporating the measurement at time $k$, that is, $\B{z}_k$.

Given the belief $b[\B{x}_k]$ and an action $\B{u}_k$, the belief before incorporating a measurement will be called the propagated belief and can be written as
\begin{equation}
b[\bar{\B{x}_{k+1}}] = \int_{\B{x}_k} p(\B{x}_{k+1}|\B{x}_{k},\B{u}_{k})b[\B{x}_{k}] 
\end{equation}
Now let us consider that a measurement $\B{z}_{k+1}$ is obtained that corresponds to observing the object $O_{k+1}^i$. The posterior distribution $b[\B{x}_{k+1}]$ can then be computed using Bayes rule and the theorem of total probability. This expansion is obtained in terms of the belief at the previous time step since the Bayes filter is recursive. Thus we have
\begin{multline}
p(\B{x}_{k+1}|\B{z}_{k+1}, O_{k+1}^i, \B{z}_{0:k},\B{u}_{0:k}) =\\ \eta p(\B{z}_{k+1}|\B{x}_{k+1}, O_{k+1}^i) p(O_{k+1}^i|\B{x}_{k+1}) \int_{\B{x}_{k}} p(\B{x}_{k+1}|\B{x}_{k},\B{u}_{k})b[\B{x}_{k}] 
\label{eq:posterior}
\end{multline}
\noindent where $\eta = 1/p(\B{z}_{k+1}|\B{z}_{0:k},\B{u}_{0:k})$ is the normalization constant. The term $p(O_{k+1}^i|\B{x}_{k+1})$ denotes the probability of observing the object $O_{k+1}^i$ from the pose $\B{x}_{k+1}$. In other words, this term models the fact, how likely it is to observe $O_{k+1}^i$ from $\B{x}_{k+1}$ and thus models the object uncertainty. The term $p(O_{k+1}^i|\B{x}_{k+1})$ also additionally model aspects such as occlusions due to static obstacles that hinder the observation, occlusions that results due to dynamic obstacles, faulty sensors or other aspects that impedes observations of objects of interest. Thus, given an object one can only reason probabilistically about observing it to obtain the corresponding measurement. However, when the object uncertainty and other additional aspects are ignored an object is observed whenever the robot is within the viewpoint space (see left hand side of Fig.~\ref{fig:concept}). Thus, in the case of such an assumption, for poses within the viewpoint space the term is equal to unity, that is, $p(O_{k+1}^i|\B{x}_{k+1}) = 1$. For poses that lie outside the viewpoint space\revtext{,} $p(O_{k+1}^i|\B{x}_{k+1}) = 0$ and hence no measurement can be obtained. As such, when the object uncertainty is ignored, the term $p(O_{k+1}^i|\B{x}_{k+1})$ can be removed from~(\ref{eq:posterior}) and the posterior belief parameters can be computed using the standard EKF update equation as
\begin{equation}
\begin{split}
K_{k+1}     & = \bar{\Sigma}_{k+1} H_{k+1}^T\left(H_{k+1} \bar{\Sigma}_{k+1}  H_{k+1}^T + Q_{k+1}\right)^{-1}\\
\bm{\mu}_{k+1} & = \bar{\bm{\mu}}_{k+1} + K_{k+1}\left(\B{z}_{k+1}-h(\bar{\bm{\mu}}_{k+1})\right)\\
\Sigma_{k+1} & = \left(I -K_{k+1} H_{k+1}\right)\bar{\Sigma}_{k+1} 
\end{split}
\label{eq:update}
\end{equation}
\noindent where $H_{k+1}$ is the Jacobian of $h(\cdot)$ with respect to $\B{x}_{k+1}$, and $K_{k+1}$ is the Kalman gain.

The exposition so far has been agnostic to the actual model of $p(O_{k+1}^i|\B{x}_{k+1})$. In general, this term can be modeled given the environment map, the sensing capabilities and the robot objectives. These aspects should hence be incorporated to obtain the actual object uncertainty model. However, in this work we approximate the object distribution as a Gaussian distribution:
\begin{equation}
p(O_{k+1}^i|\B{x}_{k+1}) \sim \mathcal{N}(\bm{\mu}_{O_{k+1}^i},\Sigma_{O_{k+1}^i}) 
\end{equation}
\noindent where $\bm{\mu}_{O_{k+1}^i}$ is the viewpoint/pose that corresponds to the maximum probability of observing $O_{k+1}^i$ and $\Sigma_{O_{k+1}^i}$ is the associated uncertainty in the observation.  

We will now consider the object uncertainty term $p(O_{k+1}^i|\B{x}_{k+1})$ and derive the Gaussian belief parameters by expanding~(\ref{eq:posterior}). Expanding the right hand side of (\ref{eq:posterior}) using the probability density
function (pdf) of multivariate Gaussian distributions, we have $b[\B{x}_{k+1}] = \eta' \int \text{exp}(-\mathcal{J}_{k+1})$, where $\eta'$ contains the non-exponential terms and $\mathcal{J}_{k+1}$ is given by
\begin{multline}
\mathcal{J}_{k+1} = \frac{1}{2}\left(\B{z}_{k+1} - h\left(\bar{\bm{\mu}}_{k+1}\right) - H_{k+1}\left(\B{x}_{k+1} - \bar{\bm{\mu}}_{k+1}\right)\right)^T
 \\ 
Q_{k+1}^{-1}\left(\B{z}_{k+1} - h\left(\bar{\bm{\mu}}_{k+1}\right) - H_{k+1}\left(\B{x}_{k+1} - \bar{\bm{\mu}}_{k+1}\right)\right) \\
 +  \frac{1}{2}(\bm{x}_{k+1} - \bm{\mu}_{O_{k+1}^i})^T\Sigma_{O_{k+1}^i}^{-1}(\B{x}_{k+1} -\bm{\mu}_{O_{k+1}^i})\\
 +  \frac{1}{2} (\B{x}_{k+1} - \bar{\bm{\mu}}_{k+1} )^T \bar{\Sigma}_{k+1}^{-1} (\B{x}_{k+1} - \bar{\bm{\mu}}_{k+1} )
 \label{eq:exponential_terms}
\end{multline}
\noindent where $H_{k+1}$ is the Jacobian of $h(\cdot)$ with respect to $\B{x}_{k+1}$. As shown in~\cite{thrun2005book}, the covariance $\Sigma_{k+1}$ is obtained as the inverse of the second derivative of $\mathcal{J}_{k+1}$ with respect to $\B{x}_{k+1}$. The expression for the second derivative is obtained as
\begin{equation}
\frac{\partial^2 \mathcal{J}_{k+1}}{\partial \B{x}_{k+1}^2} = H_{k+1}^TQ_{k+1}^{-1}H_{k+1} + \Sigma_{O_{k+1}^i}^{-1} +\bar{\Sigma}_{k+1}^{-1}
\label{eq:second_derivative}
\end{equation}
\noindent Therefore the posterior covariance is obtained as
\begin{equation}
\Sigma_{k+1}^{-1} =  H_{k+1}^TQ_{k+1}^{-1}H_{k+1} + \Sigma_{O_{k+1}^i}^{-1} +\bar{\Sigma}_{k+1}^{-1}
\label{eq:cov}
\end{equation}
\noindent The mean of $b[\B{x}_{k+1}]$ is the value that maximizes $b[\B{x}_{k+1}]$ and hence is obtained by equating the first derivative of $\mathcal{J}_{k+1}$ to zero. The expression for the mean $\bm{\mu}_{k+1}$ is obtained as (see Appendix A for derivation)
\begin{equation}
 \bm{\mu}_{k+1} =  \bar{\bm{\mu}}_{k+1} + K_{k+1}\left(\B{z}_{k+1} - h\left(\bar{\bm{\mu}}_{k+1}\right)\right) + 
 \Sigma_{k+1}\Sigma_{O_{k+1}^i}^{-1}\left(\bm{\mu}_{O_{k+1}^i}- \bar{\bm{\mu}}_{k+1}\right)
\label{eq:object_mu}
\end{equation}

\noindent where $K_{k+1}=\Sigma_{k+1} H_{k+1}^TQ_{k+1}^{-1}$ is the Kalman gain. We note that when no object uncertainty is considered the update step of the standard EKF gives $\bm{\mu}_{k+1} =  \bar{\bm{\mu}}_{k+1} + K_{k+1}\left(\B{z}_{k+1} - h\left(\bar{\bm{\mu}}_{k+1}\right)\right)$. The additional term in~(\ref{eq:object_mu}) rightly adjusts the mean $\bm{\mu}_{k+1}$ accounting for the fact that the object location is uncertain. 

As in the standard EKF based Bayes filter, the expression for the covariance $\Sigma_{k+l}$ can also be derived in terms of the Kalman gain $K_{k+1}$ and the predicted covariance $\bar{\Sigma}_{k+1}$. Using the matrix inversion lemma on~(\ref{eq:cov}), the following expression is obtained (see Appendix B for derivation)
\begin{equation}
\Sigma_{k+1} = \left(I - K_{k+1}H_{k+1}\right)\bar{\Sigma}_{k+1}\tilde{\Sigma}_{k+l}\Sigma_{O_{k+1}^i}
\label{eq:object_sigma}
\end{equation}
\noindent where $\tilde{\Sigma}_{k+l} = \left(\bar{\Sigma}_{k+1} +\Sigma_{O_{k+1}^i} \right)^{-1}$.

When object uncertainty is not considered, the update step of the standard EKF gives $\Sigma_{k+1} = \left(I - K_{k+1}H_{k+1}\right)\bar{\Sigma}_{k+1}$. The extra terms in~(\ref{eq:object_sigma}) account for the object uncertainty and scale the posterior covariance accordingly. We note that when object uncertainty is not considered, $p(O_{k+1}^i|\B{x}_{k+1}) = 1$ and hence the results in~(\ref{eq:object_mu}) and~(\ref{eq:object_sigma}) reduce to that of the standard EKF case in~(\ref{eq:update}). The method presented above is easily generalized to multiple objects observed at any time instant. This is done by following the sequential-sensor method~(\cite{durrant2016multisensor}), considering the fact that given the current state estimate, the observations are independent of each other.
\subsection{\revtext{Implications}}
Let us now analyse the effect of object uncertainty. As discussed above when object uncertainty is not assumed, $p(O_{k+1}^i|\B{x}_{k+1}) = 1$, and therefore the posterior belief parameters reduce to that of the standard EKF case. However, in practice, one should consider object uncertainty and the posterior belief parameters are as delineated in~(\ref{eq:object_mu}) and~(\ref{eq:object_sigma}). Yet, the impact of considering object uncertainty in localisation depends on the covariance of the estimated object location. When the covariance of the object location is much larger compared to the predicted robot belief state covariance, the impact of considering object uncertainty is greatly reduced.
\begin{lemma}
When the covariance of the estimated object location is much larger than the predicted robot belief state covariance, that is, when $\Sigma_{O_{k+1}^i} \gg \bar{\Sigma}_{k+1}$, then the object uncertainty has limited impact and can be ignored. 
\label{lemma_object}
\end{lemma} 
\textit{Proof.} In order to prove the above lemma, it suffices to show that when $\Sigma_{O_{k+1}^i} \gg \bar{\Sigma}_{k+1}$, the posterior belief parameters reduce to that of the standard EKF update case as given in~(\ref{eq:update}). Let us first consider the expession in~(\ref{eq:cov}). Using the fact that $\Sigma_{O_{k+1}^i} \gg \bar{\Sigma}_{k+1}$, then $\Sigma_{O_{k+1}^i}^{-1} \ll \bar{\Sigma}_{k+1}^{-1}$ and hence it can be neglected when compared to $\bar{\Sigma}_{k+1}^{-1}$. This gives $\Sigma_{k+1} = \left( H_{k+1}^TQ_{k+1}^{-1}H_{k+1} +\bar{\Sigma}_{k+1}^{-1}\right)^{-1}$, and is the expression for the posterior belief covariance when object uncertainty is not considered. Again, using $\Sigma_{O_{k+1}^i} \gg \bar{\Sigma}_{k+1}$, $\bar{\Sigma}_{k+1}$  can be neglected from the sum $\left(\bar{\Sigma}_{k+1} +\Sigma_{O_{k+1}^i} \right)$. The expression for Kalman gain thus reduces to
\begin{multline}
K_{k+1} = \bar{\Sigma}_{k+1}\left(\Sigma_{O_{k+1}^i} \right)^{-1}\Sigma_{O_{k+1}^i} H_{k+1}^T \\ \left(H_{k+1}\bar{\Sigma}_{k+1}\left(\Sigma_{O_{k+1}^i} \right)^{-1}\Sigma_{O_{k+1}^i} H_{k+1}^T + Q_{k+1}\right)^{-1}\\
= \bar{\Sigma}_{k+1}H_{k+1}^T \left(H_{k+1}\bar{\Sigma}_{k+1} H_{k+1}^T + Q_{k+1}\right)^{-1}
\end{multline}
\noindent Thus, as can be seen in~(\ref{eq:update}), the Kalman gain is exactly the gain obtained when object uncertainty is not considered. Similarly, we have $ \Sigma_{k+1}\Sigma_{O_{k+1}^i}^{-1} \ll 1$, thus $\bm{\mu}_{k+1} =  \bar{\bm{\mu}}_{k+1} + K_{k+1}\left(\B{z}_{k+1} - h\left(\bar{\bm{\mu}}_{k+1}\right)\right)$. Following a similar argument, it is easily seen that $\Sigma_{k+1} = \left(I -K_{k+1} H_{k+1}\right)\bar{\Sigma}_{k+1}$. This completes the proof of Lemma~\ref{lemma_object}.

Although the above result might seem counter-intuitive at first, we note here that the viewpoint space, when object uncertainty is not considered, is the space centred around the mean of the viewpoint space when the object uncertainty is considered. When the covariance of the object location is very high, then the probability values for viewpoints slightly away from the mean reduces drastically. Consequently considering these viewpoints adds little impact.

\section{Exact Collision Probability}
\label{sec:coll_prob}
We denote by $\mathcal{R}$ the set of all points occupied by a rigid-body robot at any given time. Similarly, let $\mathcal{S}$ represent the set of all points occupied by a rigid-body obstacle. A collision occurs if there exits a point such that it is in both $\mathcal{R}$ and $\mathcal{S}$. Thus the collision condition is defined as
\begin{equation}
\mathcal{R} \cap \mathcal{S} \neq \{\phi\}
\end{equation}
\noindent and we denote the probability of collision as $P\left(\mathcal{R} \cap \mathcal{S} \neq \{\phi\}\right)$. In this work we assume spherical geometries for $\mathcal{R}$ and $\mathcal{S}$ with radii $r_1$ and $s_1$, respectively. We assign body-fixed reference frames to both the robot and the obstacle located at $\B{x}_k$ and $\B{s}_k$, respectively in the global frame. By abuse of notation we will use $\B{x}_k$ and $\B{s}_k$ equivalently to $\mathcal{R}$ and $\mathcal{S}$. The collision condition is thus defined in terms of the body-fixed frames as
\begin{equation}
\mathcal{C}_{\B{x}_k,\B{s}_k}: \mathcal{R} \cap \mathcal{S} \neq \{\phi\}
\end{equation}
We recall here that the locations of the obstacles are in general uncertain. Let us now consider an obstacle at any given time instant, distributed according to the Gaussian $\B{s}_k \sim \mathcal{N}\left(\bm{\mu}_{\B{s}_k},\Sigma_{\B{s}_k}\right)$, where $\bm{\mu}_{\B{s}_k}$ represents the mean and $\Sigma_{\B{s}_k}$ the associated covariance. Given the current robot state $\B{x}_k$ and the obstacle state $\B{s}_k$, the probability of collision can be formulated if the joint distribution between the robot and the obstacle state is known. In such a case the collision probability is given by
\begin{equation}
P\left(\mathcal{C}_{\B{x}_k,\B{s}_k}\right) = \int_{\B{x}_k} \int_{\B{s}_k} I_c(\B{x}_k,\B{s}_k)p(\B{x}_k,\B{s}_k)
\label{eq:collision_prob}
\end{equation}
where $\mathcal{C}_{\B{x}_k,\B{s}_k}$ as defined above represents the fact that the robot configuration $\B{x}_k$ and its collision with an obstacle at location $\B{s}_k$ is considered, and $I_c$ is an indicator function defined as
\begin{equation}
   I_c(\B{x}_k,\B{s}_k)= 
   \begin{cases}
     1 \ &\text{if} \ \mathcal{R} \cap \mathcal{S} \neq \{\phi\} \\
     0 \ &\text{otherwise}.
   \end{cases}
\end{equation}
\noindent and $p(\B{x}_k,\B{s}_k)$ is the joint distribution of the robot and the obstacle.~\cite{zhu2019RAL} compute an approximate upper bound for the collision probability by linearizing the collision condition.~\cite{lambert2008ICCARV} use MCI to compute (\ref{eq:collision_prob}). However, the resulting double summation is approximated to a single summation to reduce computational complexity.~\cite{dutoit2011IEEE},~\cite{park2018IEEE} approximate the integral in (\ref{eq:collision_prob}) as $Vp(\B{x}_k,\B{s}_k)$, where $V$ is the volume occupied by the robot. For computing $p(\B{x}_k,\B{s}_k)$, they first assume a distribution centered around the obstacle with the covariance being the sum of the robot and obstacle location uncertainties. Then the density $p(\B{x}_k,\B{s}_k)$ is computed by assuming a constant robot location. Du Toit and Burdick use the robot center, whereas in~\cite{park2018IEEE} the maximum of $p(\B{x}_k,\B{s}_k)$ on the surface of the robot is used to obtain an upper bound. However, the approximation is valid only when the robot radius is negligible. To demonstrate, let us re-write the collision condition as
\begin{equation}
P\left(\mathcal{C}_{\B{x}_k,\B{s}_k}\right) = \int_{\B{x}_k} \left[\int_{\B{s}_k \in \mathcal{R} } p(\B{s}_k|\B{x}_k)\right] p(\B{x}_k)
\end{equation}
\noindent If the robot radius is negligible then it can be assumed that $\B{s}_k=\B{x}_k$, giving
\begin{equation}
P\left(\mathcal{C}_{\B{x}_k,\B{s}_k}\right) = \int_{\B{x}_k} \left[p(\B{s}_k = \B{x}_k| \B{x}_k) \int_{\B{s}_k \in \mathcal{R} } \right] p(\B{x}_k)
\end{equation}
\noindent Thus assuming a constant value of the obstacle evaluated at the robot location, we have 
\begin{equation}
V = \int_{\B{s}_k \in \mathcal{R} }
\end{equation}
\noindent where $V$ is the volume occupied by the robot. The approximate collision probability is thus
\begin{equation}
P\left(\mathcal{C}_{\B{x}_k,\B{s}_k}\right) \approx V \int_{\B{x}_k} p(\B{s}_k = \B{x}_k| \B{x}_k) p(\B{x}_k)
\end{equation}

Assuming that the robot and the obstacle locations are independent Gaussian distributions with $\B{s}_k \sim \mathcal{N}\left(\bm{\mu}_{\B{s}_k},\Sigma_{\B{s}_k}\right)$ and $\B{x}_k \sim \mathcal{N}\left(\bm{\mu}_{\B{x}_k},\Sigma_{\B{x}_k} \right)$, the collision probability can be approximately written as
\begin{equation}
P\left(\mathcal{C}_{\B{x}_k,\B{s}_k}\right) \approx V  p(\B{x}_k = \bm{\mu}_{\B{x}_k}, \B{s}_k = \bm{\mu}_{\B{s}_k})
\end{equation}
\noindent where
\begin{multline}
p(\B{x}_k = \bm{\mu}_{\B{x}_k}, \B{s}_k = \bm{\mu}_{\B{s}_k}) =  det\left(2\pi \left(\Sigma_{\B{s}_k} + \Sigma_{\B{x}_k}\right) \right)^{-\frac{1}{2}} \\ \textrm{exp}\left(-\frac{1}{2}(\bm{\mu}_{\B{x}_k} - \bm{\mu}_{\B{s}_k})^T \Sigma^{-1} (\bm{\mu}_{\B{x}_k} - \bm{\mu}_{\B{s}_k}) \right)
\end{multline}
Other existing approaches truncate the state distributions or compute approximate upper bounds using chance-constraints. As such, these approaches compute an approximation of the collision probability. In contrast, we formulate the collision constraint as a quadratic form in random variables, allowing us to compute an exact expression for the collision probability. In the remainder of this~\revtext{section} a rigorous treatment of the same is presented.

Since the robot and the obstacles are assumed to be spherical objects, the collision constraint is written as
\begin{equation}
\norm{\B{x}_k -\B{s}_k}^2 \leq (r_1+s_1)^2
\label{eq:coll_condition}
\end{equation} 
\noindent where (as before) $\B{x}_k$ and $\B{s}_k$ are the random vectors that denote the robot and obstacle locations, respectively. Let the current estimates of the two random vectors be distributed according to $\B{s}_k \sim \mathcal{N}\left(\bm{\mu}_{\B{s}_k},\Sigma_{\B{s}_k}\right)$ and $\B{x}_k \sim \mathcal{N}\left(\bm{\mu}_{\B{x}_k},\Sigma_{\B{x}_k} \right)$. Let us denote the difference between the two random variables by $\B{\textit{w}} = \B{x}_k -\B{s}_k$. Using the expression for the difference between two Gaussian distributions, we have $\B{\textit{w}}  \sim \mathcal{N} \left(\bm{\mu}_{\B{x}_k} - \bm{\mu}_{\B{s}_k}, \Sigma_{\B{x}_k} + \Sigma_{\B{s}_k} \right)$. The collision constraint in~(\ref{eq:coll_condition}) can now be written in terms of $\B{\textit{w}}$,
\begin{equation}
\B{\textit{y}} = \norm{\B{\textit{w}}}^2 = \B{\textit{w}}^T\B{\textit{w}} \leq (r_1+s_1)^2
\label{eq:collision}
\end{equation}
\noindent where $\B{\textit{y}}$ is a random vector distributed according to the squared $L_2$-norm of $\B{\textit{w}}$. Now, given the probability density function (pdf) of $\B{\textit{y}}$, the collision constraint reduces to solving the integral
\begin{equation}
P\left(\mathcal{C}_{\B{x}_k,\B{s}_k}\right) = \int_{0}^{(r_1+s_1)^2} p(y)
\end{equation}
where $p(y) = P_{\B{\textit{y}}}(\B{\textit{y}} = y)$ is the pdf of $\B{\textit{y}}$. It is noteworthy that the above integral is the cumulative distribution function (cdf) of $\B{\textit{y}}$, that is, $P\left(\mathcal{C}_{\B{x}_k,\B{s}_k}\right) = F_{\B{\textit{y}}}(y)$, where $F_{\B{\textit{y}}}(y)$ denotes the cdf. Thus the collision condition reduces to finding the cdf of $\B{\textit{y}}$ such that $\B{\textit{y}} \leq (r_1+s_1)^2$. As a consequence, we have 
\begin{equation}
 P\left(\mathcal{C}_{\B{x}_k,\B{s}_k}\right) =  P\left(\B{\textit{y}} \leq (r_1+s_1)^2\right) = F_{\B{\textit{y}}}\left((r_1+s_1)^2\right) 
 \label{eq:cdf}
\end{equation}
In the following Sections, we will first show that the collision constraint is a quadratic form in random variables and later derive an exact expression for the cdf of the quadratic from.
\subsection{Quadratic Form in Random Variables}
\label{sec:param_quadratic}
We define a quadratic form in random variables:
\begin{definition}
Let $\B{\textit{x}} = \left(x_1, \ldots,x_n \right)^T$ denote a random vector with mean $\bm{\mu} = \left(\mu_1,\ldots,\mu_n\right)^T$ and covariance matrix $\Sigma$. Then the quadratic form in the random variables $x_1, \ldots,x_n $ associated with an $n \times n$ symmetric matrix $A = (a_{ij})$, with $i$ and $j$ in $1,\ldots,n$, is
\begin{equation}
Q(\B{\textit{x}}) = Q(x_1, \ldots,x_n) = \B{\textit{x}}^TA\B{\textit{x}} = \sum_{i=1}^{n}\sum_{j=1}^n a_{ij}x_ix_j
\end{equation}
\end{definition}
Let us define $\B{\textit{v}} = \Sigma^{-\frac{1}{2}}\B{\textit{x}}$ and define a random vector $\B{\textit{z}} = \left(\B{\textit{v}} - \Sigma^{-\frac{1}{2}}\bm{\mu}\right)$. The resulting distribution of $\B{\textit{z}}$ is thus zero mean with covariance being the identity matrix. Therefore, the quadratic form becomes
\begin{equation}
Q(\B{\textit{x}})  = \left(\B{\textit{z}} + \Sigma^{-\frac{1}{2}}\bm{\mu}\right)^T\Sigma^{\frac{1}{2}}A\Sigma^{\frac{1}{2}}\left(\B{\textit{z}} + \Sigma^{-\frac{1}{2}}\bm{\mu}\right)
\end{equation}
Let us suppose there exists an orthogonal matrix $P$, that is, $PP^T = I$ which diagonalizes $\Sigma^{\frac{1}{2}}A\Sigma^{\frac{1}{2}}$, then $P^T\Sigma^{\frac{1}{2}}A\Sigma^{\frac{1}{2}}P = \textrm{diag}\left(\lambda_1,\ldots,\lambda_n\right)$, where $\lambda_1,\ldots,\lambda_n$ are the eigenvalues of $\Sigma^{\frac{1}{2}}A\Sigma^{\frac{1}{2}}$. The quadratic form can now be written as
\begin{equation}
\begin{split}
Q(\B{\textit{x}}) & = \left(\B{\textit{z}} + \Sigma^{-\frac{1}{2}}\bm{\mu}\right)^T\Sigma^{\frac{1}{2}}A\Sigma^{\frac{1}{2}}\left(\B{\textit{z}} + \Sigma^{-\frac{1}{2}}\bm{\mu}\right)\\
& = \left(\B{\textit{u}} + \B{\textit{b}}\right)^T\textrm{diag}\left(\lambda_1,\ldots,\lambda_n\right)\left(\B{\textit{u}} + \B{\textit{b}}\right)
\end{split}
\label{eq:quad_form}
\end{equation}
\noindent where $\B{\textit{u}} = P^T \B{\textit{z}} = (u_1,\ldots,u_n)^T$ and $\B{\textit{b}} = P^T \Sigma^{-\frac{1}{2}}\bm{\mu} = (b_1,\ldots,b_n)^T$. The expression in (\ref{eq:quad_form}) can be written concisely,
\begin{equation}
Q(\B{\textit{x}}) = \B{\textit{x}}^TA\B{\textit{x}} = \sum_{i=1}^n \lambda_i (u_i + b_i)^2
\end{equation}

It is easily verified that the left hand side of (\ref{eq:collision}), that is $\B{\textit{w}}^T\B{\textit{w}}$, is in the quadratic form $Q(\B{\textit{w}})$ with $A = I$, that is, the identity matrix. Thus the collision probability can be computed from the cdf of the quadratic form.
\subsection{Series Expansion for the Quadratic Form}
We describe below the most general method used to obtain a series expansion for the pdf and cdf of the quadratic form in random variables. Various other methods exists in the literature and we refer the interest readers to~\cite{provost1992book} for a brief survey. The series expansion that we seek for the pdf of the quadratic form is of the form
\begin{equation}
p_{\B{y}}(y) = p(\B{y} = y) = \sum\limits_{k=0}^{\infty} c_kh_k(y)
\end{equation} 
\noindent where $c_k$ is a sequence of complex number and $\{h_k\}$ is a known sequence of the form $y^k$. Let the Laplace transform of $h_k(y)$ be denoted by $L(h_k(y))$. In the expansion sought here, the Laplace transform is of the special form~(\cite{kotz1967AMS})
\begin{equation}
L(h_k(y)) = \xi(s)\eta^k(s)
\end{equation} 
\noindent where, for $Re(s) > \alpha$ and $\alpha$ being a real constant, $\xi(s)$ is a non-vanishing (non-zero everywhere) analytic function and $\eta(s)$ is an analytic function with an inverse function $\eta(\zeta(\theta)) = \theta$. Now we are interested in the case where the series expansion is convergent, that is, $\sum_{k=0}^{\infty} c_kh_k(y) < \infty$. For any real number $\beta$, let us define
\begin{equation}
\sum_{k=0}^{\infty} c_kh_k(y) \leq \sum_{k=0}^{\infty} |c_k||h_k(y)| \leq \alpha e^{\beta y}, \ y \in [0, \infty]
\end{equation}
\noindent If the above equation is satisfied almost everywhere, then computing the Laplace transform, we have 
\begin{equation}
\int_0^{\infty} e^{-sy} \alpha e^{\beta y} = \alpha \int_0^{\infty} e^{-(s-\beta)y}dy < \infty 
\end{equation}
\noindent if $(s-a) > 0$. Therefore, from Lebesgue's dominated convergence theorem, we have the following lemma.
\begin{lemma}
Let $\{h_k\}_0^{\infty}$ be a sequence of measurable complex valued functions on $[0, \infty]$ and $\{c_k\}_0^{\infty}$ be a sequence of complex numbers such that almost everywhere the following is satisfied
\begin{equation}
\sum_{k=0}^{\infty} |c_k||h_k(y)| \leq \alpha e^{\beta y}, \ y \in [0, \infty]  
\label{eq:convergence1}
\end{equation}
\noindent where $\alpha$, $\beta$ are real constants. Then when $s > 0$ and $p_{\B{y}}(y) =  \sum\limits_{k=0}^{\infty} c_kh_k(y)$, we have
\begin{equation}
L\left(p_{\B{y}}(y) \right) = \sum\limits_{k=0}^{\infty} c_kL(h_k(y))
\end{equation}
\label{lemma11}
\end{lemma}
The implications of the above lemma are twofold. The first is that the series expansion is convergent. This however is rather straightforward from our construction of the series expansion. The second is the fact that the Laplace transform of the series $p_{\B{y}}(y)$ can be obtained by taking the Laplace transform of the individual terms of the series. This fact will be used below to derive the pdf and the cdf of the quadratic from. 
We now state the following theorem without proof. The proof may be found in~\cite{provost1992book}.
\begin{theorem}
For $Q(\B{x}) = \B{y} = \B{x}^TA\B{x}$ with $A = A^T > 0, \B{x} \sim \mathcal{N}(\bm{\mu},\Sigma), 
\Sigma > 0$, the moment generating function $M_Q(t)$ of $Q$ is given by
\begin{equation}
M_Q(t) = 
\normalfont \text{exp}\left(-\frac{1}{2} \sum\limits_{i=1}^n b_i^2\right)\text{exp}\left(\frac{1}{2} \sum\limits_{i=1}^n b_i^2(1-2t\lambda_i)^{-1}\right) \prod_{i=1}^n(1-2t\lambda_i)^{-\frac{1}{2}}
\label{eq:laplace}
\end{equation}
\end{theorem}
\noindent where the $b_i$, $\lambda_i$ are the parameters of the quadratic form as defined in Section~\ref{sec:param_quadratic}. Let us now define the series $M(\theta)$ such that
\begin{equation}
M(\theta) = \sum_{k=0}^{\infty} c_k \frac{L(h_k(y))}{\xi(\zeta(\theta))} =  \sum_{k=0}^{\infty} c_k \theta^k
\label{eq:M_theta_def}
\end{equation}
\noindent where the infinite series is a uniformly convergent series for $\theta$ in some region with $M(\theta) > 0$, $M(0) = c_0$ and $s = \zeta(\theta)$.  We note here that if $p_{\B{y}}(y) = 0$ for $y < 0$, then $M_Q(-t)$ represents the Laplace transform of $p_{\B{y}}(y)$. Thus, from~(\ref{eq:laplace}) we have
\begin{equation}
L\left(p_{\B{y}}(y) \right) = 
\text{exp}\left(-\frac{1}{2} \sum\limits_{i=1}^n b_i^2\right)\text{exp}\left(\frac{1}{2} \sum\limits_{i=1}^n b_i^2(1+2s\lambda_i)^{-1}\right) \prod_{i=1}^n(1+2s\lambda_i)^{-\frac{1}{2}}
\end{equation}
\noindent Using $\zeta(\theta) = \theta^{-1}$, we have
\begin{equation}
L\left(p_{\B{y}}(y) \right) = s^{-\frac{n}{2}} M(\theta)
\label{eq:l_m_theta}
\end{equation}
Thus we obtain,
\begin{equation}
c_0 = M(0) = \text{exp}\left(-\frac{1}{2} \sum\limits_{i=1}^n b_i^2\right) \prod_{i=1}^n(2\lambda_i)^{-\frac{1}{2}}
\label{eq:c_nought}
\end{equation}
Differentiating the natural logarithm of $M(\theta)$, we get the following form
\begin{equation}
\ln M(\theta) = d_0 + \sum_{k=1}^{\infty} d_k \frac{\theta^k}{k}
\label{eq:ln_M}
\end{equation}
\noindent where
\begin{equation}
\begin{split}
& d_0 = -\frac{1}{2}\sum\limits_{i=1}^n b_i^2 + \ln \prod_{i=1}^n(2\lambda_i)^{-\frac{1}{2}}\\
& d_k = \frac{1}{2}\sum\limits_{i=1}^n \left(1-kb_i^2\right)\left(2\lambda_i\right)^{-k}
\end{split}
\label{eq:d_noughtNk}
\end{equation}
From~(\ref{eq:l_m_theta}), we have the following lemma.
\begin{lemma}
\begin{equation}
L\left(p_{\B{y}}(y) \right) = \sum\limits_{k=0}^{\infty} c_k(-1)^{k}s^{-\left(\frac{n}{2} +k \right)}
\end{equation}
\label{lemma:laplace}
\end{lemma}
We now obtain the required expressions for the pdf and cdf of the quadratic form of $Q(\B{x})$.
\begin{lemma}
The cdf of $Q(\B{x}) = \B{y} = \B{x}^TA\B{x}$ with $A = A^T > 0, \B{x} \sim \mathcal{N}(\bm{\mu},\Sigma), 
\Sigma > 0$ is 
\begin{equation}
F_{\B{y}}(y) = p(\B{y}\leq y) = \sum_{k=0}^{\infty}(-1)^k c_k \frac{y^{\frac{n}{2} + k}}{ \Gamma\left(\frac{n}{2} +k +1\right)}
\label{eq:collision_probability}
\end{equation}

and its pdf is given by 
\begin{equation}
p_{\B{y}}(y) = p(\B{y} = y) = \sum_{k=0}^{\infty}(-1)^k c_k \frac{y^{\frac{n}{2} + k -1}}{ \Gamma\left(\frac{n}{2} +k \right)}
\label{eq:pdf}
\end{equation}
\label{theorem1}
\end{lemma}
\noindent
where $\Gamma$ denotes the gamma function, $c_0$ and $d_0$, $d_k$ are the terms defined in~(\ref{eq:c_nought}) and~(\ref{eq:d_noughtNk}), respectively. The expression for $c_k$ is given by (see Appendix C for derivation)
\begin{equation}
c_k = \frac{1}{k}\sum\limits_{j=0}^{k-1}d_{k-j}c_j, \ k \geq 1 
\label{eq:c_k}
\end{equation}
\textit{Proof.} From Lemma~\ref{lemma:laplace}, we have $L\left(p_{\B{y}}(y) \right) = \sum\limits_{k=0}^{\infty} c_k(-1)^{k}s^{-\left(\frac{n}{2} +k \right)}$. The lemma is proved by noting that $s^{-\left(\frac{n}{2} +k \right)}$ is Laplace transform of $y^{\frac{n}{2} + k -1}/ \Gamma\left(\frac{n}{2} +k \right)$. Integrating the expression for $p_{\B{y}}(y)$, we obtain the required expression for $F_{\B{y}}(y)$.
\begin{theorem}
The collision probability for the collision constraint formulated in~(\ref{eq:collision}) is given by
\begin{equation}
 P\left(\mathcal{C}_{\B{x}_k,\B{s}_k}\right) = \sum_{k=0}^{\infty}(-1)^k c_k \frac{y^{\frac{n}{2} + k -1}}{ \Gamma\left(\frac{n}{2} +k \right)}
\end{equation}
\noindent where $y = (r_1+s_1)^2$.
\end{theorem}
\textit{Proof.} From~(\ref{eq:collision}), the collision constraint is in the quadratic form $Q(\B{\textit{y}})$, with $\B{\textit{w}}  \sim \mathcal{N} \left(\bm{\mu}_{\B{x}_k} - \bm{\mu}_{\B{s}_k}, \Sigma_{\B{x}_k} + \Sigma_{\B{s}_k} \right)$. We recall here that $\B{\textit{w}} = \B{x}_k -\B{s}_k$, where $\B{x}_k$ and $\B{s}_k$ are the random vectors that denote the robot and obstacle locations, respectively and are distributed according to $\B{s}_k \sim \mathcal{N}\left(\bm{\mu}_{\B{s}_k},\Sigma_{\B{s}_k}\right)$ and $\B{x}_k \sim \mathcal{N}\left(\bm{\mu}_{\B{x}_k},\Sigma_{\B{x}_k} \right)$. As noted before, the collision probability is the cdf of the quadratic form $Q(\B{\textit{y}})$. Thus from Lemma~\ref{theorem1}, the above theorem is proved.   
\subsection{Revisiting Convergence of the Series Expansion}
As seen in Lemma~\ref{lemma11}, the cdf and the pdf of the quadratic form is convergent. In the following, we will derive upper bounds for the truncation error of the series expansions for the pdf and the cdf of the quadratic form. 

If the infinite series pdf in (\ref{eq:pdf}) is truncated after $N$ terms, the truncation error is
\begin{equation}
e(N) = \sum_{k=N+1}^{\infty}|c_kh_k(y)| = \left|\sum_{k=N+1}^{\infty} c_k \frac{y^{\frac{n}{2} + k -1}}{ \Gamma\left(\frac{n}{2} +k \right)}\right|
\end{equation}
Using Cauchy's inequality, we get
\begin{equation}
|c_k| \leq \frac{m(\rho)}{\rho^k}, \ \ m(\rho) = \textrm{max}_{|\theta| = \rho} |M(\theta)|
\label{eq:cauchy}
\end{equation}
\noindent Thus we have
\begin{multline}
e(N) \leq \frac{m(\rho)}{\rho^k}\left|\sum_{k=N+1}^{\infty} \frac{y^{\frac{n}{2} + k -1}}{ \Gamma\left(\frac{n}{2} +k \right)}\right|\leq  \\ 
m(\rho)\left(\Gamma\left(\frac{n}{2} \right)N!\right)^{-1}\left(\frac{y}{2}\right)^{\frac{n}{2} -1}\left(\frac{y}{2\rho}\right)^{N+1}\text{exp}\left(\frac{y}{2\rho}\right)
\end{multline}
where we have used the gamma function identity, $\forall \varsigma > 0, \ \Gamma(\varsigma + 1) = \varsigma\Gamma(\varsigma)$. In a similar manner, we obtain the truncation error for the infinite series cdf in~(\ref{eq:collision_probability})
\begin{equation}
E(N) \leq  m(\rho)\left(\Gamma \left(\frac{n}{2}\right)(N+1)!\right)^{-1}\left(\frac{y}{2}\right)^{\frac{n}{2}}\left(\frac{y}{2\rho}\right)^{N+1}\text{exp}\left(\frac{y}{2\rho}\right)
\label{eq:truncation}
\end{equation}  

\noindent The expression for $m(\rho)$ is obtained from~\cite{kotz1967AMS2},
\begin{equation}
m(\rho) = \prod_{i=1}^n \lambda_i^{-\frac{1}{2}} \text{exp} \left(-\frac{1}{2} \sum_{i=1}^n \frac{b_i^2\lambda_i}{ \lambda_i + \rho}\right)\prod_{i=1}^n (1 - \frac{\rho}{\lambda_i})^{-\frac{1}{2}}
\label{eq:positive}
\end{equation}

For the expression in~(\ref{eq:positive}) to be valid, it is required that $\rho < \lambda_i$ and therefore we have $\rho < \textrm{min} \ \lambda_i$. As a result, $m(\rho)$ vanishes with $\sum_{i=1}^n b_i^2 \rightarrow \infty$. We recall here that $\B{\textit{b}} = P^T \Sigma^{-\frac{1}{2}}\bm{\mu} = (b_1,\ldots,b_n)^T$. Thus, larger the distance from the obstacles and lower the uncertainty in the robot and obstacle positions, the greater is the $b_i$ value. In such scenarios, based on our experience, convergence is often attained within the first few terms of the series. 
\begin{figure}[]
\centering
  \subfloat[Configuration A]{\includegraphics[scale=0.3]{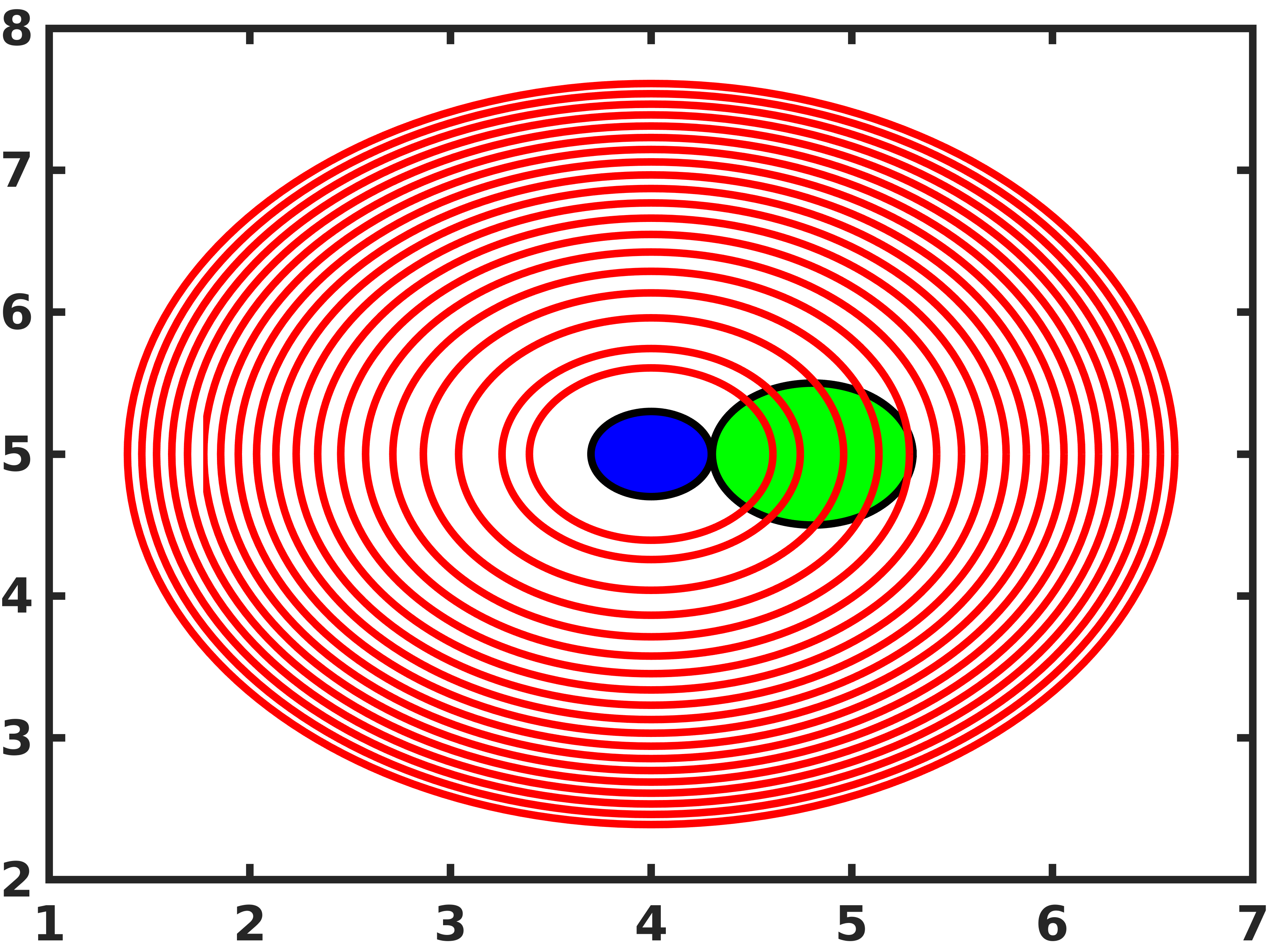}}  
   \subfloat[\revtext{Collision probability evolution}]{\includegraphics[scale=0.3]{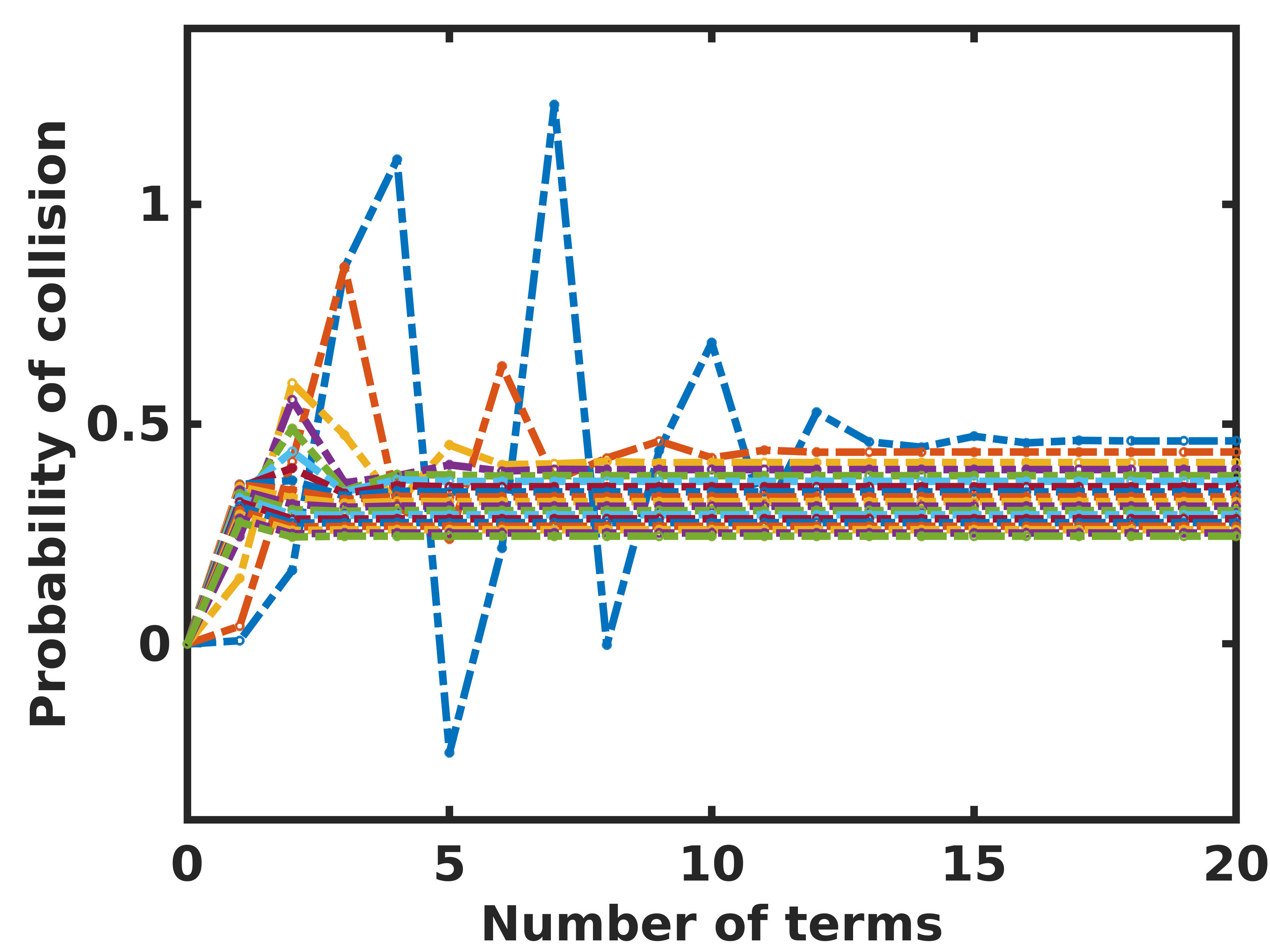}}\\
   \vspace{-0.1cm}
    \subfloat[Configuration B]{\includegraphics[scale=0.3]{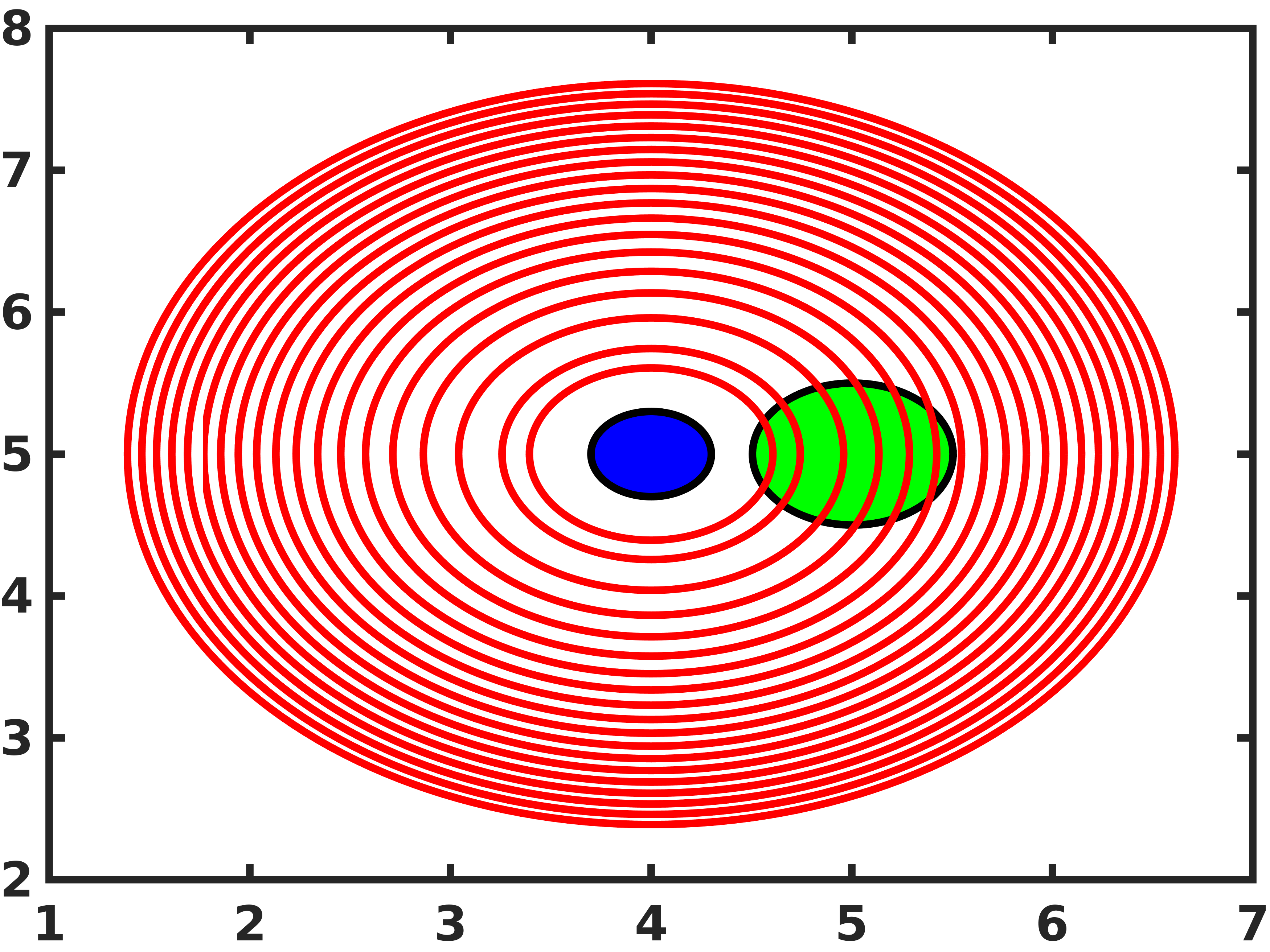}}
      \subfloat[\revtext{Collision probability evolution}]{\includegraphics[scale=0.3]{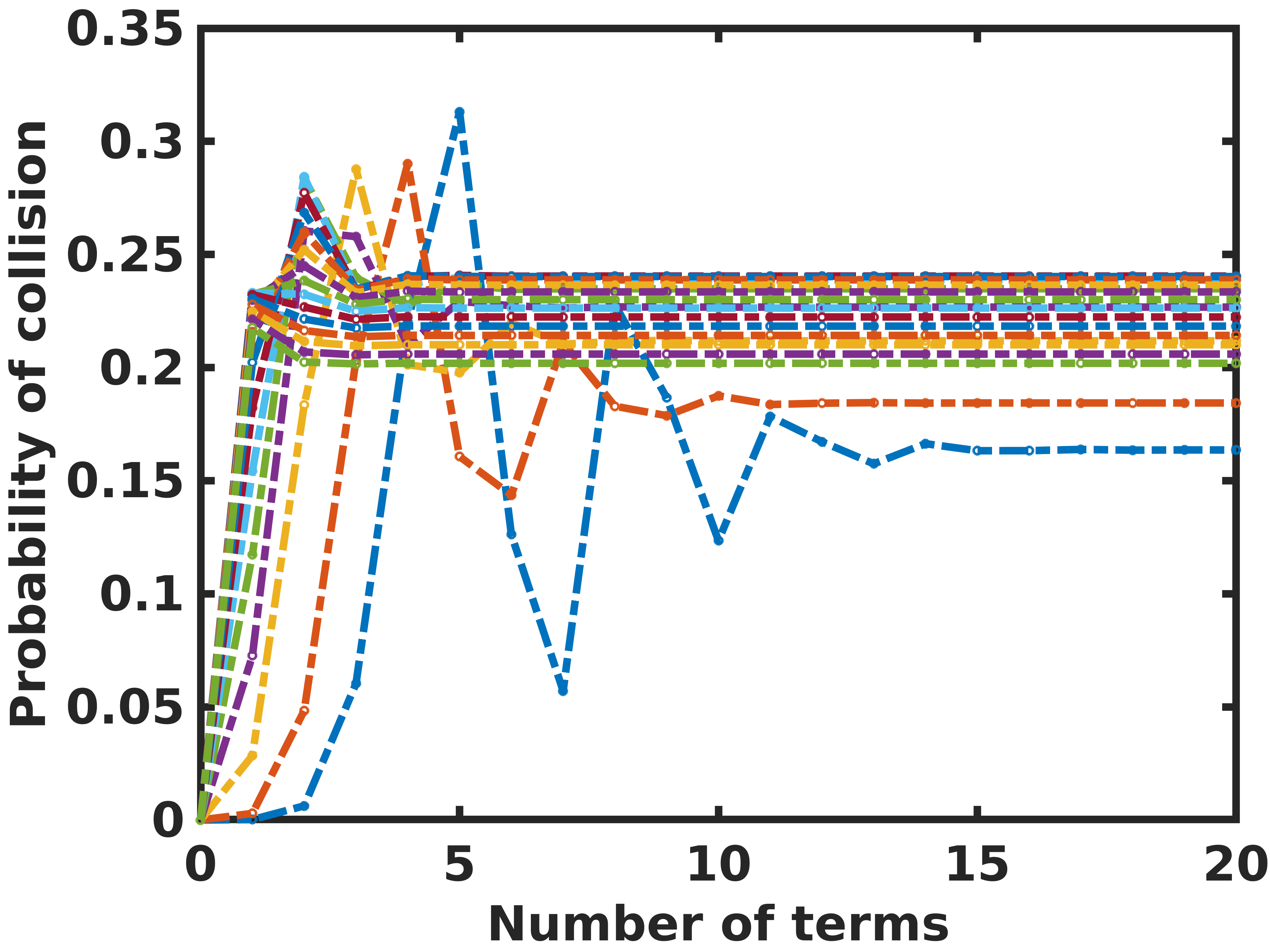}}\\
       \vspace{-0.1cm}
    \subfloat[Configuration C]{\includegraphics[scale=0.3]{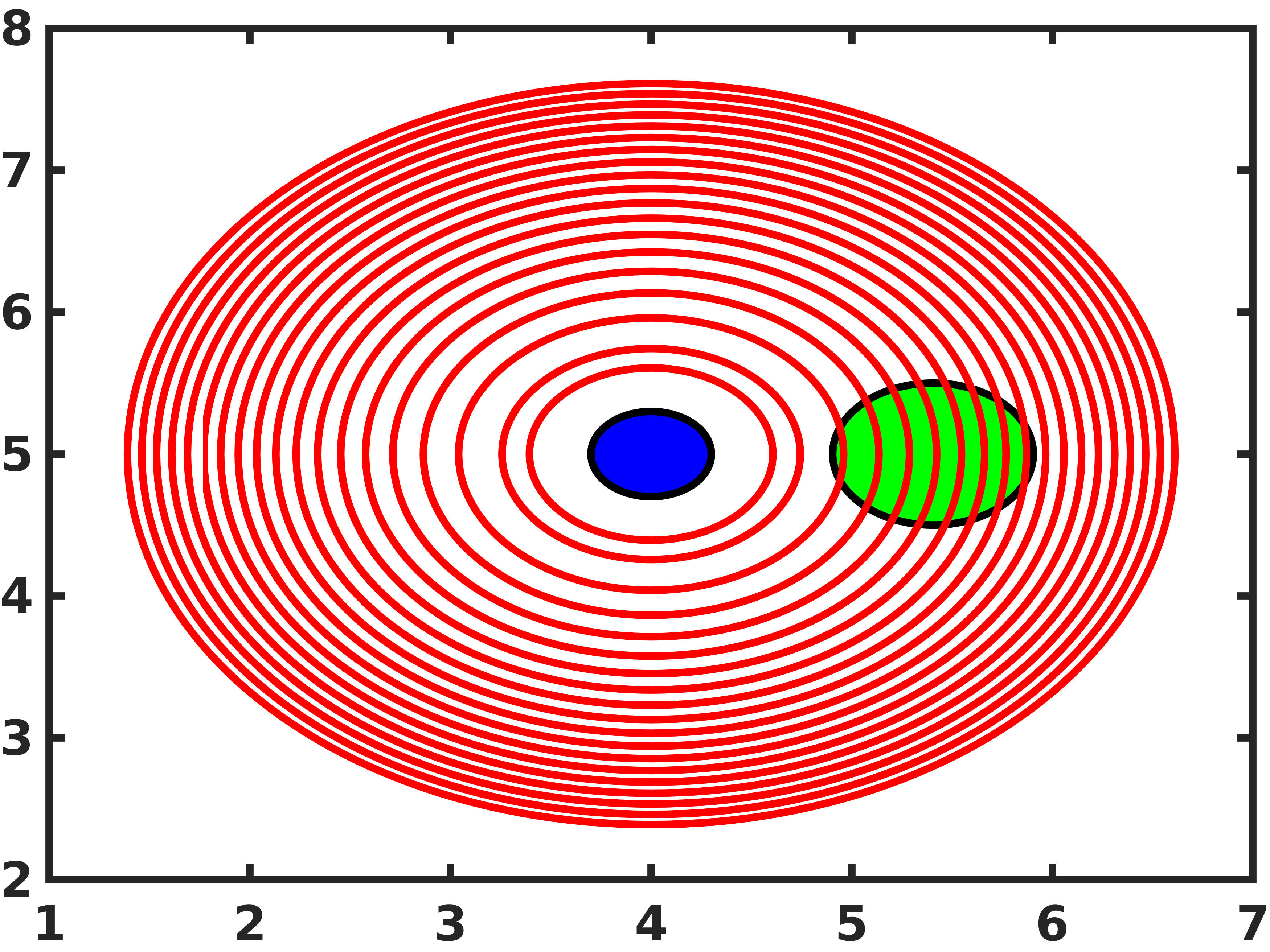}} 
    \subfloat[\revtext{Collision probability evolution}]{\includegraphics[scale=0.3]{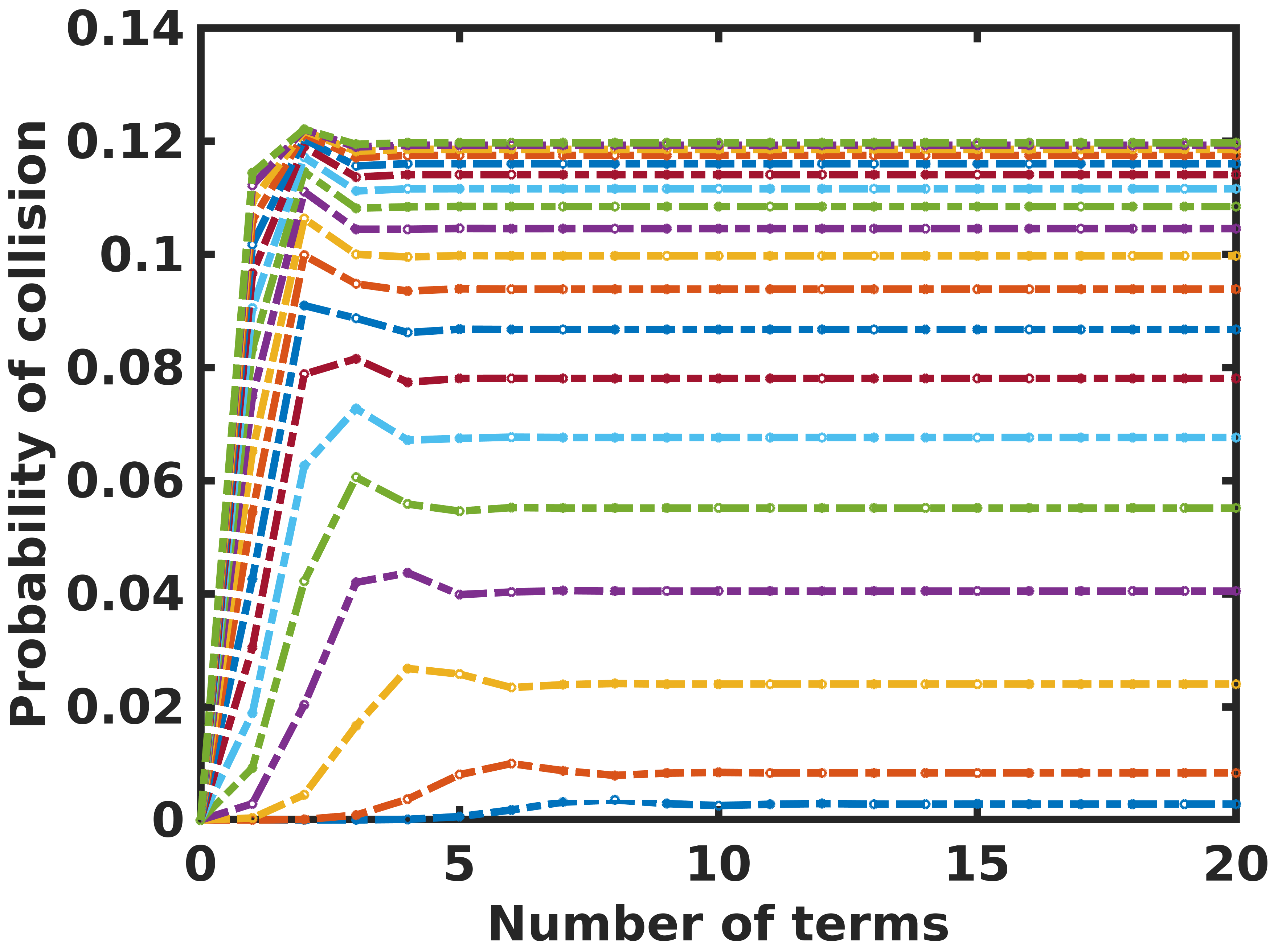}}\\
     \vspace{-0.1cm}
   \subfloat[Configuration D]{\includegraphics[scale=0.3]{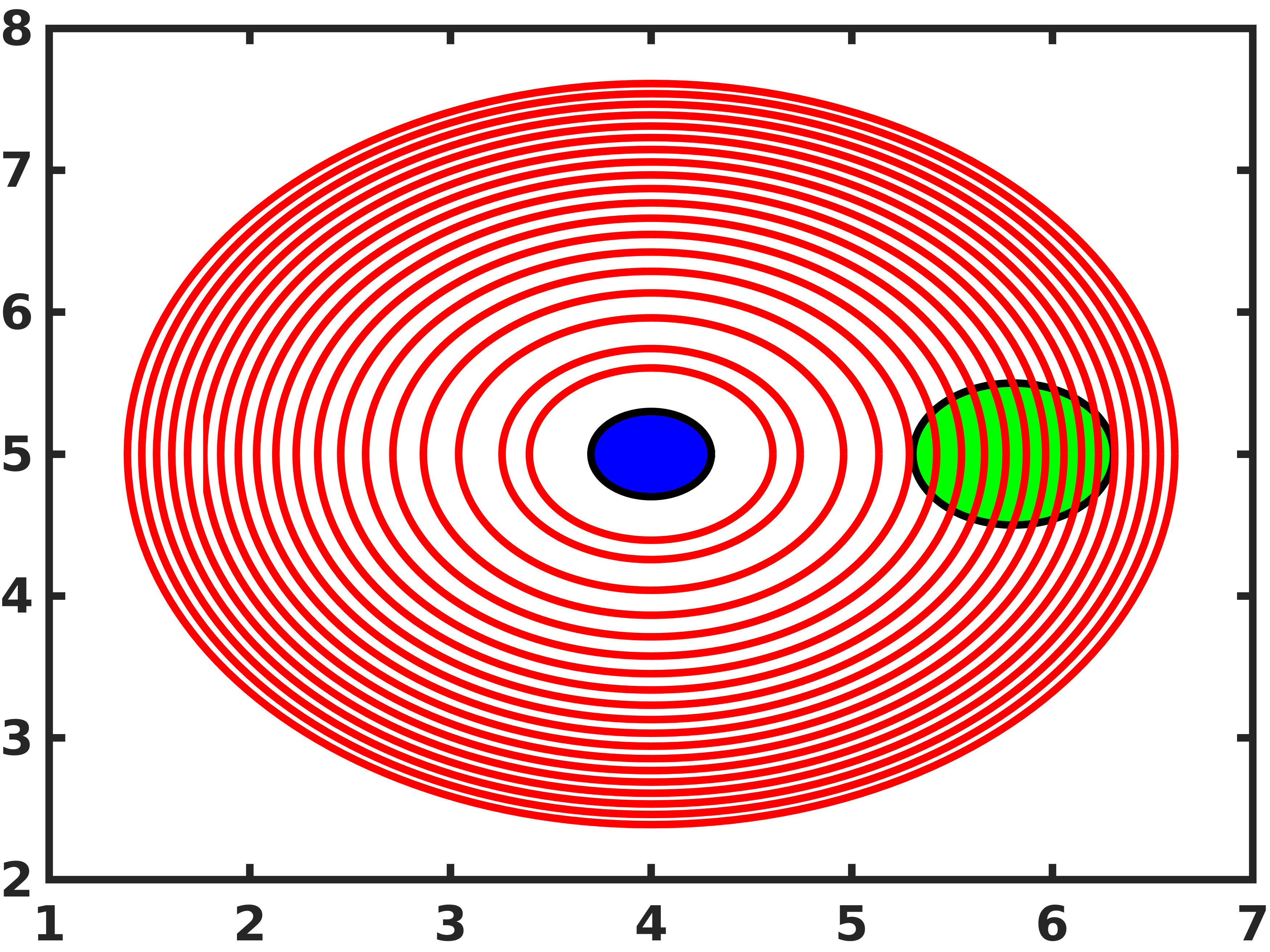}} 
    \subfloat[\revtext{Collision probability evolution}]{\includegraphics[scale=0.3]{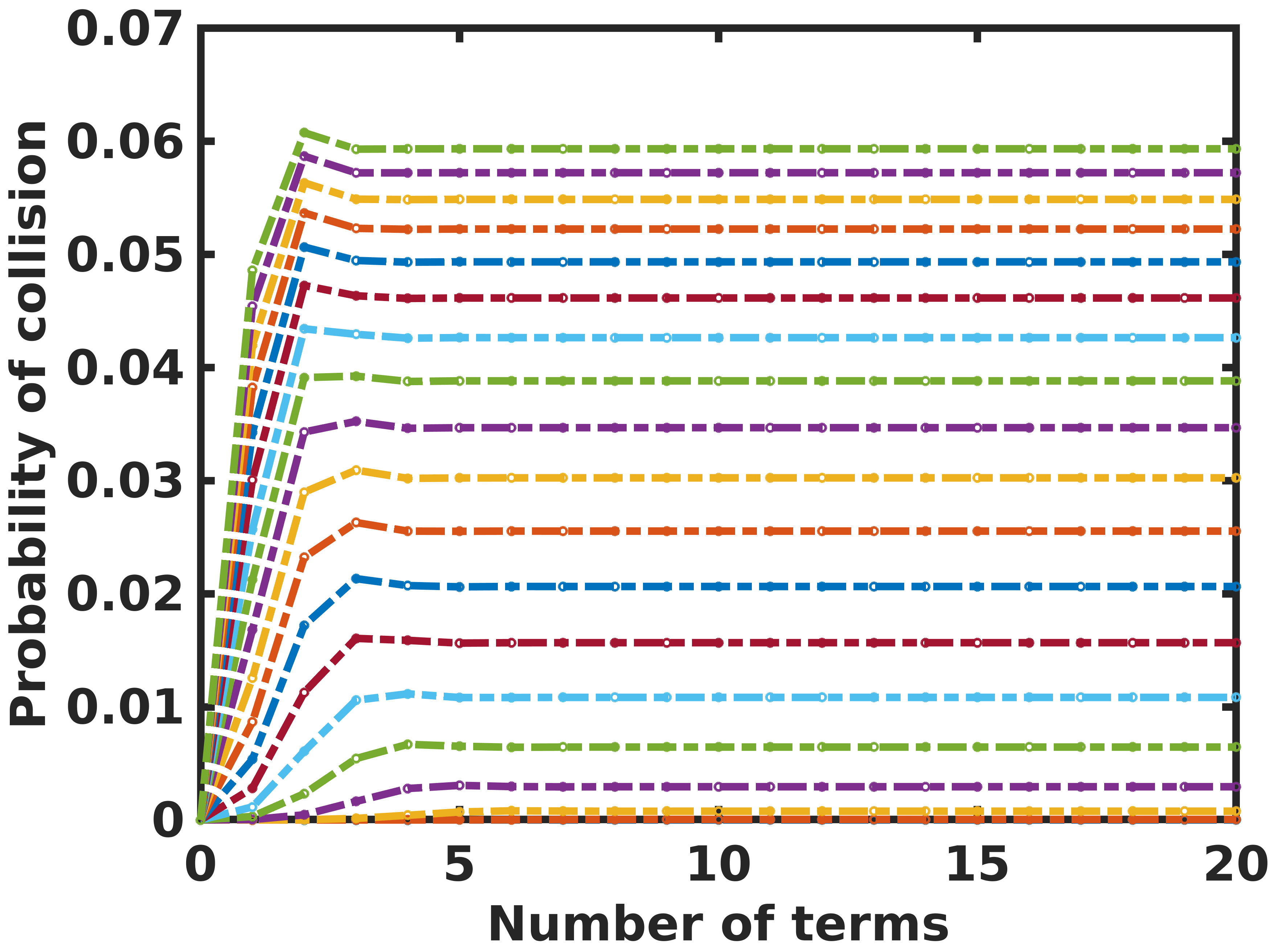}}
     \vspace{-0.1cm}
  \caption{Different configurations for a robot of radius 0.3$m$ and obstacle of radius 0.5$m$.~\revtext{Different covariances are plotted as red circles}. For each configuration the evolution of collision probability is plotted for different covariances~\revtext{in (b), (d), (f) and (h)}. In each of the 4 configurations, the maximum terms for convergence correspond to the minimum covariance of $diag(0.04,0.04)$.}
  \label{fig:convergence}
\end{figure}
\begin{figure}[]
\centering
  \subfloat[Configuration E]{\includegraphics[scale=0.3]{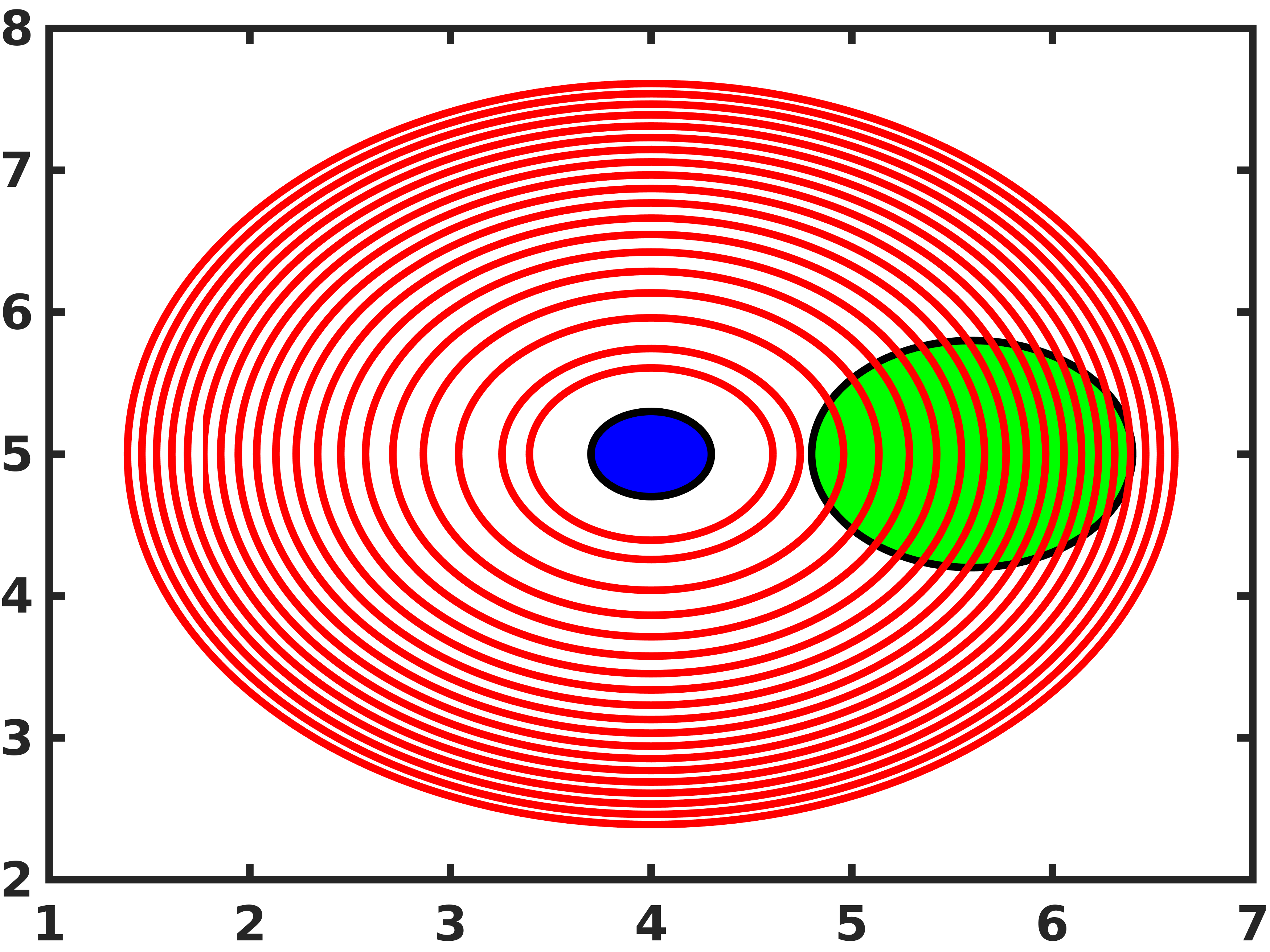}} 
   \subfloat[\revtext{Collision probability evolution}]{\includegraphics[scale=0.3]{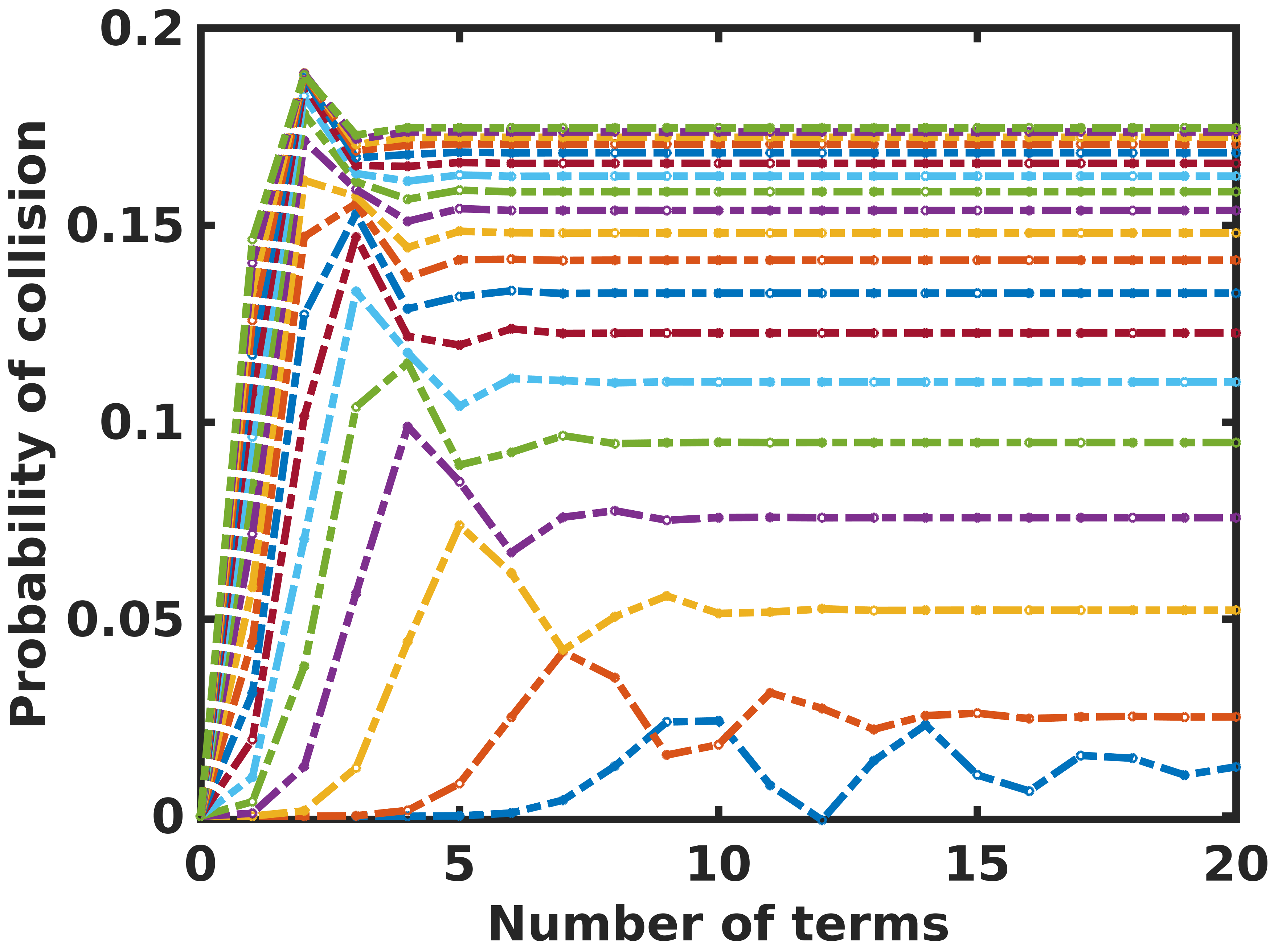}}\\
    \subfloat[Configuration F]{\includegraphics[scale=0.3]{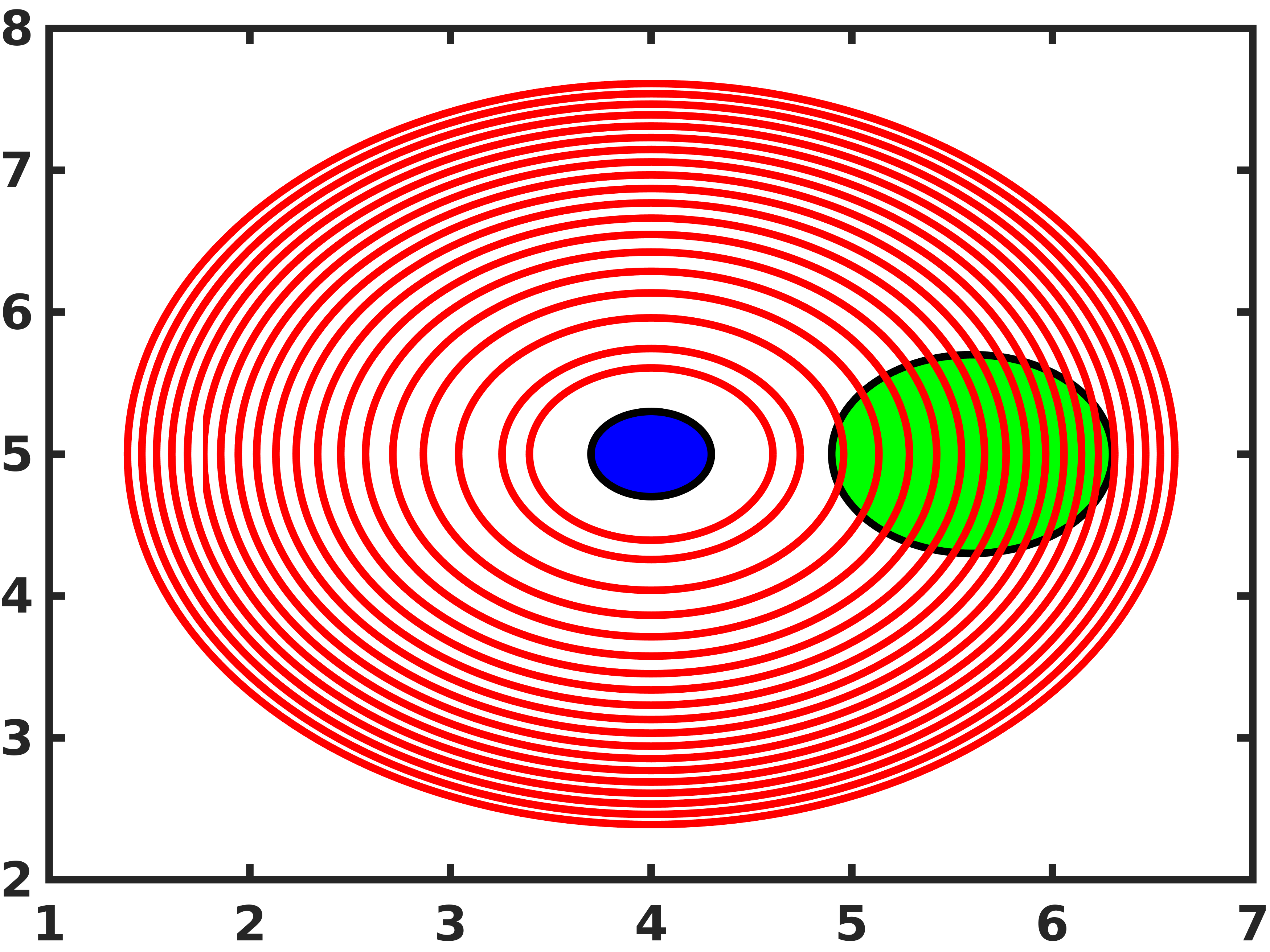}}
      \subfloat[\revtext{Collision probability evolution}]{\includegraphics[scale=0.3]{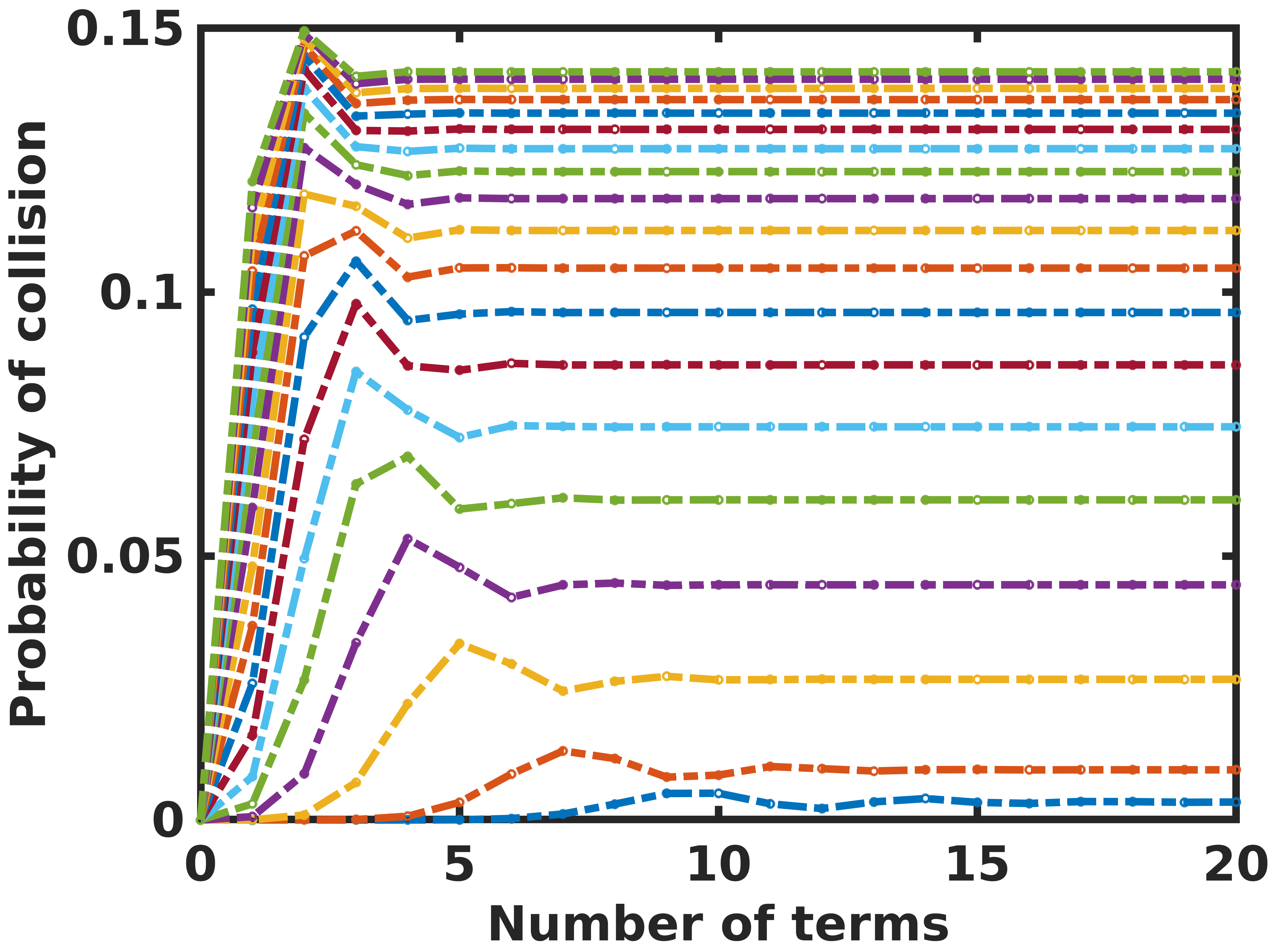}}\\
    \subfloat[Configuration G]{\includegraphics[scale=0.3]{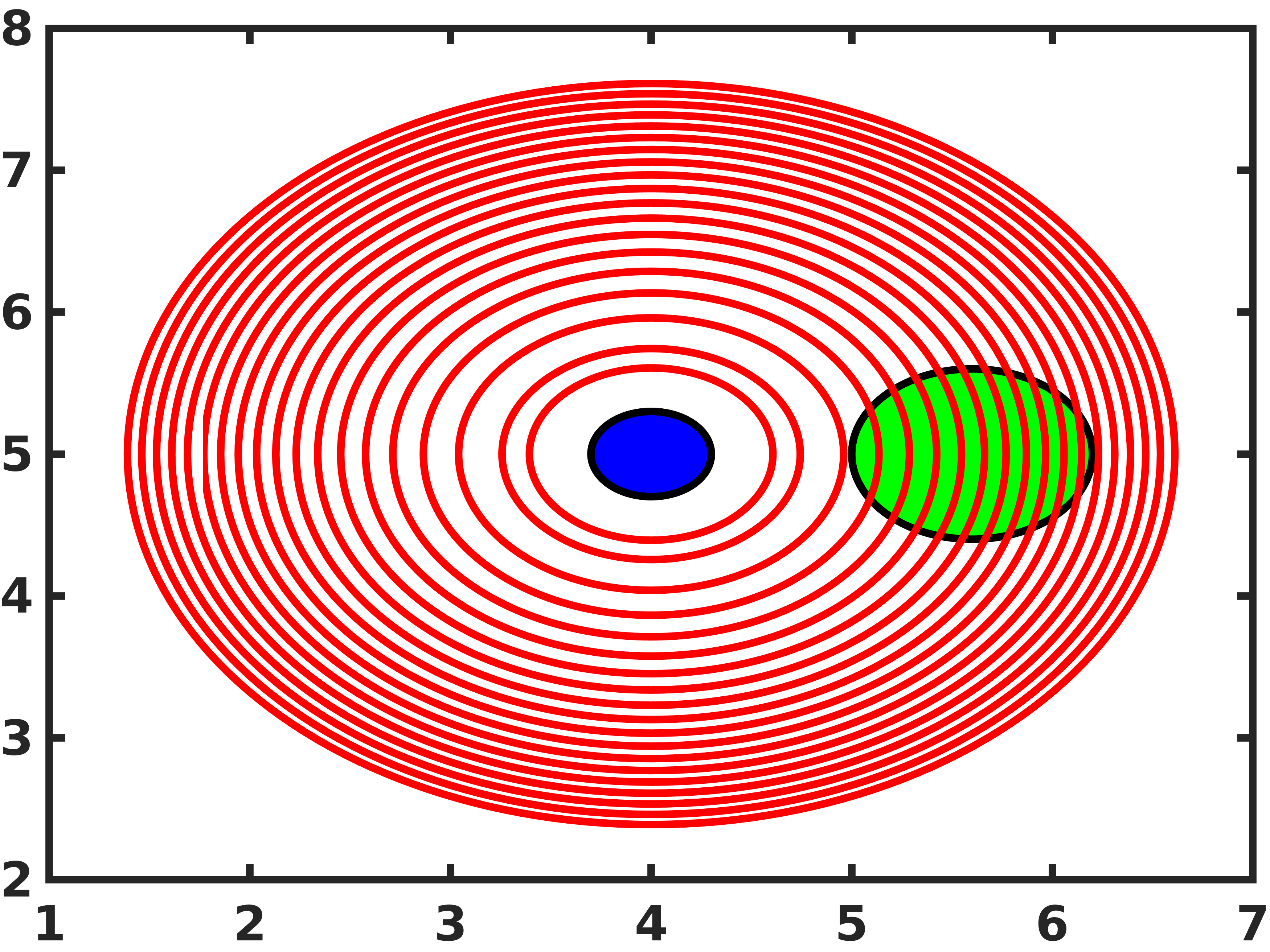}} 
    \subfloat[\revtext{Collision probability evolution}]{\includegraphics[scale=0.3]{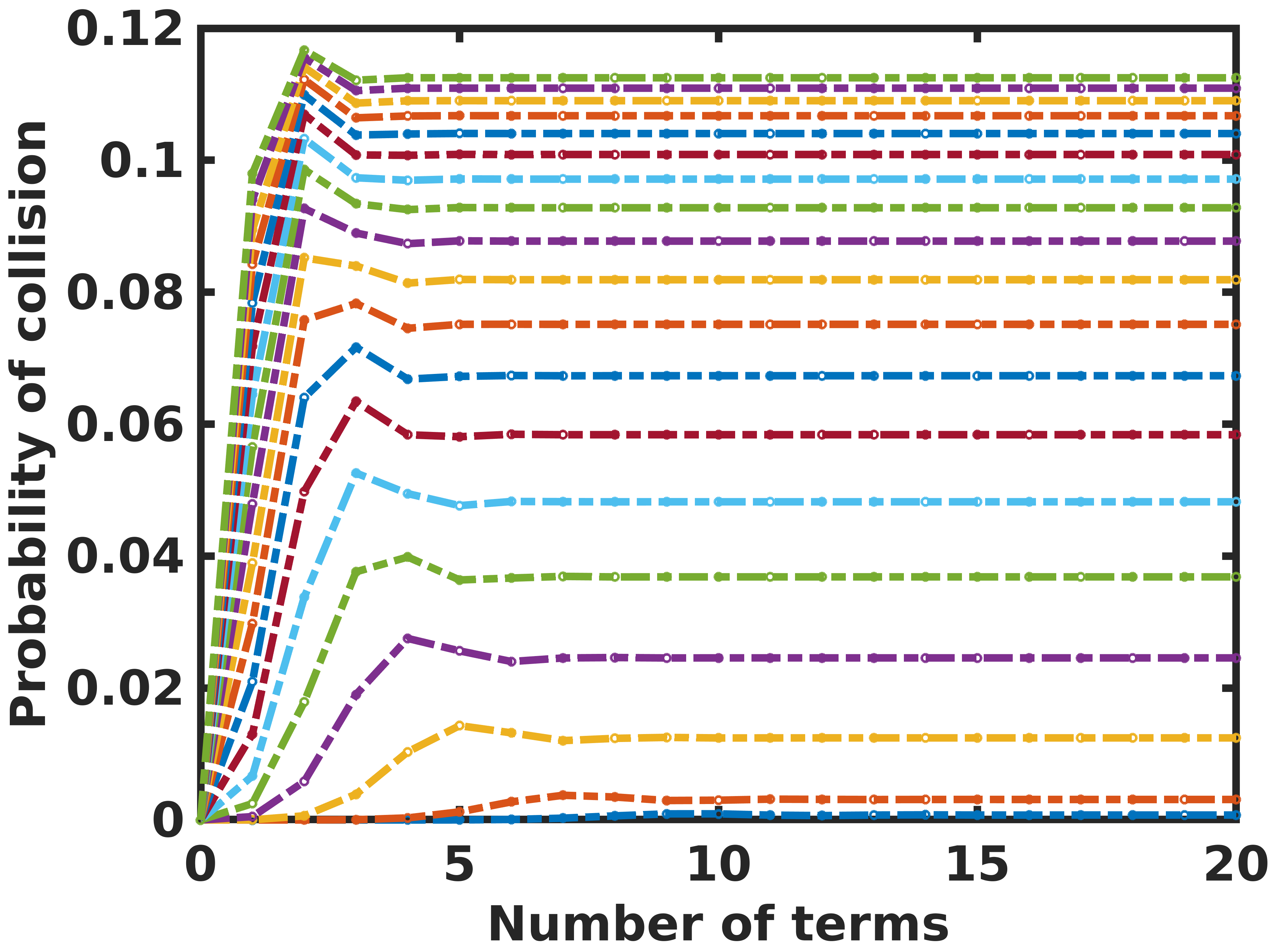}}\\
   \subfloat[Configuration G]{\includegraphics[scale=0.3]{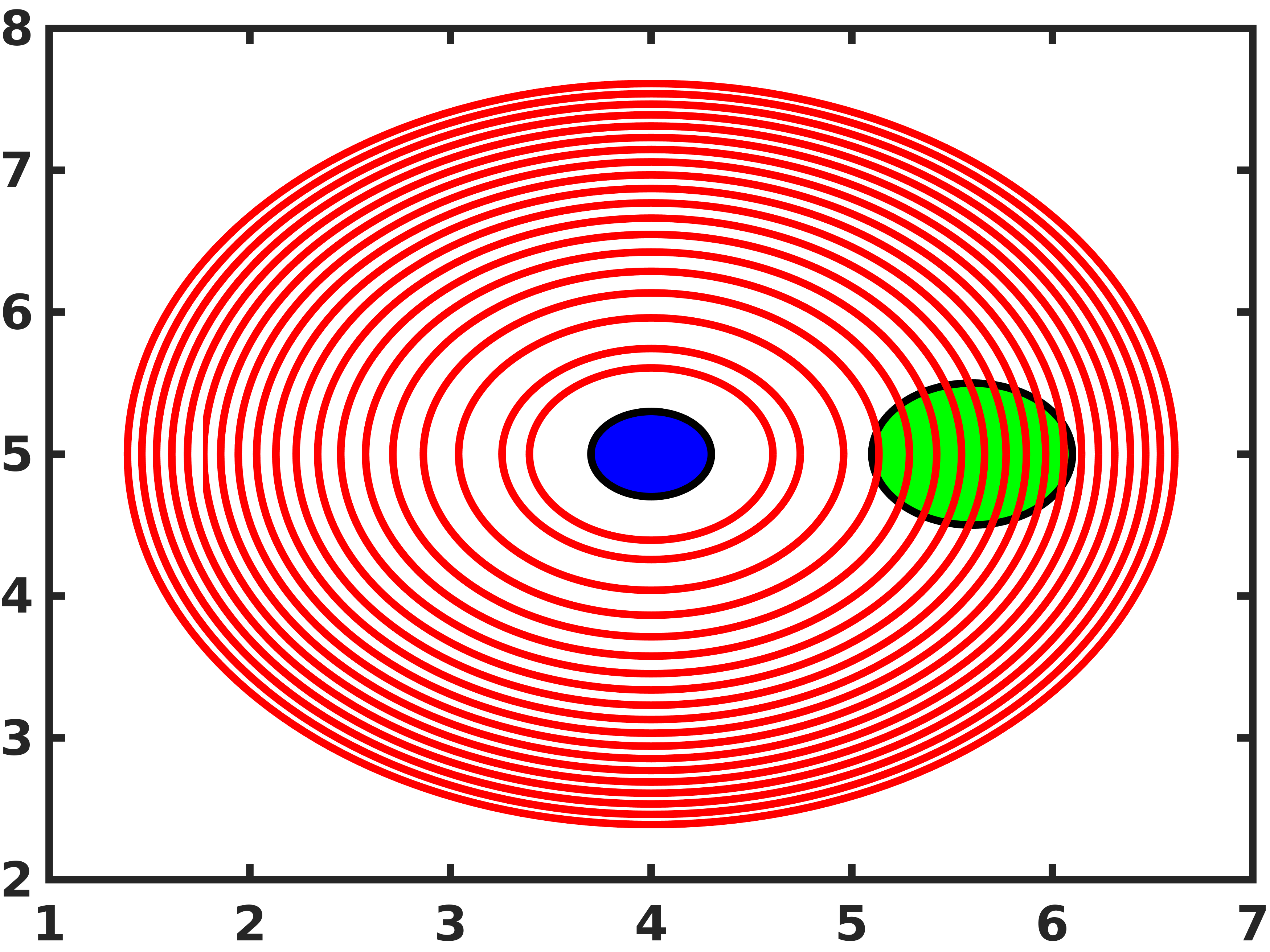}} 
    \subfloat[\revtext{Collision probability evolution}]{\includegraphics[scale=0.3]{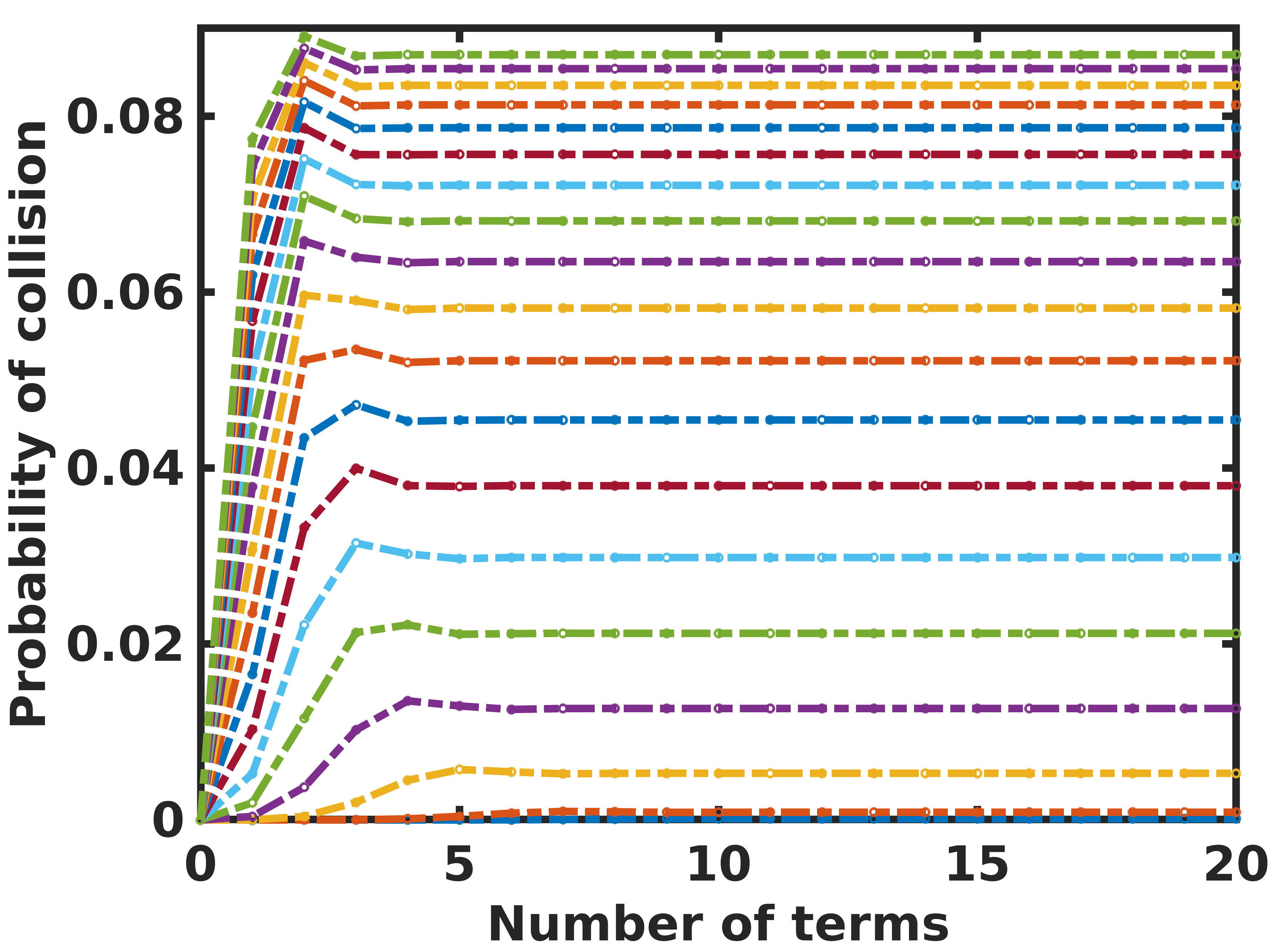}}
  \caption{Different configurations for a robot of radius 0.3$m$ and obstacle of radius (a) 0.8$m$, (c) 0.7$m$, (e) 0.6$m$ and (g) 0.5$m$. In the second column, for each of these configurations the evolution of collision probability is plotted against different covariances\revtext{--- the covariances are plotted as red circles in the figures on the left.}}
  \label{fig:y_convergence}
\end{figure}

It is worth noting that for a given robot configuration and obstacle parameters, the varying term in (\ref{eq:truncation}) is $(y/2\rho)^{N+1}/(N+1)!$. This term is inversely proportional to the parameter $\rho$. As discussed above $\rho$ depend on $\lambda_i$'s, that is, the eigenvalues of  $\Sigma_{\B{x}_k} + \Sigma_{\B{s}_k} $. Thus at time instant $k$, the parameter that influences the convergence is the degree of uncertainty in both the robot and obstacle locations, that is, $\Sigma_{\B{x}_k} + \Sigma_{\B{s}_k} $. This is visualized for different configurations in Fig.~\ref{fig:convergence}. The blue and green circles represent a robot and an obstacle, respectively. The red ellipses corresponds to the 3$\sigma$ uncertainties for different covariances $diag(0.04,0.04),\ diag(0.08,0.08),\ \ldots,\ diag(0.74,0.74)$. For all the scenarios discussed we choose $E(N) = 0.001$. In Fig.~\ref{fig:convergence}(a) the robot and the obstacle are touching each other. For each of these covariances, the number of terms for convergence is shown in Fig.~\ref{fig:convergence}(b). The worst case corresponds to the covariance of $diag(0.04,0.04)$, requiring 16 terms for convergence (dashed blue line with spikes in Fig.~\ref{fig:convergence}(b)). In Fig.~\ref{fig:convergence}(c) the distance between the robot and the obstacle is increased by 0.2$m$ and the covariance $diag(0.04,0.04)$ needs 12 terms for convergence. The distances are further increased by 0.4$m$ and 0.8$m$ in Fig.~\ref{fig:convergence}(e),(g) and their worst case convergences are 9 and 5 respectively, as seen in Fig.\ref{fig:convergence}(f),(h). The number of terms for the worst case convergence corresponds to covariance $diag(0.04,0.04)$ and the respective timings for collision probability computation are shown in Table~\ref{table1}.  

Similarly, the term $(y/2\rho)^{N+1}/(N+1)!$ is directly proportional to $y$ which quantifies the size of the robot and the obstacle. We recall here from~(\ref{eq:cdf}) that $y = (r_1+s_1)^2$, that is, the square of the sum of robot and obstacle radius. By keeping the robot size constant and varying the obstacle size, the influence of $y$ on convergence is visualized for four different configurations in Fig.~\ref{fig:y_convergence}. In Fig.~\ref{fig:y_convergence}(a) $y = 1.1^2 (m^2)$ and convergence is obtained within 7 terms. In Fig.~\ref{fig:y_convergence}(c),(e),(g) we have $y = 1^2, 0.9^2, 0.8^2$ and the number of terms required for convergence are  4, 3 and 2, respectively. The collision probability computation times are as given in Table~\ref{tab:size}. For $y > 1.1^2$, it can be seen that the number of terms for convergence did not exceed 7 and for $y < 0.8^2$ convergence is achieved with the first two terms. Thus this shows that $\rho$ plays a much larger role in convergence than $y$.

\begin{table}
\small\sf\centering
 \caption{The maximum number of terms required for convergence and the corresponding collision probability computation time. The values correspond to the covariance $diag(0.04,0.04)$ for each of the configurations.}
\begin{tabular}{ |c|c|c| } 
 \hline
 Configuration & Terms for convergence & Computation time (s) \\
 \hline 
 A & 16 & 0.0412 $\pm$ 0.0086 \\ 
  \hline 
 B & 12 & 0.0044 $\pm$ 0.0041 \\ 
 \hline
 C & 9  & 0.0008 $\pm$ 0.0003 \\
 \hline
 D & 5  & 0.0004 $\pm$ 0.0002 \\
 \hline
\end{tabular}
 \label{table1}
\end{table}
\begin{table}
\small\sf\centering
 \caption{The maximum number of terms required for convergence and the corresponding collision probability computation time. Each configuration corresponds to different $y$ values with the robot and obstacle locations remaining the same; only obstacle size varies.}
\begin{tabular}{ |c|c|c| } 
 \hline
 Configuration & Terms for convergence & Computation time (s) \\
 \hline 
 E & 7 & 0.0006 $\pm$ 0.0005 \\ 
  \hline 
 F & 4 & 0.0004 $\pm$ 0.0002 \\ 
 \hline
 G & 3  & 0.0004 $\pm$ 0.0001 \\
 \hline
 H & 2  & 0.0001 $\pm$ 0.0000 \\
 \hline
\end{tabular}
 \label{tab:size}
\end{table}
\subsection{Safe Configurations}
In the presence of perception and motion uncertainty, providing safety guarantees for robot navigation is imperative. In this Section, we certify safety by defining the notion of a ``safe" robot configuration. Let us assume that the obstacle position is known with high certainty as a result of perfect sensing, that is, no significant noise is present. However, since the true state of the robot is not known and only a distribution of these states can be estimated, collision checking has to be performed for this distribution of states. Moreover, in practice, the observations are noisy and this renders the estimated obstacle location (and shape) uncertain. Hence, this uncertainty should be taken into account while considering collision avoidance. 

Given a robot configuration $\B{x}_k$, we define the following notion of $\epsilon-$safe configuration.

\begin{definition}
A robot configuration $\B{x}_k$ is an $\epsilon-$safe configuration with respect to an obstacle configuration $\B{s}$, if the probability of collision is such that $P\left(\mathcal{C}_{\B{x}_k,\B{s}}\right) \leq 1 - \epsilon$.
\end{definition}
For example, a $0.99-$safe configuration implies that the probability of this configuration colliding with the obstacle is at most $0.01$. On the one hand, sampling-based motion planning approaches such as the Probabilistic Roadmap (PRM)~\cite{kavraki1996IEEE} consider a discrete state space or a set of controls. As a result, it can only guarantee probabilistic completeness for returning $\epsilon-$safe configurations since the PRM motion planner is probabilistically complete~\cite{karaman2011IJRR}, that is the probability of failure decays to zero exponentially with the number of samples used in the construction of the roadmap. As a result, for sampling-based BSP approaches~\cite{agha_mohammadi2014IJRR, prentice2009IJRR}, the failure to find an $\epsilon-$safe configuration might be because such a configuration indeed does not exist or simply because there are not enough samples. On the other hand, continuous state and action space BSP approaches~\cite{van_den_berg2012IJRR, platt2010RSS, patilWAFR14, indelman2015IJRR} do not always guarantee $\epsilon-$safe configurations. This is merely because there might not be enough measurements to localize the robot or to estimate obstacle locations or both and hence this may preclude computing appropriate control commands.
\subsection{Comparison to Other Approaches} 
\label{sec:comparison}
\begin{figure}[t]
  \subfloat[]{\includegraphics[scale=0.3]{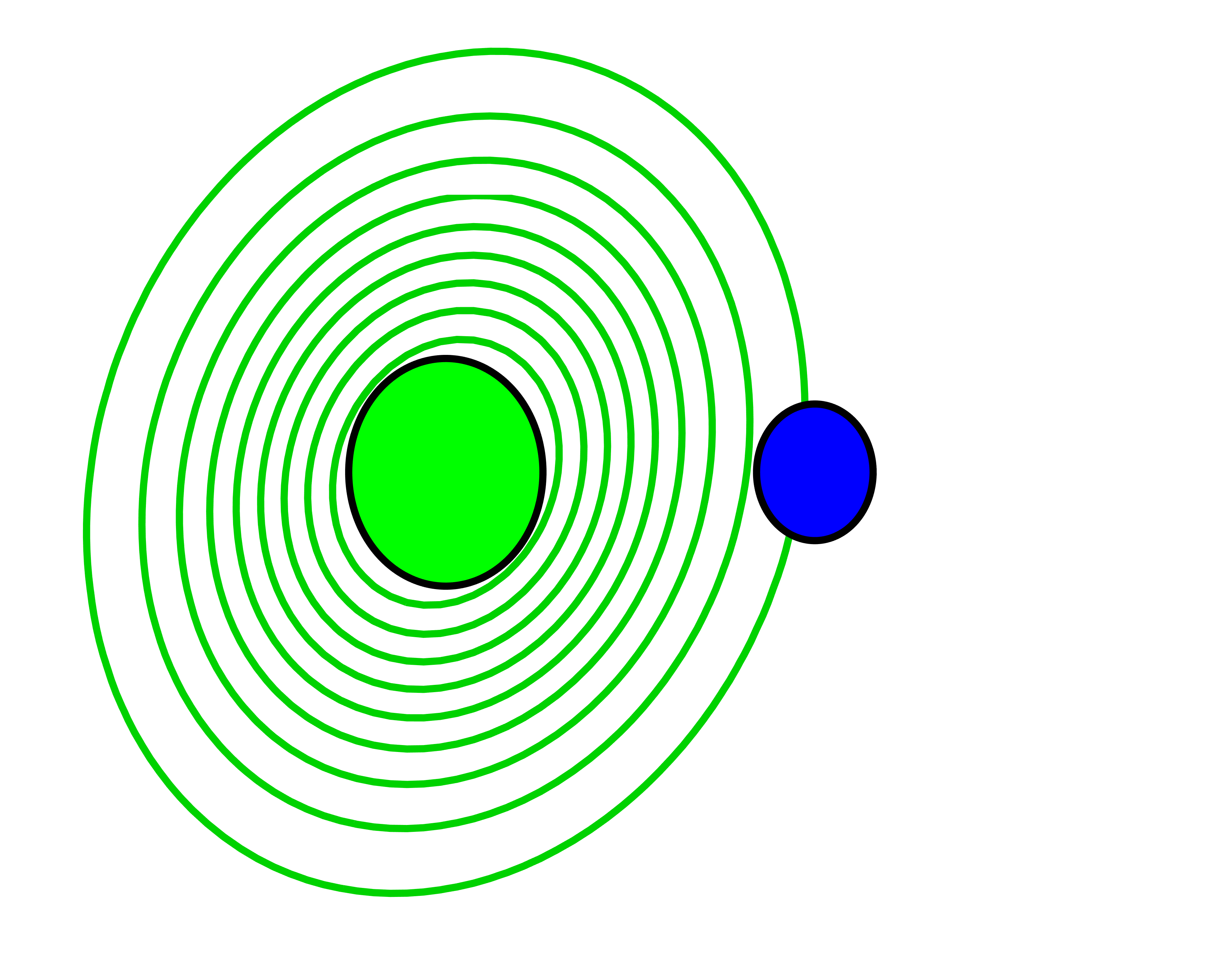}} 
  \subfloat[]{\includegraphics[scale=0.3]{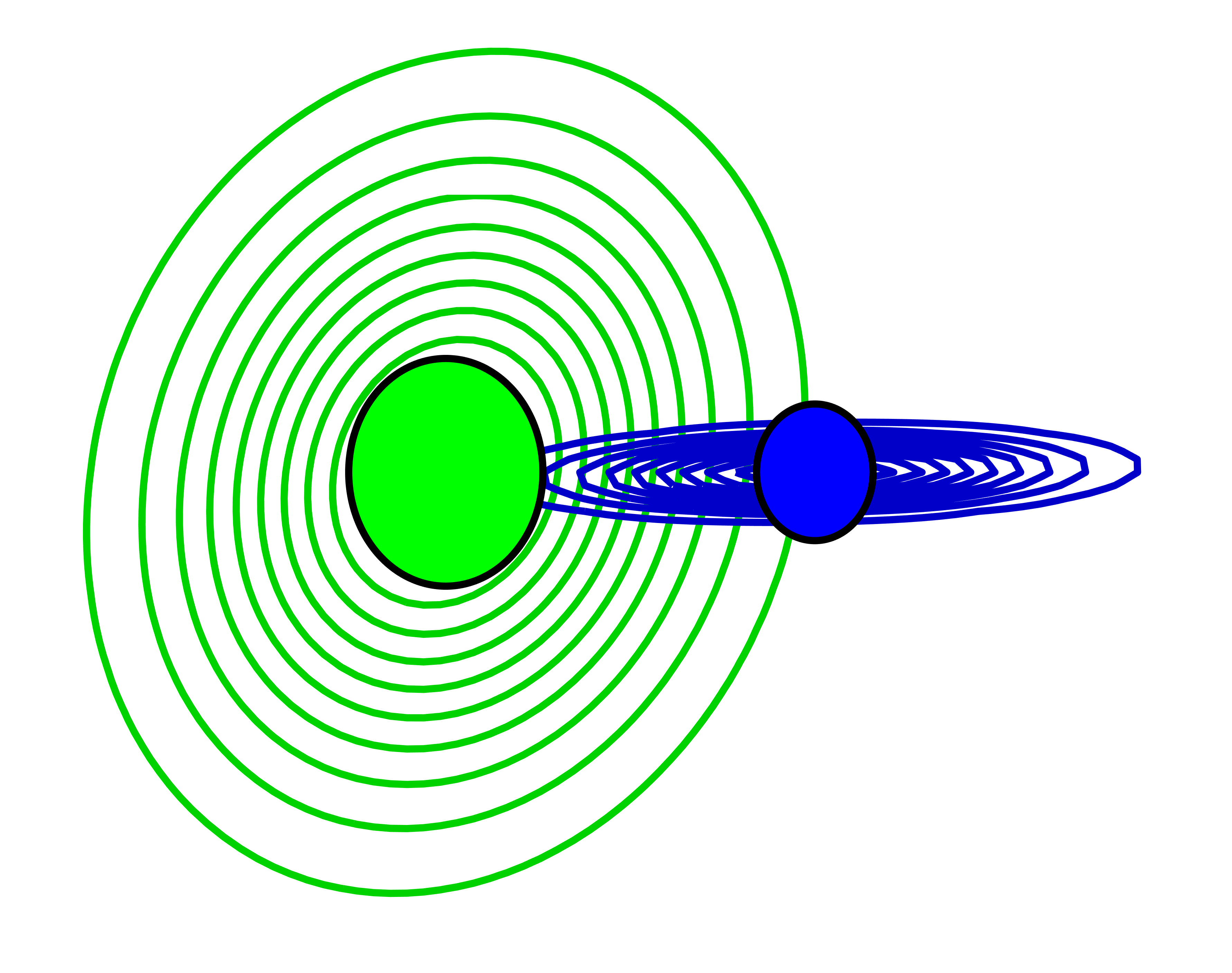}}\hspace{0.2cm} 
   \subfloat[]{\includegraphics[scale=1,height=2.3cm]{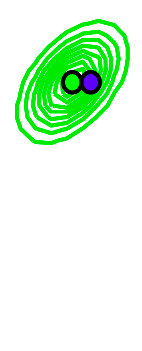}}
  \caption{Comparison of our approach with~\cite{lambert2008ICCARV,zhu2019RAL,park2018IEEE,dutoit2011IEEE}. (a) The robot state (in blue) is known perfectly, however the obstacle location (in green) is uncertain.~\revtext{The green ellipses denote the Gaussian uncertainty contours.}  (b) Robot state uncertainty is considered~\revtext{and the uncertainty contours are shown as blue ellipses}. The approaches in~\cite{dutoit2011IEEE,park2018IEEE, zhu2019RAL} computes higher values. (c) Point-like robot and obstacle considered. The values computed with~\cite{dutoit2011IEEE,park2018IEEE} are much lower than expected while that of~\revtext{\cite{zhu2019RAL}} is very high.}
  \label{fig:comparison}
\end{figure}

\begin{table*}
\small\sf\centering
\caption{Comparison of collision probability approaches.}
\begin{tabular}{ |c|c|c|c|c| }
\hline
Case & Algorithm & Collision probability & Computation time (s)& Feasible \\ 
 \hline
\multirow{6}{*}{(a)} & Numerical integral & $4.62\%$ & 0.8648 $\pm$ 0.0418 &Yes \\& ~\cite{lambert2008ICCARV} & $4.41\%$ & 0.0272 $\pm$ 0.0023 &Yes \\
   & ~\cite{dutoit2011IEEE} & $5.84\%$ &0.0017 $\pm$ 0.0002 &Yes\\ 
   & ~\cite{park2018IEEE} & $33.26\%$ &0.2495 $\pm$ 0.3093 &No \\  
  & ~\cite{zhu2019RAL} & $9.60\%$ &0.0021 $\pm$ 0.0003 &No \\
  & ~Our approach & $4.61\%$ & 0.0254 $\pm$ 0.0034 &Yes \\ \hline
 \multirow{6}{*}{(b)} & Numerical integral & $8.25\%$ & 1.1504 $\pm$ 0.0318 &Yes \\ & ~\cite{lambert2008ICCARV} & $7.87\%$ &0.0325 $\pm$ 0.0024 &Yes\\
   & ~\cite{dutoit2011IEEE} & $14.20\%$ &0.0011$\pm$ 0.0002 & No\\ 
 & ~\cite{park2018IEEE} & $36.31\%$ & 0.2156 $\pm$ 0.4068& No \\
 & ~\cite{zhu2019RAL} & $16.73\%$ & 0.0013 $\pm$ 0.0003&No \\
 & ~Our approach & $8.22\%$ &0.0216 $\pm$ 0.0023 &Yes \\ \hline
 \multirow{6}{*}{(c)} & Numerical integral & $14.82\%$ & 1.1341 $\pm$ 0.0211 &No \\ & ~\cite{lambert2008ICCARV} & $15.26\%$ & 0.0287 $\pm$ 0.0059 &No \\
 & ~\cite{dutoit2011IEEE} & $0.46\%$ & 0.0015 $\pm$ 0.0007 &Yes\\ 
 & ~\cite{park2018IEEE} & $0.61\%$ & 0.3233 $\pm$ 0.5405&Yes \\
   & ~\cite{zhu2019RAL} & $50.00\%$ & 0.0018 $\pm$ 0.0007 &No \\
 & ~Our approach & $14.83\%$ & 0.0280 $\pm$ 0.0093 &No \\ \hline
\end{tabular}
\label{tab:comparison}
\end{table*}

\cite{lambert2008ICCARV} compute the collision probability by performing MCI. The joint distribution between the robot and the obstacle $p(\B{x}_k,\B{s}_k)$ is simplified as the product of the individual distributions. This MCI approach results in an expression with double summation for computing the probability of collision.~\cite{lambert2008ICCARV} approximate this to a single summation expression to decrease computational complexity.~\revtext{Though this approximation compute values closer to the actual collision probability, it can either be bounded from below (when uncertainty is too large) or above. The approches in~\cite{dutoit2011IEEE,park2018IEEE,zhu2019RAL} compute upper bounds for collision probability.~\cite{zhu2019RAL} compute an upper bound using Gaussian chance constraints.}~\cite{park2018IEEE} compute the collision probability by finding the $\B{x}_k$ that maximizes $p(\B{x}_k,\B{s}_k)$ and formulate the problem as an optimization problem with a Lagrange multiplier. Unlike in~\cite{park2018IEEE}, which computes the maximum density,~\cite{dutoit2011IEEE} use the density associated with the center of the robot.~\revtext{Yet,~\cite{dutoit2011IEEE,park2018IEEE} compute lower values when the joint robot and obstacle covaraince is very small.} We formulate the problem as exactly given in each of the works mentioned above to compare it with our approach\footnote{For comparison, the computation of other approaches have been reproduced to the best our understanding and the reproduced codes can be found here- \url{https://bitbucket.org/1729antony/comparison/src/master/}}. The MCI approach of~\cite{lambert2008ICCARV} is evaluated using $10,000$ samples.~\revtext{The numerical integration of the expression in~(\ref{eq:collision_prob}) gives the exact collision probability value. Thus to validate the value computed using our approach, we perform the numerical integration of~(\ref{eq:collision_prob}), using Monte Carlo method with $10,000$ samples.} 

Three different cases are considered as shown in Fig.~\ref{fig:comparison}. The solid green circle denotes an obstacle of radius $0.5m$ and its corresponding uncertainty contours are shown as green circles. The solid blue circle denotes a robot of radius $0.3m$ with the blue circles showing the Gaussian contours. We define a collision probability threshold of $0.09$, that is, a $0.91-$safe configuration. The collision probability values and the computation times are summarized in Table~\ref{tab:comparison}. In Fig.~\ref{fig:comparison}(a), the robot position is known with high certainty.~\revtext{The numerical integration of~(\ref{eq:collision_prob}) gave a value of $4.62\%$ and hence the given configuration is a $0.91-$safe configuration. Our approach computes the collision probability as $4.61\%$, corroborating the exactness}. 
~\revtext{The approach of~\cite{lambert2008ICCARV} gave a close value of $4.41\%$ but is a lower bound for the actual value. The other three approaches compute upper bounds as discussed previously.~\cite{dutoit2011IEEE} estimates the configuration to be feasible, giving a collision probability value of $5.84\%$.} The collision probability computed as given in~\cite{park2018IEEE} is $33.26\%$ (not a $0.91-$safe configuration). Moreover, the value computed is almost seven times higher than the one computed using our approach. Similarly, the value computed using the approach in~\cite{zhu2019RAL} is $9.60\%$, predicting the configuration to be unsafe. The higher values are due to the overly conservative nature of the estimates as loose upper bounds are computed. In Fig.~\ref{fig:comparison}(b), there is robot uncertainty along the horizontal axis and~\revtext{the numerical integration gave a collision probability value of $8.25\%$. As compared to the previous case, the probability has almost doubled. This is quite intuitive as seen from the robot location uncertainty spread and hence there is greater chance for intersection between the two spheres. The collision probability computed using our approach is $8.22\%$}.  The increased chance for collision is also rightly communicated by the values computed using other approaches. The value computed using the approach in~\cite{park2018IEEE} gave a much higher value of $36.31\%$, an increase by $342\%$ as compared to our approach. As in the previous case, the approaches in~\cite{dutoit2011IEEE},~\cite{zhu2019RAL} also gave higher values of $14.20\%$ and $16.73\%$, respectively, while~\cite{lambert2008ICCARV} gave a feasible value of $7.87\%$~\revtext{but a value lower than the actual}.  

The approaches in~\cite{dutoit2011IEEE,park2018IEEE} assume that the robot radius is negligible and that the obstacle size is relatively small compared to their location uncertainties. We also compute the collision probabilities for a robot and an obstacle with radius $0.05m$ each, where the robot and the obstacle are touching each other (Fig.~\ref{fig:comparison}(c)). The obstacle location is also much more certain, with the uncertainty reduced by $97\%$ as compared to cases in Fig.~\ref{fig:comparison}(a),(b). The numerical integration gave a collision probability value of $14.82\%$. The probability of collision computed using our approach is $14.83\%$, whereas, using the approach in~\cite{park2018IEEE}, the computed value is $0.61\%$. A lower value of $0.46\%$ is obtained using the approach in~\cite{dutoit2011IEEE}.~\revtext{As noted before, the lower values are a consequence of the covariance being very small}. The approach of~\cite{zhu2019RAL} gave an overly conservative estimate of $50\%$. The value computed using~\cite{lambert2008ICCARV} is $15.26\%$~\revtext{, an upper bound to the actual value}. To get a sense of the actual value, we compute the area of the covariance matrix, which is $6.28 \times 10^{-4}m^2$. This clearly indicates that $0.61\%$, $0.46\%$ and $1.69\%$ are too small values while $50.00\%$ is a very high value. Our approach computes the exact probability of collision and outperforms the approaches in~\cite{lambert2008ICCARV,dutoit2011IEEE,park2018IEEE,zhu2019RAL}.

\begin{figure}[t]
\centering
  \subfloat[]{\includegraphics[scale=0.3]{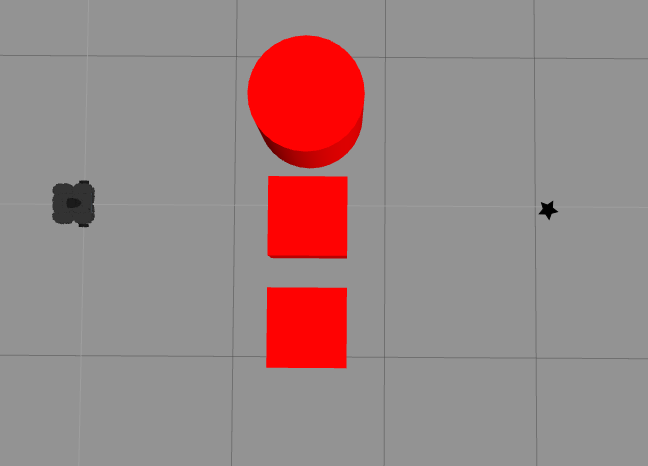}} \hspace{0.4cm}%
  \subfloat[]{\includegraphics[scale=0.34]{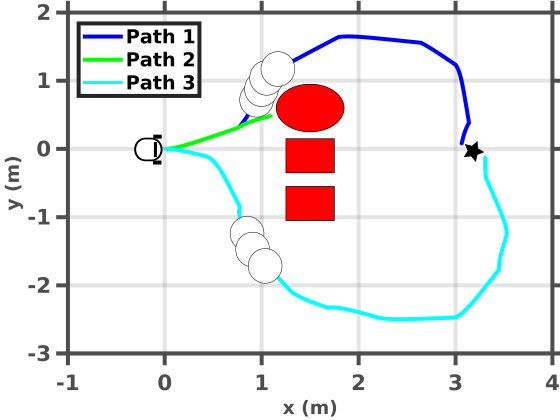}}
  \caption{Comparison to other approaches in simulation: (a) Top view of the environment in gazebo. (b) Path 1 is the trajectory executed by the robot following our approach. The trajectory is executed in seven planing sessions.~\revtext{The robot footprint can be seen as it navigates past the obstacle. Path 2 leads to collision as upper bounds are computed by other approaches deeming the plans infeasible-- robot is hence halted.} The action set is extended and a longer trajectory is executed by the robot using the approach of~\cite{zhu2019RAL}. The goal is reached in 15 planning sessions.}
  \label{fig:gazebo_compare}
\end{figure}

We now provide a comparison in simulation using a scenario shown in Fig.~\ref{fig:gazebo_compare}(a). The robot has to reach the goal position (black star) by avoiding the obstacles in-between. To make the implications of overly conservative \revtext{estimates~\cite{dutoit2011IEEE,park2018IEEE,zhu2019RAL}} explicit, we make the following assumptions. During each planning session\footnote{By a planning session we mean an $L$ look-ahead step planning at the current time and choosing an optimal control. Thus if $n$ planning sessions are required to reach a goal this means that the control action was executed $n$ times.}, the robot can~\revtext{choose from a restrcitive set of nine different actions}. A description of the motion and observation models can be found in Section~\ref{experiments}. An action is chosen based on an additive cost of distance to the goal and the collision probability~\revtext{value}. If the collision probability for an action is greater than $0.01$, then~\revtext{the collision probability value for this action} is penalized by redefining the value to be equal to a large number $M$.~\revtext{In this way, if all the actions lead to configurations with collision probability greater than $0.01$, these actions are assigned a cost $M$. In reality this would mean that there exist no feasible plan and the robot would not proceed ahead. Thus if all actions are assigned a collision probability value of M, all these actions lead to collision and therefore we mark the trajectory as stopped.} The trajectory executed by the robot using our collision probability computation approach is shown in blue (Path 1) in Fig.~\ref{fig:gazebo_compare}(b).~\revtext{The robot footprint (bounded circle) is also shown as the robot curves past the obstacle.} The goal was reached in seven planning sessions.~\revtext{For the approaches in~\cite{dutoit2011IEEE,park2018IEEE,zhu2019RAL} the trajectory is stopped (Path 2) before the obstacle since all actions are assigned a value of M due to collision probability values greater than 0.01}.~\revtext{As noted before, the action set from which the robot can choose an action is restriced and each action from this restricted set gives configurations with collision probability greater than $0.01$. This is due to the fact that these approaches compute loose upper bounds and hence the values in the collision cost are redefined to $M$.}~\revtext{The restrictive action set does not affect our approach as the exact value is computed} and hence the robot reaches the goal safely.~\revtext{We now remove the restriction on the action set and} all the other approaches are able to compute a path with a greater curve than Path 1. One such trajectory is shown in cyan (Path 3), with the collision probabilities computed using the approach by~\cite{zhu2019RAL}.~\revtext{The planner is now able to choose an action with collision probability less than $0.01$.} Thus it is seen that loose upper bounds for collision probability can lead to longer trajectories or in some cases deem all plans to be infeasible. 

\subsection{Non Circular Geometry}
Given two objects (represented as convex polygons), in this Section we derive the collision constraint as a measure of the distance between the mid points of the objects. As before, let us consider two objects, $\mathcal{R} \subset \mathbb{R}^2$ and $\mathcal{S} \subset \mathbb{R}^2$. Let us assume that $\mathcal{S}$ is static and $\mathcal{R}$ can perform translational motions and is approaching $\mathcal{S}$. Then, subtracting $\mathcal{R}$ from $\mathcal{S}$ gives a convex polygon $\mathcal{P}$ such that for any $c \in \mathcal{P}$, then $\mathcal{R} \cap \mathcal{S} \neq \{\phi\}$~(\cite{perez1983TCO}), that is, the convex polygon $\mathcal{P}$ is the set of configurations of $\mathcal{R}$ that leads to collision with obstacle $\mathcal{S}$. Note that, $\mathcal{R}$ and $\mathcal{S}$ are essentially two sets whose elements are the $(x,y)$ pairs belonging to the respective polygons that they represent. Therefore, $\mathcal{P}$ is essentially the Minkowski difference between the two sets $\mathcal{R}$ and $\mathcal{S}$.
\begin{definition}
The Minkowski sum of two sets $\mathcal{S}$, $\mathcal{R} \subseteq \mathbb{R}^d$ is
\begin{equation}
\mathcal{S} + \mathcal{R} = \{s + r \ | \ s \in \mathcal{S}, \ r \in \mathcal{R} \}
\end{equation}
\end{definition}

\begin{definition}
The Minkowski difference of two sets $\mathcal{S}$, $\mathcal{R} \subseteq \mathbb{R}^d$ is
\begin{equation}
\mathcal{S} - \mathcal{R} = \{s - r \ | \ s \in \mathcal{S}, \ r \in \mathcal{R} \}
\end{equation}
\end{definition}

\begin{figure}[t]
  \subfloat[Convex objects]{\includegraphics[scale=0.6]{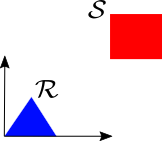}} \hfill
  \subfloat[Minkowski sum]{\includegraphics[scale=0.6]{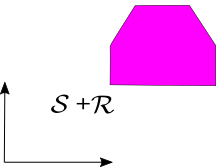}} \hfill
   \subfloat[Minkowski difference]{\includegraphics[scale=0.6]{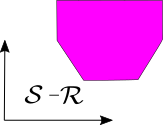}}
  \caption{(a) Two convex objects and the object formed by considering the (b) Minkowski sum and (c) Minkowski difference.}
  \label{fig:minkowski}
   \end{figure}

The Minkowski sum and difference of two objects are visualized in Fig.~\ref{fig:minkowski}. The Minkowski difference between the two sets $\mathcal{S}$ and $\mathcal{R}$, also called the configuration space obstacle, is the set of (translational) configurations of $\mathcal{R}$ that brings it into collision with $\mathcal{S}$~(\cite{perez1983TCO,cameron1986ICRA}). However, we would like to obtain a collision constraint of the form~(\ref{eq:coll_condition}). In order to obtain such a constraint, we first compute the Minkowski difference between the set $\mathcal{S}$ and the mid-point of $\mathcal{R}$. This gives a new convex set $\mathcal{P'}$ whose elements are formed by subtracting each element of the set $\mathcal{S}$ by the mid-point of object $\mathcal{R}$. In other words, the set $\mathcal{P'}$ is the set of all configurations of the mid-point of $-\mathcal{R}$\footnote{It holds that $-\mathcal{R} = \{-r \ | \ r \in \mathcal{R} \}$} obtained by shifting/translating this point by each element in the set $\mathcal{S}$.
%
\begin{lemma}
The maximum distance from a point $P$ to any other point on a polygon $\mathcal{Q}$ is obtained by computing 
\begin{equation}
\sup{\{PV_i \ | \ V_i \ \textrm{is the vertex of } \ \mathcal{Q}\}}
\end{equation}
where $PV_i$ denotes the line segment from point $P$ to the vertex $V_i$.
\label{lemma3}
\end{lemma}
\textit{Proof.} Given a point inside (or outside) a polygon,~\revtext{farthest distance to any other point on the polygon is obtained when this point lies on the boundary of the polygon. It is known that the maximum distance from a point to a line segment occurs at the end-points of the line segment. Hence it is only sufficient to compute the distances to the vertices of the polygon and the maximum among them is the required distance.} 

For convex polygons $\mathcal{R}$ and $\mathcal{S}$, the boundary configurations of the Minkowski difference represents configurations that lead to contact between $\mathcal{R}$ and $\mathcal{S}$~(\cite{cameron1986ICRA}), that is, the configurations where $\mathcal{R}$ and $\mathcal{S}$ touch each other. Also note that the polygon $\mathcal{P'}$ obtained by computing the Minkowski difference between the mid-point of $\mathcal{R}$ and the set $\mathcal{S}$ is fundamentally the set of all translated configurations of the mid-point of $-\mathcal{R}$ in the set $\mathcal{P} = \mathcal{S}- \mathcal{R}$. Thus the collision constraint is obtained by computing the maximum distance between the mid-point of the obstacle $\mathcal{S}$ and the polygon $\mathcal{P'}$.

\begin{theorem}
Given a convex polygonal set $\mathcal{R}$ and an obstacle set $\mathcal{S}$, the collision constraint is given by
\begin{equation}
\sup{\{SV_i \ | \ V_i \ \textrm{is the vertex of } \ \mathcal{P'}\}}
\end{equation}
where $S$ is the mid-point of $\mathcal{S}$ and $\mathcal{P'}$ is the set obtained by computing the Minkowski difference between $\mathcal{S}$ and the mid-point of $\mathcal{R}$.
\label{theorem2}
\end{theorem}
\textit{Proof.} We saw above that the collision constraint is obtained by computing the maximum distance between the mid-point of the obstacle $\mathcal{S}$, that is $S$ and the polygon $\mathcal{P'}$. From Lemma~\ref{lemma3}, the maximum distance is achieved at the vertices of the polygon. Hence, it follows from Lemma~\ref{lemma3} that the collision constraint is $\sup{\{SV_i \ | \ V_i \ \textrm{is the vertex of } \ \mathcal{P'}\}}$. 

Thus, if $\mathcal{R}$ and $\mathcal{S}$ correspond to the set of points occupied by the robot and the obstacle, respectively, the collision constraint in~(\ref{eq:coll_condition}) can be written as $\norm{\B{x}_k - \B{s}_k}^2 \leq \left( \sup{\{SV_i\}}\right)^2$.

The Minkowski sum or difference are not invariant to rotations and hence rotation about a reference axis elicits different sets. The resulting sets are obtained by pre-multiplying the starting configuration with the standard rotation matrix of the corresponding angle. This renders different collision constraints for the two given sets. However, while planning for future control commands, the robot pose is often estimated using the motion model and by simulating possible future observations. As a result, an estimate of the robot orientation  is computed. Moreover, for static obstacles, both in known and unknown environments, the geometry of the obstacle is a constant\footnote{In this work we assume non-deformable objects.}. In the case of dynamic obstacles, the orientation of this geometry changes. Thus, assuming that the orientation of the obstacle is known and using the estimated robot orientation, the collision constrained is obtained as elucidated in Theorem~\ref{theorem2}. 

\subsection{Complexity Analysis} 
Finding a trajectory to the goal requires performing Bayesian (EKF) update operations. This involves performing matrix operations, that is, matrix multiplication and inversion of matrices. For a state of size $n$, the covariance matrix is of size $O(n^2)$. Therefore, each step of the Bayesian update has a complexity of $O(n^3)$. Let $L$ denote the number of time steps in the trajectory or the look-ahead horizon, then the overall computational complexity is $O(n^3L)$. Note that this is the complexity while computing the objective function at each time step. The number of times the computation is to be performed cannot be expressed beforehand as it depends on the specific application and objective to be achieved. Let us now analyze the complexity of collision probability computation. From (\ref{eq:truncation}) we see that for each iteration, the truncation error varies with $(y/2\rho)$. Therefore, to achieve $E(N) \leq \delta$, for an $\epsilon-$safe configuration, $k = O\left(\log \frac{\delta \rho}{y(1-\epsilon)}\right)$ iterations are required. We note that for each obstacle, the runtime is increased by this factor.

\section{Obstacle State Estimation}
\label{sec:obs}
\begin{figure}[t!]
\centering
 \subfloat{\includegraphics[scale=0.4]{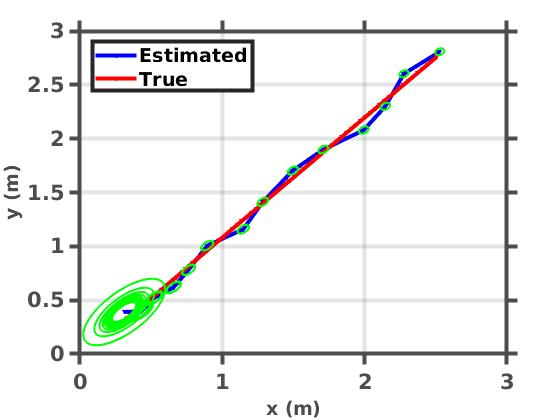}} 
  \subfloat{\includegraphics[scale=0.4]{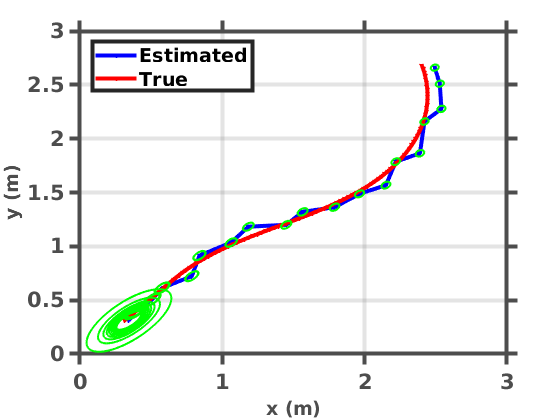}}\\ \vspace{-0.2cm}
    \subfloat{\includegraphics[scale=0.4]{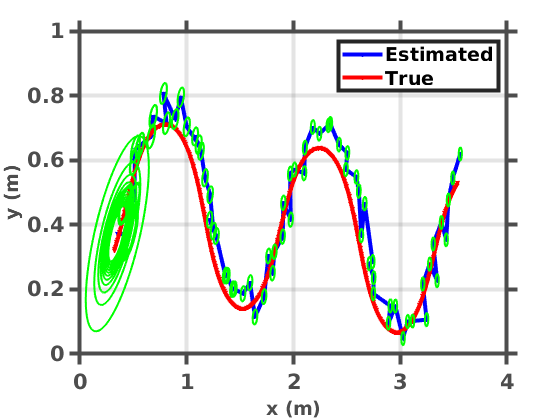}}
   \subfloat{\includegraphics[scale=0.4]{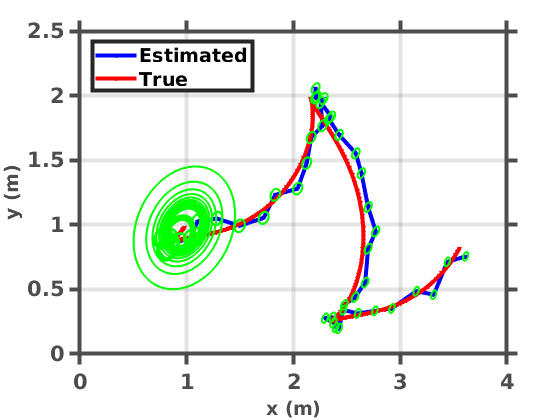}}
       \caption{True obstacle trajectories plotted along with the estimated obstacle trajectories. In all the cases the linear velocity of the obstacle is greater than or equal to 0.5$m/s$. A laser rangefinder is placed at the origin pointing towards the north-east direction. The green ellipses show the estimated covariances. Initial large ellipses correspond to the prior uncertainties. The prior uncertainties shrink as measurements are obtained due to obstacle detection.}
  \label{fig:obs_predict}
   \end{figure}
 We adapt the approaches in~\cite{schulz2001RAS, park2018IEEE} and describe below the process for estimating future obstacle states. Let us consider that at time instant $k$, the robot at state is $\B{x}_k$ and the estimated obstacle location is $\B{s}$. Since the obstacle is following an unknown trajectory, the robot receives a series of measurements $\B{z}_k^1,\ldots,\B{z}_k^n$. Note that since the obstacle is moving, then each measurement $\B{z}_k^i$ corresponds to a different location of the dynamic obstacle. Given the robot pose $\B{x}_k$ and the measurement $\B{z}_k^i$, the obstacle location $\B{s}^i$ can be estimated using the Bayesian approach,
\begin{equation}
p(\B{s}^i|\B{x}_k, \B{z}_k^i) = \eta \ p(\B{z}_k^i|\B{x}_k,\B{s}^i) \ p(\B{s}^i|\B{x}_k)
\end{equation}
\noindent where $\eta = 1/p(\B{z}_k^i|\B{x}_k)$ is the normalization constant. Since the obstacle state $\B{s}^i$ is independent of the robot state $\B{x}_k$, we obtain
\begin{equation}
p(\B{s}^i|\B{x}_k, \B{z}_k^i) = \eta \ p(\B{z}_k^i|\B{x}_k,\B{s}^i) \ p(\B{s}^i)
\label{eq:obs_update}
\end{equation} 
\noindent where $p(\B{s}^i)$ is the prior density. Given the current robot belief $b[\B{x}_k]$ and the measurement $\B{z}_k^i$, the expression $p(\B{z}_k^i|\B{x}_k,\B{s}^i)$ is computed using the measurement model (\ref{eq:measurement_model}). Therefore, the mean of the obstacle state $\B{s}^i$ is obtained as
\begin{equation}
\bar{\B{s}}^i = \argmax_{\B{s}^i} \  p(\B{s}^i|\B{x}_k, \B{z}_k^i)
\end{equation}
\noindent with the covariance matrix defined accordingly. Once $n$ measurements are acquired at time $k$, we use it to estimate future obstacle states within in Model Predictive Control (MPC) strategy, where the robot plans for an optimal sequence of controls for $L$ look-ahead steps. At each look-ahead step, the second term in (\ref{eq:obs_update}), that is the obstacle belief has to be updated as per the obstacle motion model which is unknown. Given the state $\B{s}^n$ obtained from the last measurement $\B{z}_k^n$, the new state $\B{s}'$ can then be predicted as 
\begin{equation}
p(\B{s}') = \int_{\B{s}^n} p(\B{s}'|\B{s}^n) \ p(\B{s}^n) 
\label{eq:obs_predict}
\end{equation}
\noindent whose state space form is given by
\begin{equation}
\B{s}(t+1) = A\B{s}(t) + B\B{u}(t) + \nu(t)
\label{eq:state_space}
\end{equation} 
\noindent where $\B{u}(t)$ is the control and $\nu(t)$ is the process noise and $A$, $B$ are matrices which will be defined later. Now we discuss how this prediction can be achieved. From each estimated location $\B{s}^i$ we can then compute the approximate velocities in the $x$ and $y$ directions using the forward difference method. Note that we assume that the obstacle does not change its velocity very drastically and that any two consecutive velocities differ by an $\varepsilon \ll 1 m/s$. Therefore, given $\B{s}^1,\ldots,\B{s}^n$ we obtain the sets 
\begin{multline}
\frac{\mathbf{\Delta x}}{\Delta t} = \left\{\frac{\bar{x}^2 - \bar{x}^1}{\Delta t},\ldots, \frac{\bar{x}^n - \bar{x}^{n-1}}{\Delta t}\right\} =   \left\{\frac{\Delta \bar{x}_1^2}{\Delta t},\ldots,\frac{\Delta \bar{x}_{n-1}^n}{\Delta t}\right\}  \\
 \frac{\mathbf{\Delta y}}{\Delta t} = \left\{\frac{\bar{y}^2 - \bar{y}^1}{\Delta t},\ldots, \frac{\bar{y}^n - \bar{y}^{n-1}}{\Delta t}\right\} = 
 \left\{\frac{\Delta \bar{y}_1^2}{\Delta t},\ldots,\frac{\Delta \bar{y}_{n-1}^n}{\Delta t}\right\}
\label{eq:veloctiy}
\end{multline}
\noindent where $\bar{x}^i, \bar{y}^i$ are the two components of $\bar{\B{s}}^i$ and $\Delta t$ is the time between two measurements. In a similar way we also compute the rate of change of the velocities in the $x$ and $y$ directions. From this computed sets, we choose the maximum change in velocities in both directions and denote the corresponding covariances\footnote{Note that since each variable is Gaussian, their differences are also Gaussian and the corresponding covariances can be computed trivially.} as $\Sigma_v^x$ and $\Sigma_v^y$. From the Taylor series, each component of $\B{s}'$ can be written as
\begin{align}
x'(t') = x^n(t+\Delta t) \approx x^n(t) + \frac{\Delta \bar{x}_{n-1}^n}{\Delta t} \Delta t + \frac{1}{2}\frac{\Delta^2 \bar{x}_{n-1}^n}{\Delta t^2} \Delta t^2 \nonumber \\
y'(t') = y^n(t+\Delta t) \approx y^n(t) + \frac{\Delta \bar{y}_{n-1}^n}{\Delta t} \Delta t + \frac{1}{2}\frac{\Delta^2 \bar{y}_{n-1}^n}{\Delta t^2} \Delta t^2 
\label{eq:obs_mean}
\end{align}
\noindent Note that the above equation is in the form of (\ref{eq:state_space}). The process noise is hence defined as 
\begin{equation}
\nu(t) \sim \mathcal{N}\left(0,\begin{bmatrix}
\frac{1}{4}\Sigma_v^x (\Delta t)^4  &  0 \\
0 & \frac{1}{4}\Sigma_v^y (\Delta t)^4
\end{bmatrix} \right)
\label{eq:obs_cov}
\end{equation}
We use a 2D laser scanner to estimate the state of dynamic obstacles. It is assumed that the geometry of the obstacle is spherical and is known beforehand. From each scan of the laser rangefinder, the ray with the minimum distance $r_j$ and the corresponding orientation is computed to form a measurement $\B{z}_k^i$. This is repeated to obtain $n$ distinct measurements. Given these measurements and the current robot state estimated using the standard EKF, the $x$ and $y$ components of the obstacle location are estimated. These estimated values are then used to compute the respective velocities using (\ref{eq:veloctiy}). The location estimates of the last scan $\B{z}_k^n$ is then used as the prior in (\ref{eq:obs_predict}) to estimate future obstacle states. The respective mean and covariance are computed using (\ref{eq:obs_mean}) and (\ref{eq:obs_cov}). To illustrate our approach, in Fig.~\ref{fig:obs_predict} we plot the true and estimated locations for different obstacle trajectories. 

The approach is readily extended to estimate the state of all obstacles detected by the laser scanner. We note here that advanced strategies exits in the literature to efficiently segment laser rangefinder's scans, but it is not the main focus of the current paper. We therefore employ a rather simpler method sufficient to demonstrate the approach discussed herein. The laser rangefinder returns a sequence of distance measurements and these distances are less than the maximum range when obstacles are encountered. We assume that the obstacles are not too close, that is, there is a least one distance measurement between two obstacles that gives the maximum range. This discontinuity in the distance measurements between two obstacles allows us to separate the laser scanner measurements into different clusters belonging to different obstacles. From each cluster, we estimate the state of the corresponding obstacle. Note that it does not guarantee estimating the state of all the obstacles since some of them could be completely occluded by the others. It is also worth mentioning that estimating the location of static obstacles is a special case of the approach discussed here since for static obstacles both $\frac{\mathbf{\Delta x}}{\Delta t}$ and $\frac{\mathbf{\Delta y}}{\Delta t}$ equate to zero. 
\section{Objective Function} 
\label{sec:obj_function}
At each time instant $k$, the robot plans for $L$ look-ahead steps to obtain a control policy $\B{u}^\star_{k:k+L-1}$ given by
\begin{equation}
\B{u}^\star_{k:k+L-1} = \argmin_{\B{u}_{k:k+L-1}} J_k(\B{u}_{k:k+L-1})
\end{equation}

\noindent where $J_k(\B{u}_{k:k+L-1})$ is the objective function. As per the standard MPC, at each time step the first control command $\B{u}^\star_k$ is then applied. At each time step, the robot is required to minimize its control usage and proceed towards the goal $\B{x}^g$ avoiding collisions, while minimizing its state uncertainty. We quantify the state uncertainty by computing the trace of the marginal covariance of the robot position. As a result, we have the following objective function
\begin{multline}
J_k(\B{u}_{k:k+L-1}) \doteq  \sum_{l=0}^{L-1} 
 \norm{\xi(\B{u}_{k+l})}^2_{M_u} \! \! \! + \!  tr\left(\norm{M_{\Sigma}}^2_{\Sigma_{k+l}}\right)\!+ \! M_CP(\mathcal{C}_{\B{x}_{k+l},\B{s}_{k+l}})\\ +\underset{\B{z}_{k+L}}{ \EX} \left[\norm{\B{x}_{k+L} - \B{x}^g}^2_{M_g} + tr\left(\norm{M_{\Sigma}}^2_{\Sigma_{k+L}}\right)  \right]
 \label{eq:objective_fn}
\end{multline}  

\noindent where $\norm{x}_S = \sqrt{x^TSx}$ is the Mahalanobis norm, $M_u, M_g, M_C$ are weight matrices and $\xi(\cdot)$ is a function that quantifies control usage. The choice of weight matrices and the control function vary with the application. The term $tr\left(\norm{M_{\Sigma}}^2_{\Sigma_k}\right) = tr\left(M_{\Sigma}^T\Sigma_kM_{\Sigma}\right)$ returns the marginal covariance of the robot location. Therefore, $M_{\Sigma} = \tau \bar{M}_{\Sigma}$, where $\tau$ is a positive scalar and $\bar{M}_{\Sigma}$ is a matrix filled with zero or identity entries. $M_C$ penalizes the belief states with higher collision probabilities. Since future observations are not available at planning time and are stochastic, the expectation is taken to account for all possible future observations. 

Our approach is summarized in Algorithm~\ref{algo}. At each time instant, the robot state is estimated using EKF (lines~\ref{algo:1},~\ref{algo:2}). As described in the previous Section, obstacles are detected using a laser rangefinder. For the $j-$th detected obstacle, its future state is then estimated (line~\ref{algo:3}) using the approach discussed in Section~\ref{sec:obs}. The total collision cost is then computed by adding the collision cost with each obstacle (line~\ref{algo:4}). Please note that if no $\epsilon-$safe configuration exists then the algorithm terminates. Finally the total cost is computed as given in (\ref{eq:objective_fn}). This is repeated for each horizon step to obtain the optimal control policy $\B{u}^\star_{k:k+L-1}$. The control command $\B{u}^\star_k$ is then applied and the process is repeated till the goal is reached. 

\begin{algorithm}[t]
\SetAlgoLined
\Input{$b[\B{x}_k], L, \epsilon, N,$ radii, $\B{z}_k^1,\ldots,\B{z}_k^n$}
$J_k = 0$, $l = 0$ \\
compute $p(\B{s}^i|\B{x}_{k+l}, \B{z}_{k+l}^i)$ $\forall i, 1 \leq i \leq n$ and $\frac{\mathbf{\Delta x}}{\Delta t}, \frac{\mathbf{\Delta y}}{\Delta t}$ \\
 \While{true}{
 $ b[\B{x}_{k+l+1}^-] \leftarrow b[\B{x}_{k+l}]p(\B{x}_{k+1+l}|\B{x}_{k+l},\B{u}_{k+l})$\label{algo:1} \\
 $ \{\B{z}_{k+l+l}\} \leftarrow$ simulate future observations \\
 \For{each  $ \{\B{z}_{k+l+l}\}$}{
    compute $b[\B{x}_{k+l+1}]$\label{algo:2} \\
    predict obstacle state $\B{s}_{k+l}$ (\,using (\ref{eq:obs_mean})\,)\label{algo:3}\\
    compute $\sum_j P(\mathcal{C}_{\B{x}_{k+l+1},\B{s}_{k+l}})$ \\
    compute total\_cost (\,using (\ref{eq:objective_fn})\,)\label{algo:4}\\
    }
$J_k \leftarrow J_k +$ total\_cost \\
 }
 $\B{u}^\star_{k:k+L-1} \leftarrow \argmin_{\B{u}_{k:k+L-1}} J_k(\B{u}_{k:k+L-1})$\\
\Return $\B{u}^\star_{k:k+L-1}$
 \caption{Safe motion planning.}
 \label{algo}
\end{algorithm}

\section{Experiments}
\label{experiments}
In this Section we describe our implementation and then illustrate and explore the capabilities of our proposed approach. First, we present a theoretical example to conceptually understand the proposed approach. Next, we consider both single and multi-robot experiments, which are performed using different Gazebo-based realistic simulations. For all the experiments we use a TurtleBot3 Waffle robot with a radius of 0.22$m$. The robot is equipped with a Laser Distance Sensor LDS-01 and we use the same to acquire obstacle range and bearing. The performance is evaluated on an Intel{\small\textregistered} Core i7-6500U CPU$@$2.50GHz$\times$4 with 8GB RAM under Ubuntu 16.04 LTS. In all the Gazebo based experiments, the initial uncertainty in robot pose is $\Sigma_0 = diag(0.1 m, 0.1 m,0.02 rad)$. The LDS detections/measurements are only from the obstacles whose motion is unknown and the EKF is employed to predict the robot state at each time step. The ground truth odometry from Gazebo is used to measure the pose of the robot, mimicking a motion capture system. This measurement is then corrupted with noise to perform state estimation. However, this estimation is not performed at each time step and we randomly select the times steps to carry out the same. In this way we explore the robustness of our approach to localization uncertainties. 
\revtext{
\begin{remark}
We note here that comparison to other approaches have been provided in Section~\ref{sec:comparison} and the computation of collision probability with these approaches have been reproduced to the best our understanding. While this may be accurate for static scenarios as demonstrated in~\ref{sec:comparison}, we believe that extending this comparison to online planning scenarios would not be an accurate portrayal of these works. For example, the work in~\cite{park2018IEEE} finds the position with the maximum probability by formulating it as an optimization problem.~\cite{zhu2019RAL} require linearizing the collision constraint and computation of the inverse of the standard error function. There are a number of ways to perform the optimization, the linearization or the computation of the error functions. So unless we know the exact methods used by these approaches, extending the comparisons to online planning would lead to an inaccurate depiction of these approaches. We thus limit the comparison to these approaches to static scenarios presented in Section~\ref{sec:comparison}. The approaches in~\cite{dutoit2011IEEE,park2018IEEE,zhu2019RAL} compute upper bounds for collision probability and if these methods are employed, we expect them to produce longer paths than the ones depicted in this section. 
\end{remark}}

\subsection{Theoretical Example}
\begin{figure}[ht]
\centering
  \subfloat[]{\includegraphics[trim=3.5cm 0cm 3.5cm 0cm, clip=true,scale=0.55]{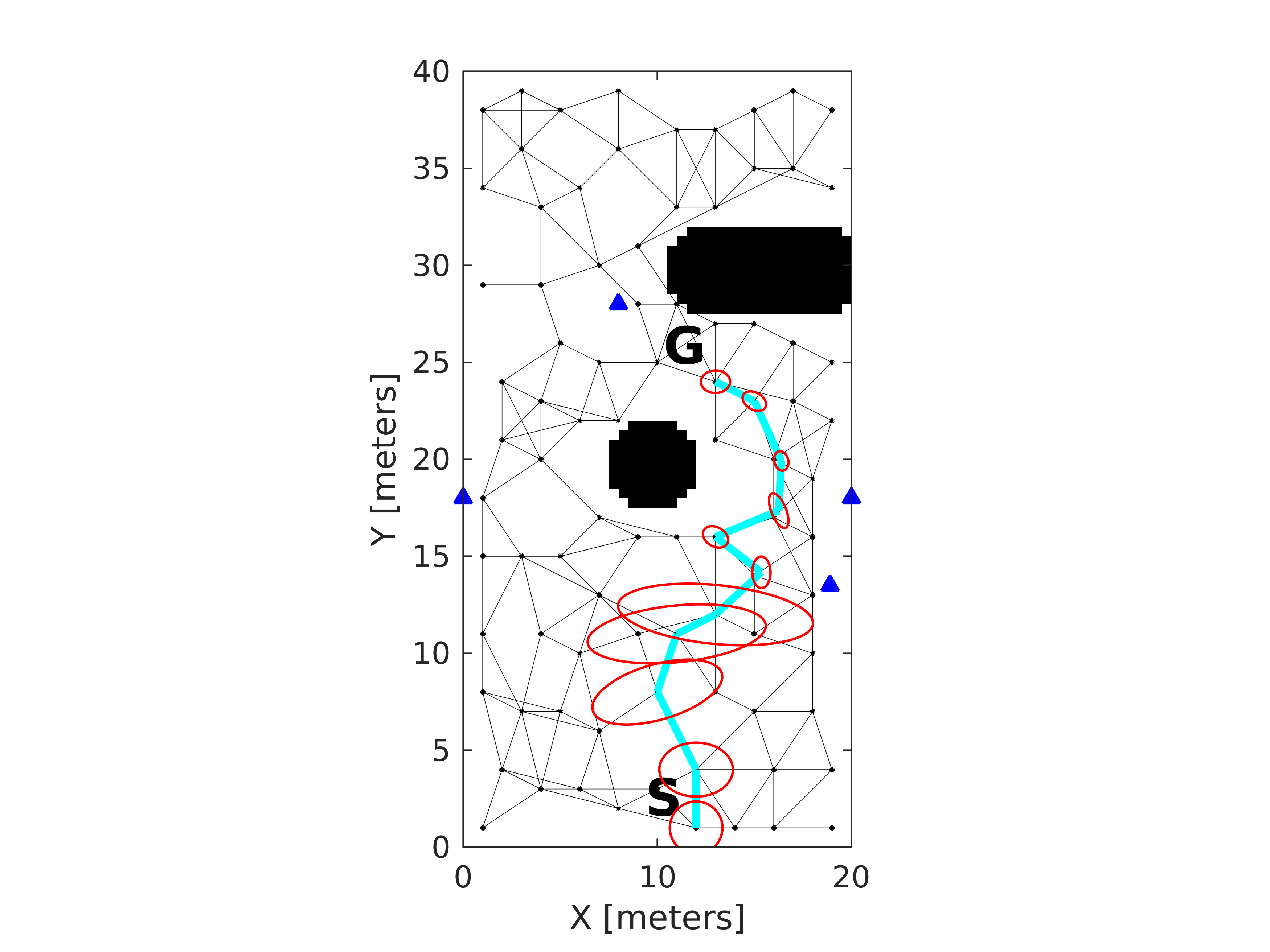}} 
  \subfloat[]{\includegraphics[trim=3.5cm 0cm 3.5cm 0cm, clip=true,scale=0.55]{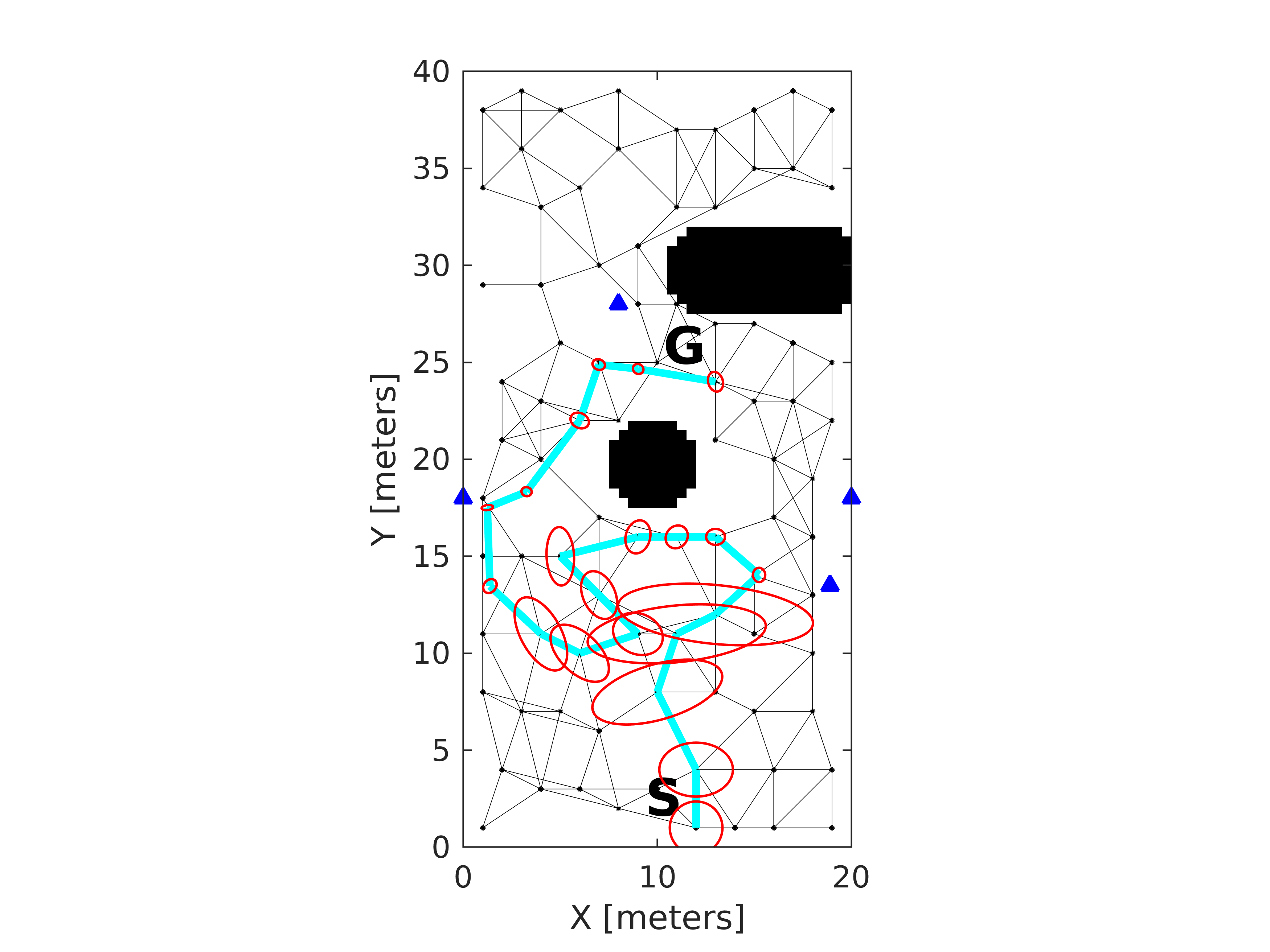}}
    \subfloat[]{\includegraphics[trim=3.5cm 0cm 3.5cm 0cm, clip=true,scale=0.55]{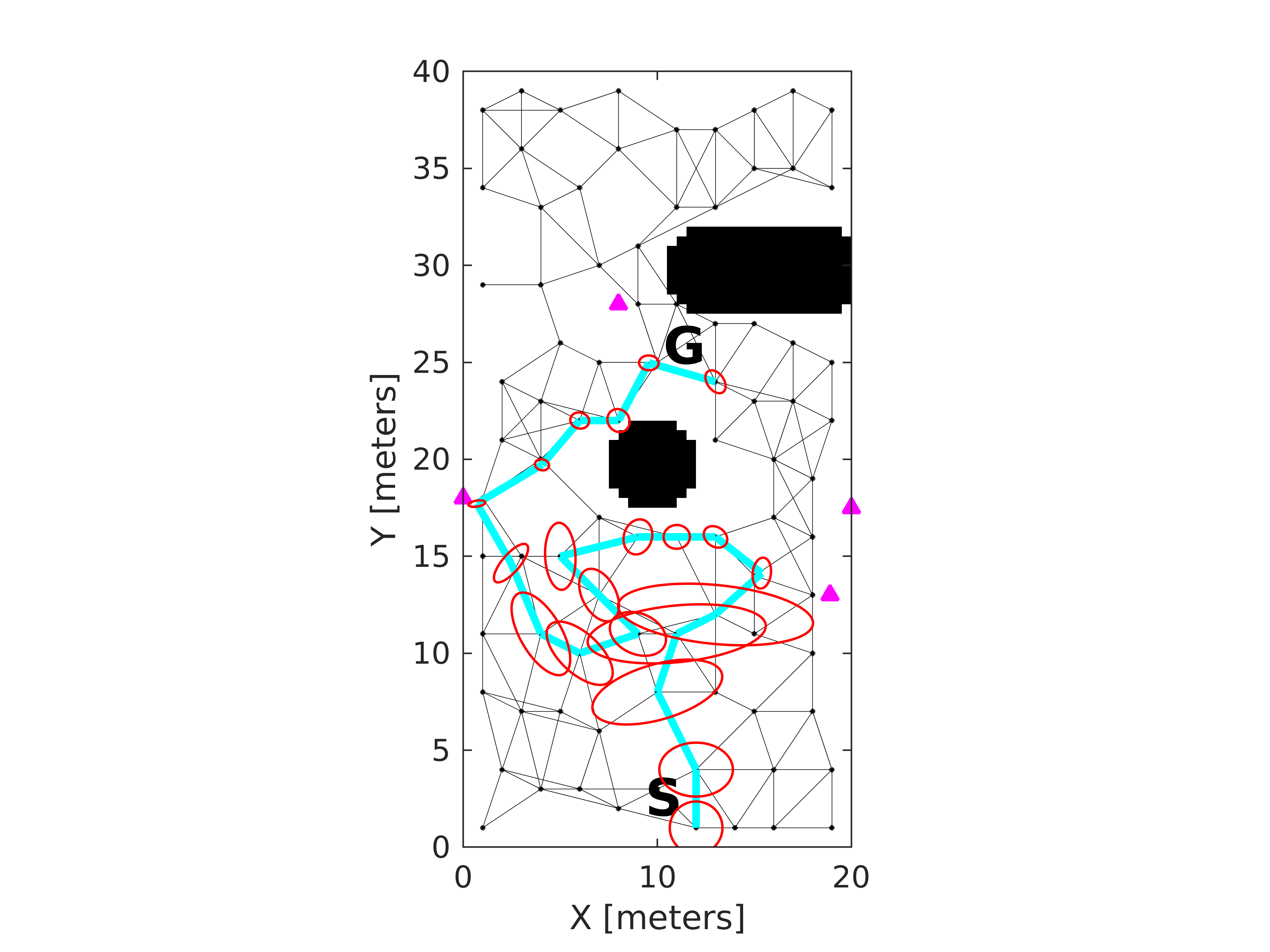}}
  \caption{Trajectory and the covariance evolution for single planning for the 2D environment. (a) Plan obtained when object uncertainty is not considered. (b) The planned trajectory when object uncertainty is considered (c) Planned trajectory with true landmark locations.}
  \label{fig:collision_2Drobot}
   \end{figure}
\begin{figure}[h]
\centering
  \subfloat[]{\includegraphics[trim=3cm 0cm 3cm 0cm, clip=true,scale=0.55]{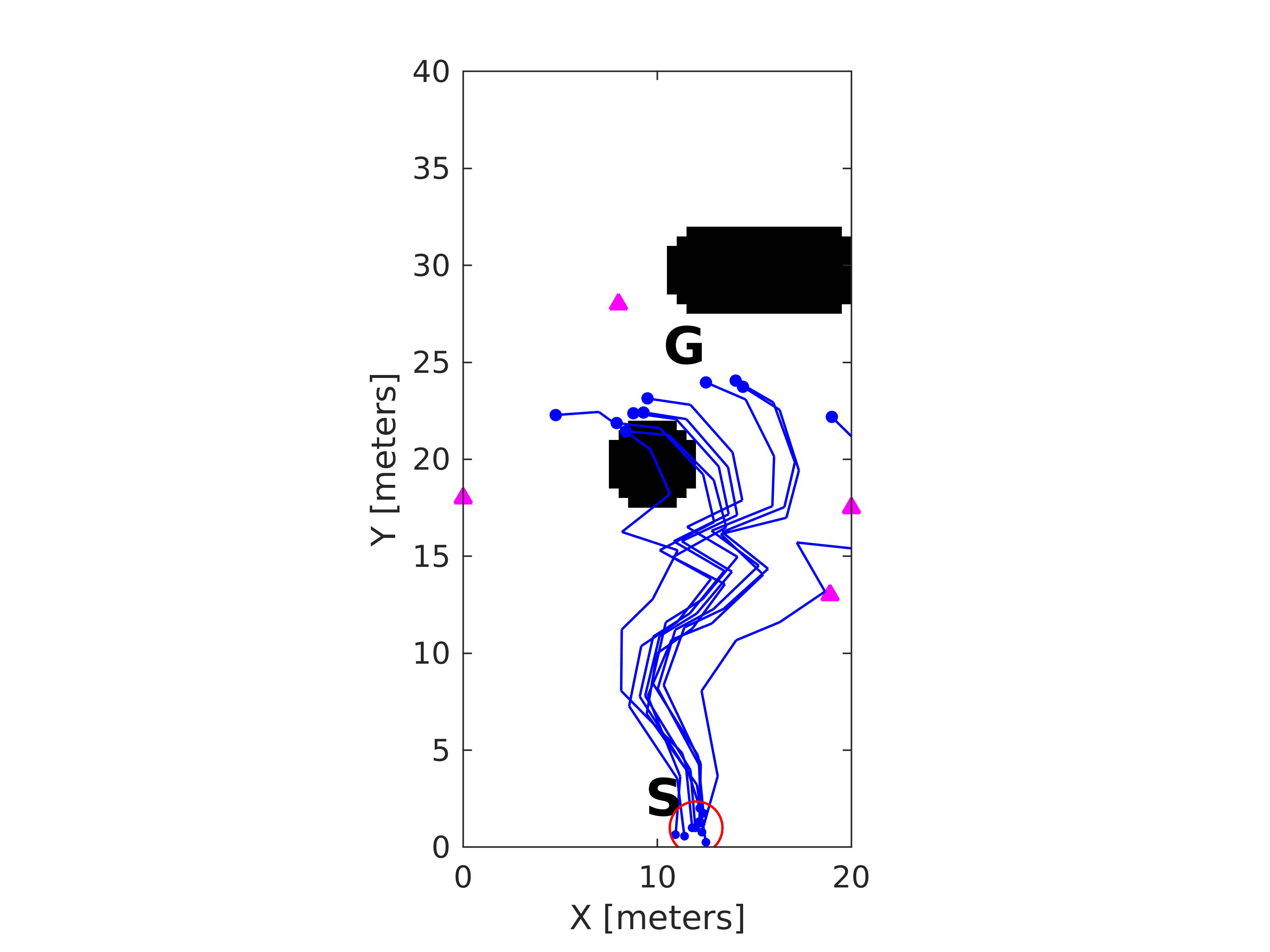}} 
  \subfloat[]{\includegraphics[trim=3cm 0cm 3cm 0cm, clip=true,scale=0.55]{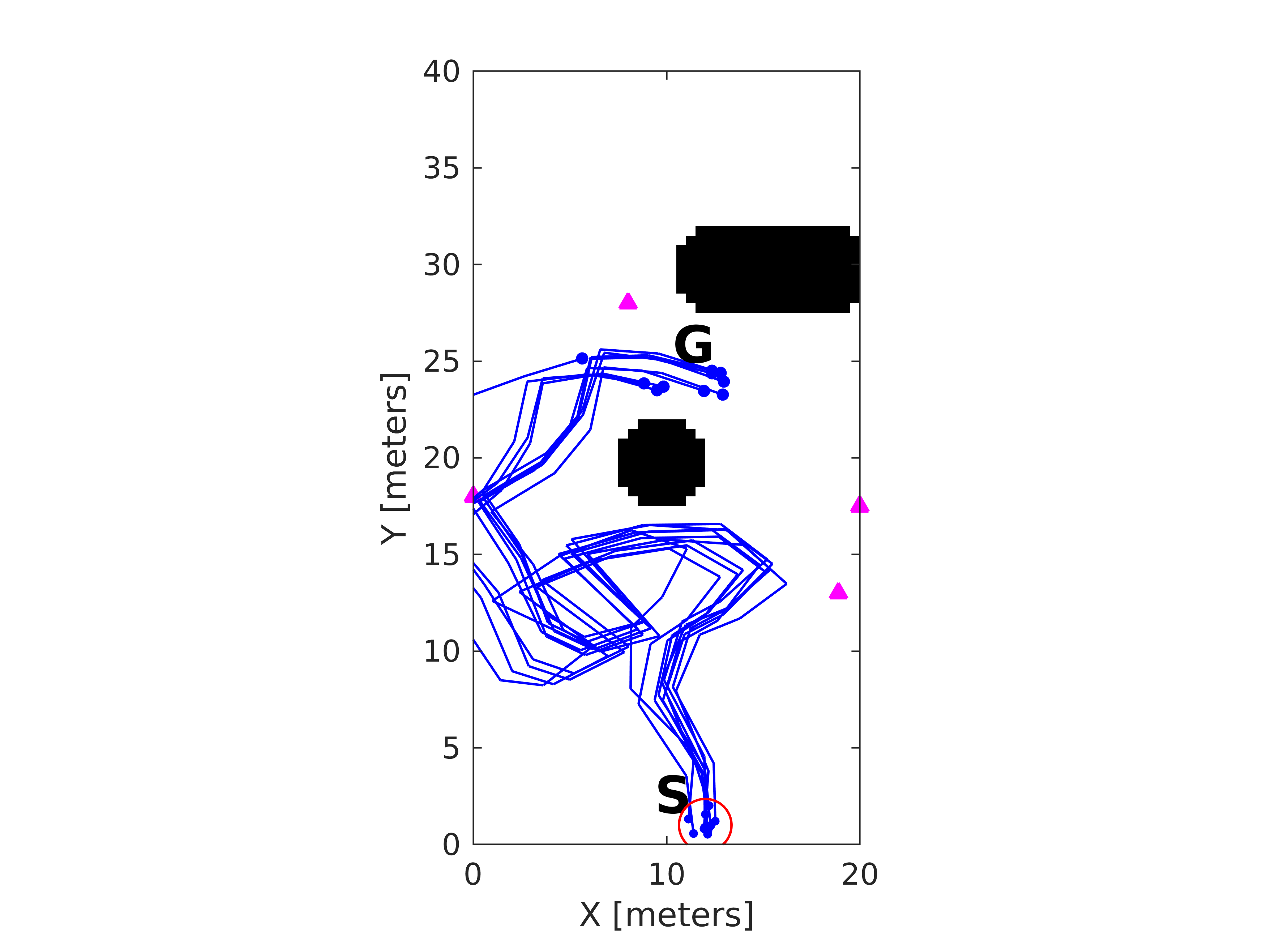}}
      \caption{Execution traces of robot's true state across ten simulation with initial state drawn from the known initial belief. (a) Computed control when object uncertainty not considered is followed. (b) Traces of robot's true state while following the computed control considering object uncertainty.}
  \label{fig:execution_sim}
   \end{figure}
   \begin{figure}[]
\centering
 \subfloat[Scenario A]{\includegraphics[width=0.45\textwidth]{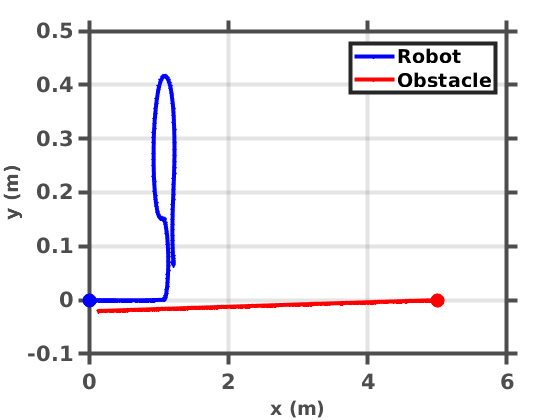}}
  \subfloat[Scenario B]{\includegraphics[width=0.45\textwidth]{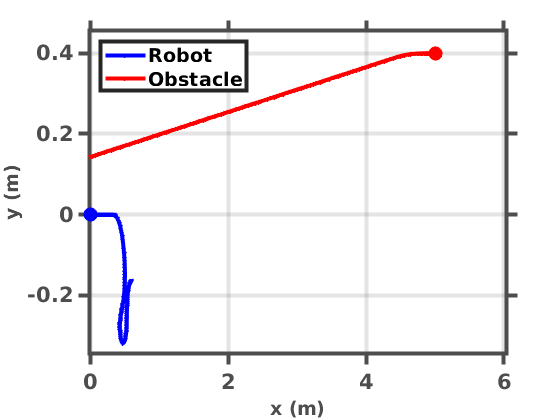}}  \\  
  \subfloat[Scenario C]{\includegraphics[width=0.44\textwidth]{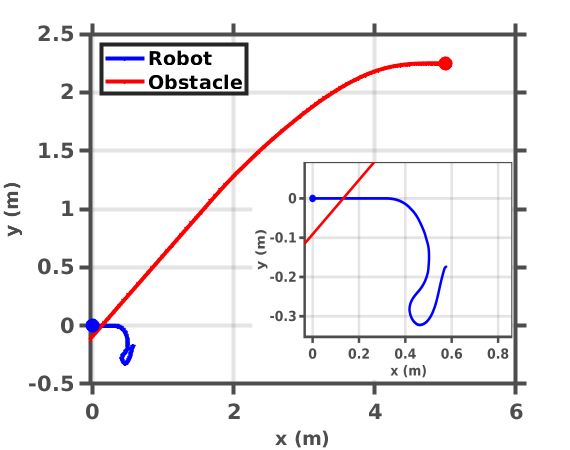}} 
 \subfloat[Scenario D]{\includegraphics[width=0.45\textwidth]{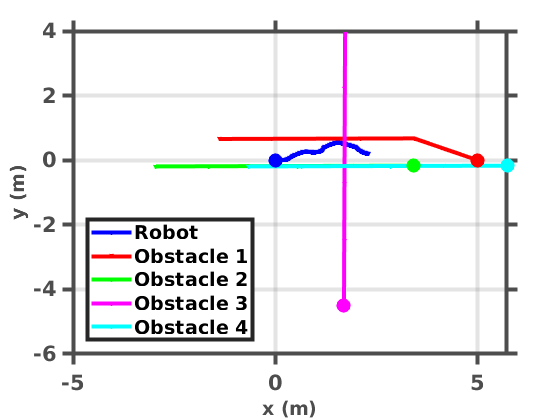}}
       \caption{Top view of robot and obstacle trajectories are plotted with the starting locations marked as round blobs. The robot trajectory is shown in blue. (a) Single obstacle with velocity of 0.5$m/s$. (b) Obstacle velocity is 1.0$m/s$. (c) Obstacle velocity is 2.5$m/s$ and the zoomed figure is shown in the inset. (d) Four obstacles with different velocities.}
  \label{fig:trajectory}
\end{figure}
\begin{figure}[]
\centering
 \subfloat{\includegraphics[trim={1cm 1cm 1cm 1cm},clip,width=0.25\textwidth]{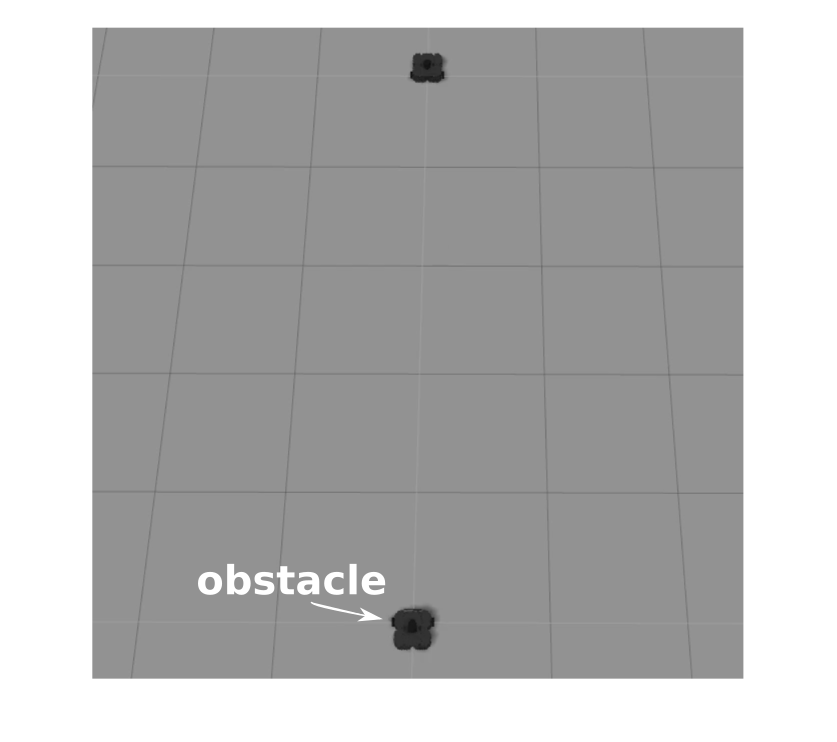}}
  \subfloat{\includegraphics[trim={1cm 1cm 1cm 1cm},clip,width=0.25\textwidth]{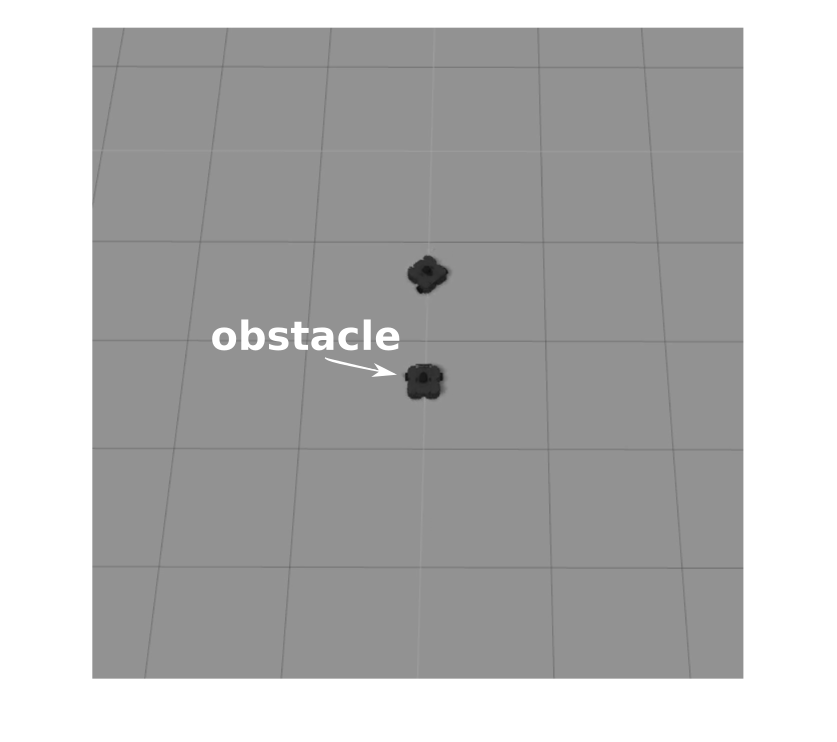}}    
  \subfloat{\includegraphics[trim={1cm 1cm 1cm 1cm},clip,width=0.25\textwidth]{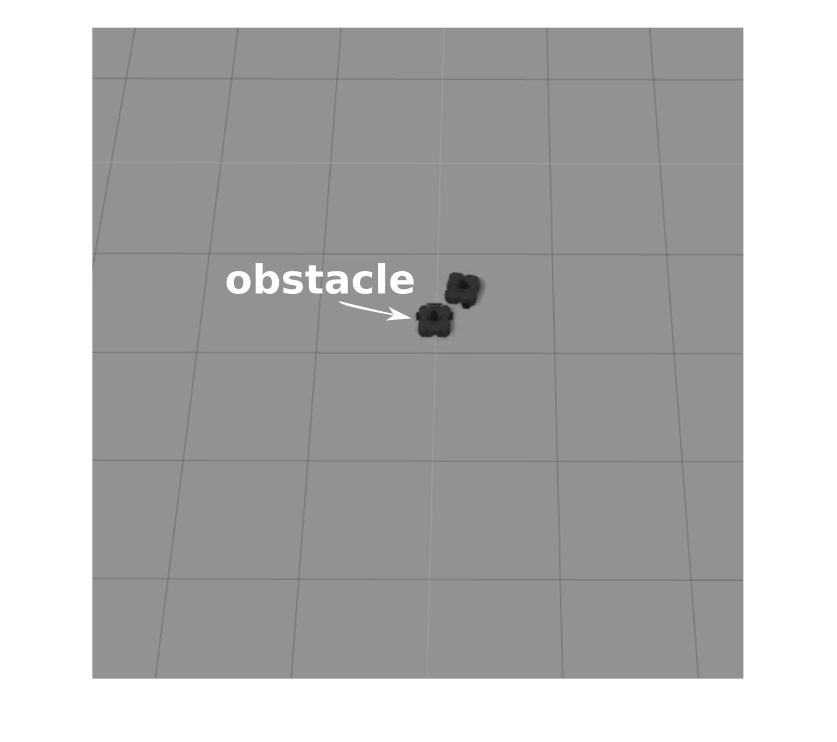}} 
 \subfloat{\includegraphics[trim={1cm 1cm 1cm 1cm},clip,width=0.25\textwidth]{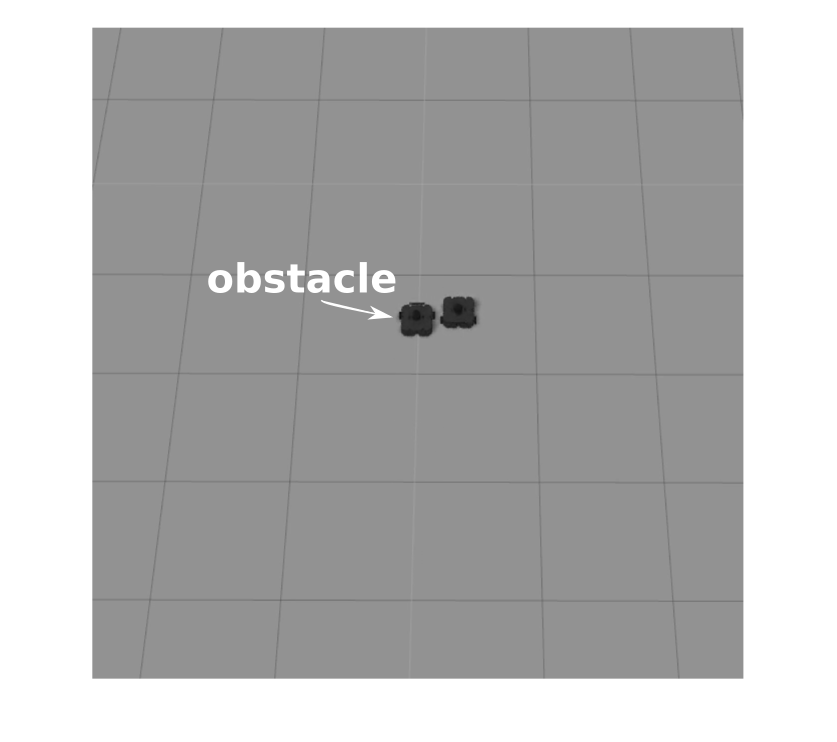}}
 
 \subfloat{\includegraphics[trim={1cm 1cm 1cm 1cm},clip,width=0.25\textwidth]{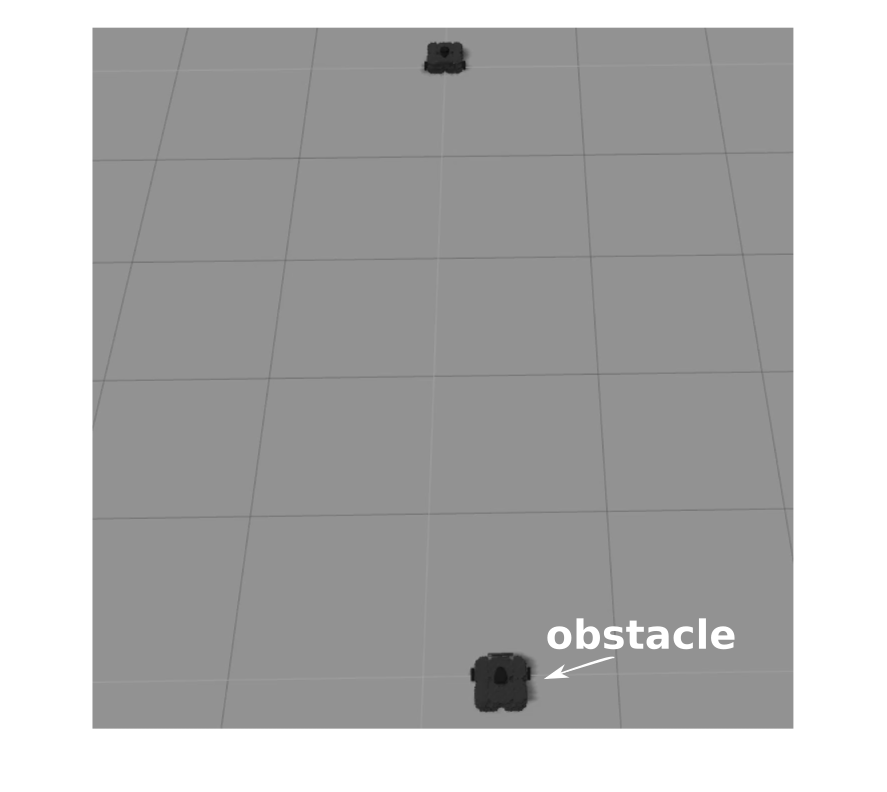}}
  \subfloat{\includegraphics[trim={1cm 1cm 1cm 1cm},clip,width=0.25\textwidth]{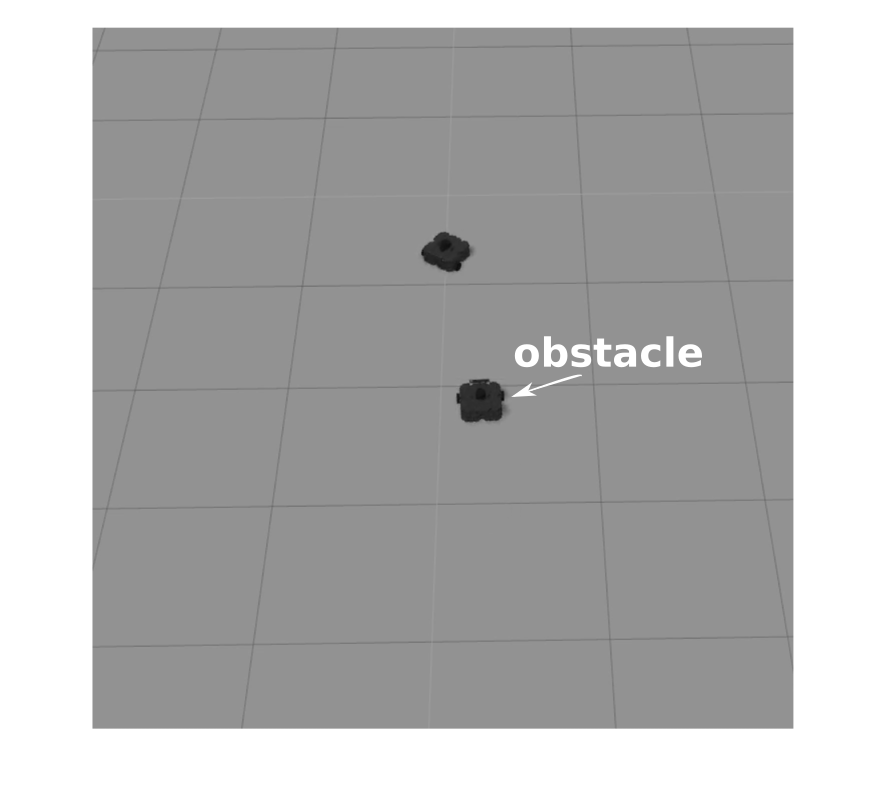}}    
  \subfloat{\includegraphics[trim={1cm 1cm 1cm 1cm},clip,width=0.25\textwidth]{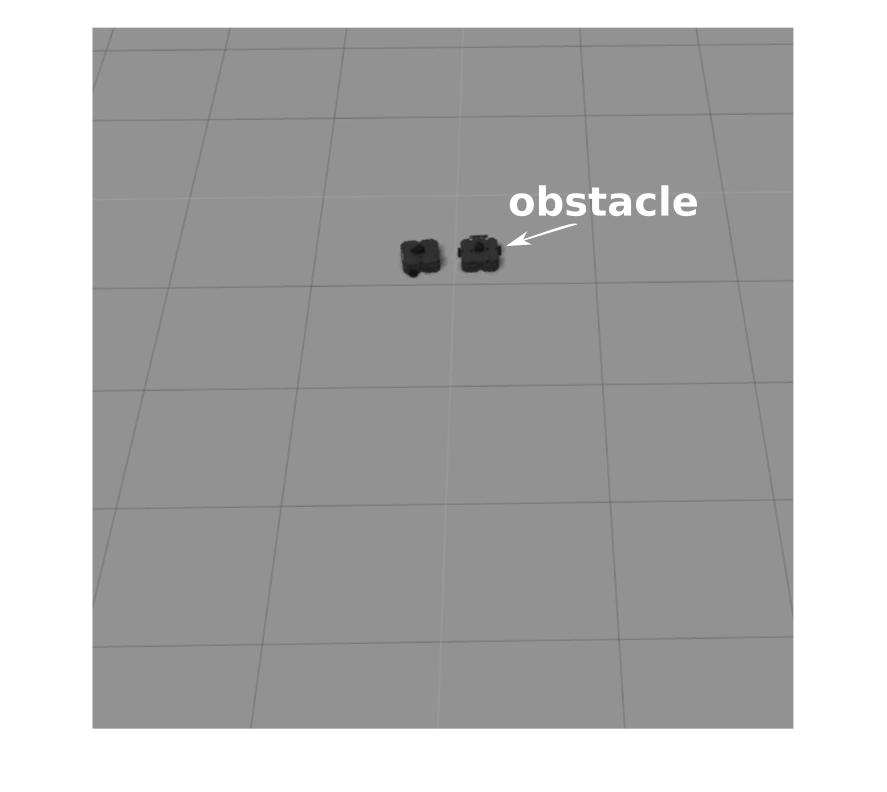}} 
 \subfloat{\includegraphics[trim={1cm 1cm 1cm 1cm},clip,width=0.25\textwidth]{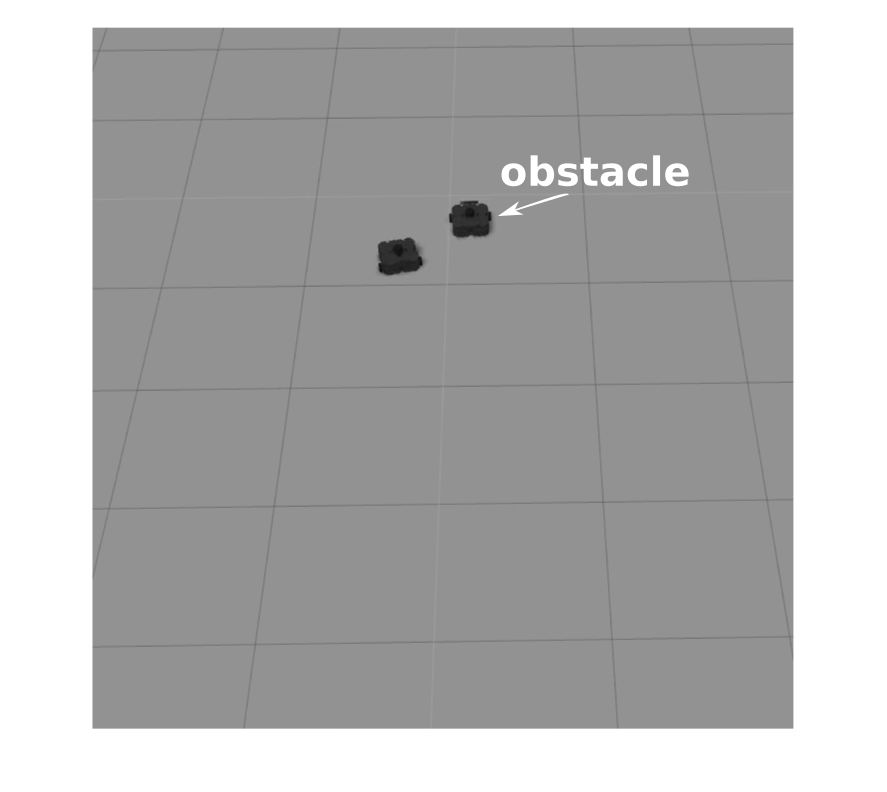}}
 
 \clearsubcaptcounter
 \subfloat[Stage 1]{\includegraphics[trim={1cm 1cm 1cm 1cm},clip,width=0.25\textwidth]{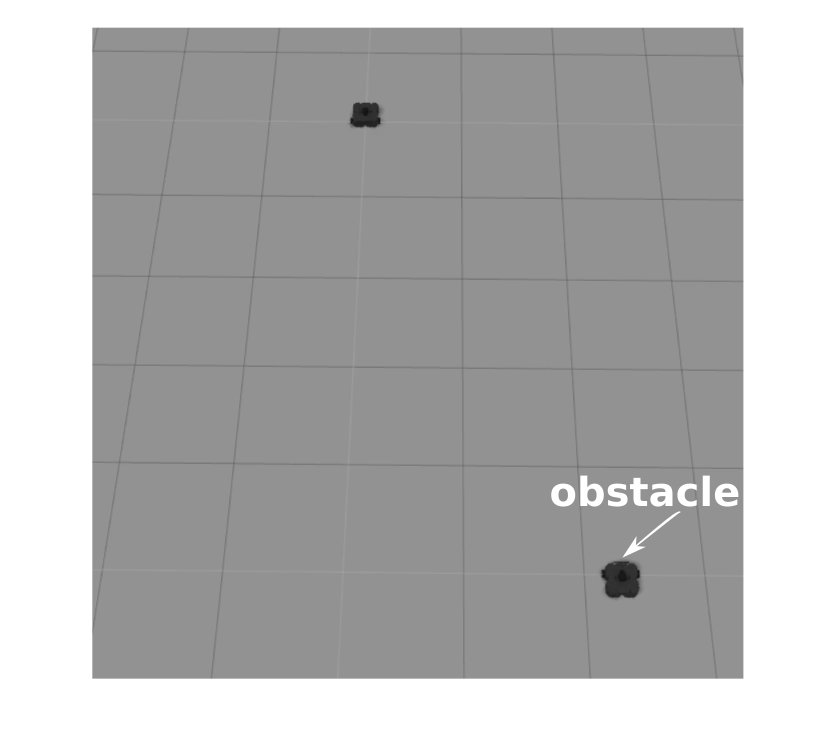}}
  \subfloat[Stage 2]{\includegraphics[trim={1cm 1cm 1cm 1cm},clip,width=0.25\textwidth]{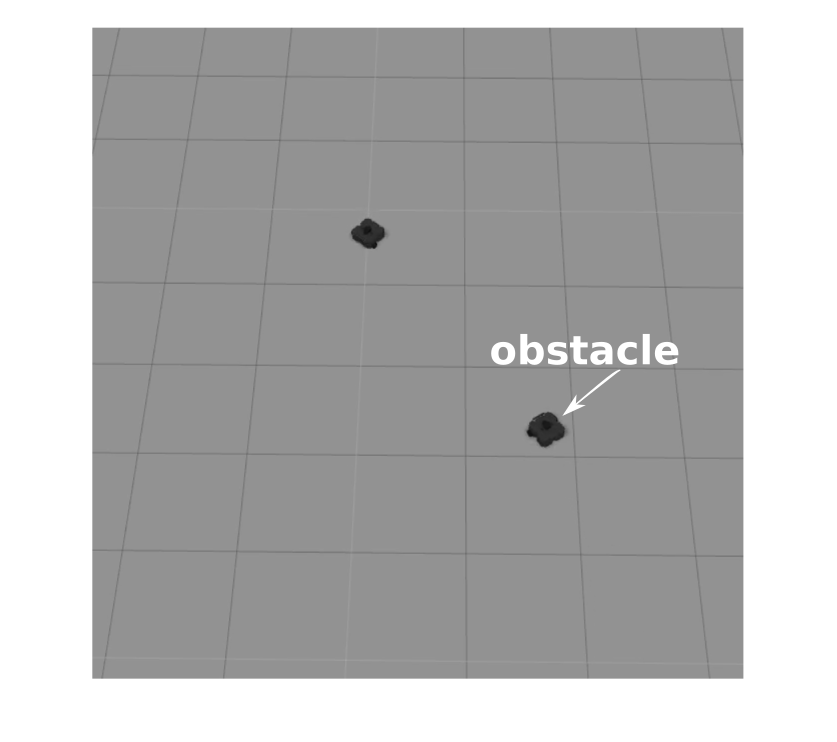}}    
  \subfloat[Stage 3]{\includegraphics[trim={1cm 1cm 1cm 1cm},clip,width=0.25\textwidth]{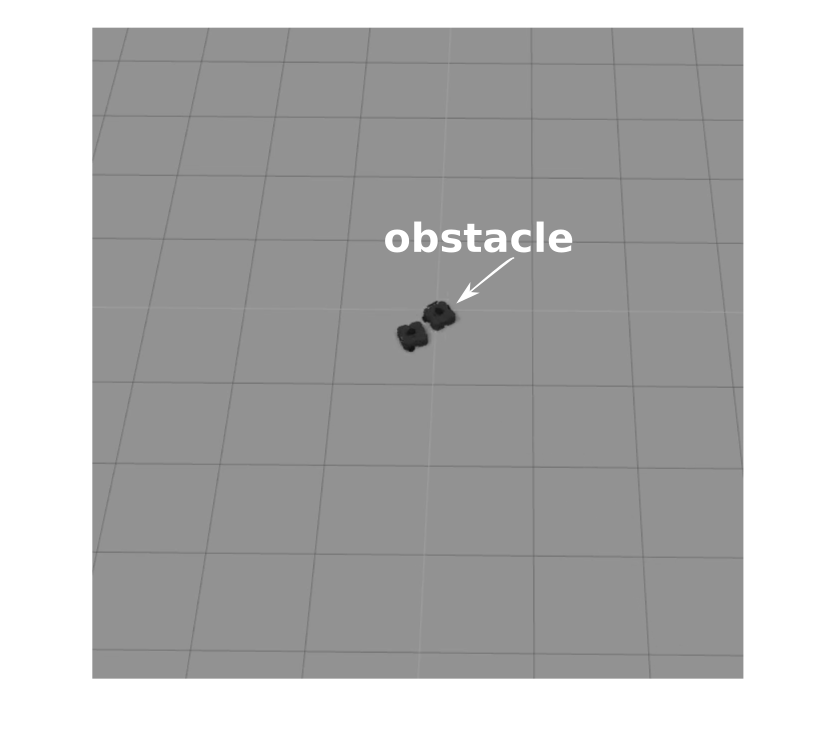}} 
 \subfloat[Stage 4]{\includegraphics[trim={1cm 1cm 1cm 1cm},clip,width=0.25\textwidth]{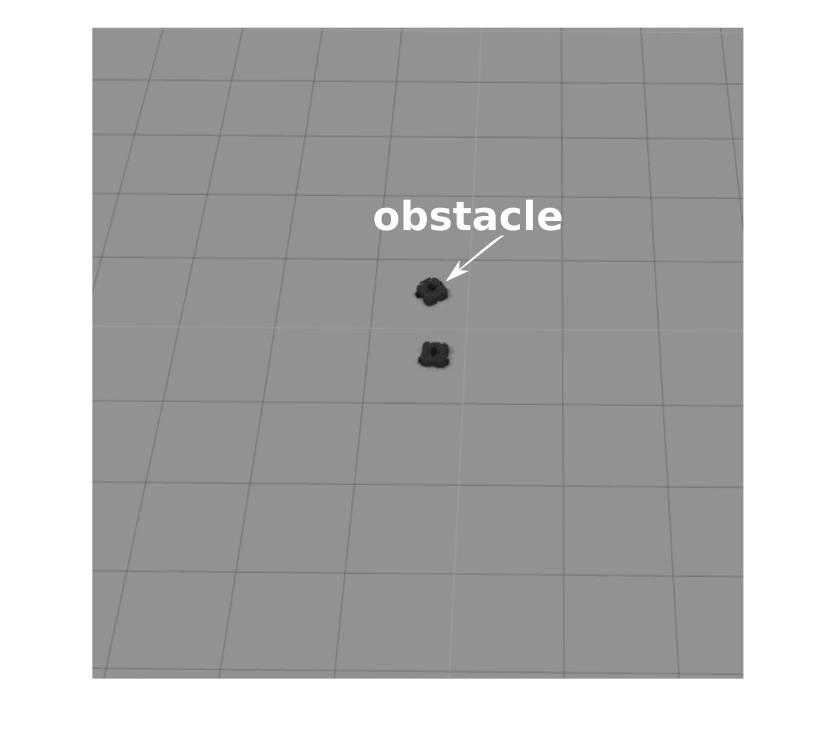}}
        \caption{Top view snapshots of the robot and the obstacle at four different stages (from left to right) of the experiment in scenarios $A$ (row 1), $B$ (row 2) and $C$ (row 3)~\revtext{shown in Fig.~\ref{fig:trajectory}}. Positive x-axis is vertically downwards.}
  \label{fig:snapshot}
\end{figure}

\begin{figure}[]
\centering
 \subfloat[Stage 1]{\includegraphics[trim={1cm 1cm 1cm 1cm},clip,width=0.25\textwidth]{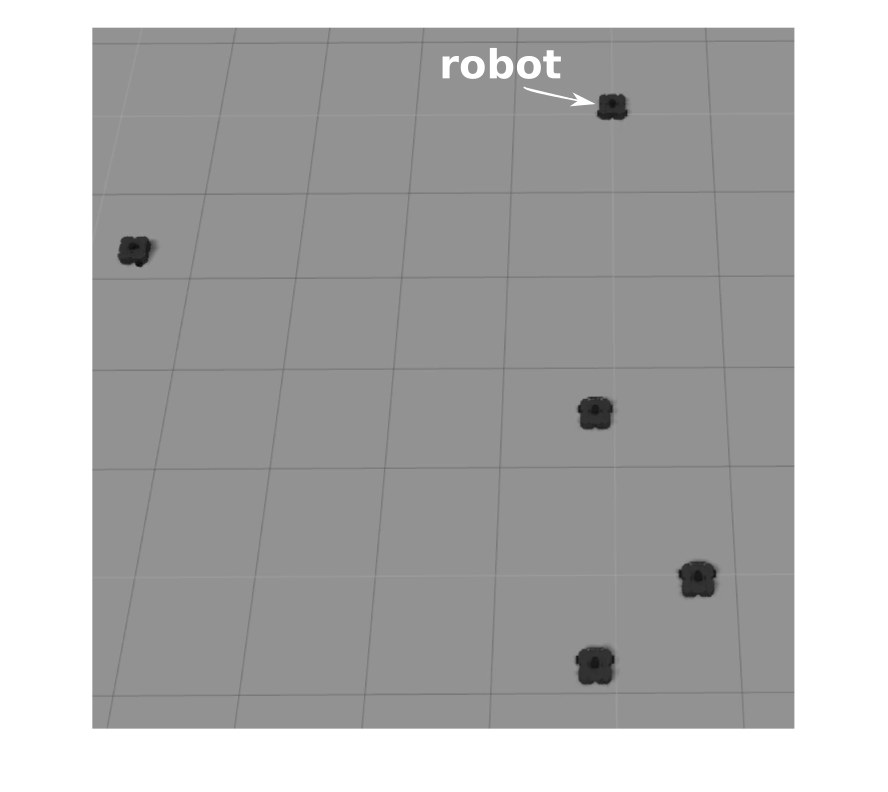}}
  \subfloat[Stage 2]{\includegraphics[trim={1cm 1cm 1cm 1cm},clip,width=0.25\textwidth]{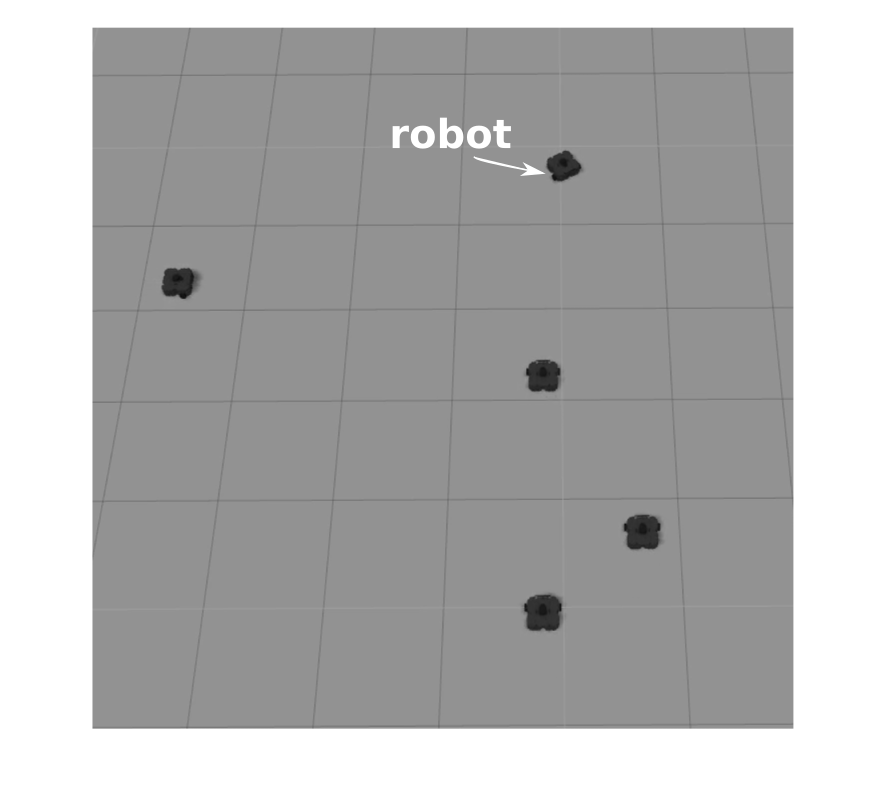}}    
  \subfloat[Stage 3]{\includegraphics[trim={1cm 1cm 1cm 1cm},clip,width=0.25\textwidth]{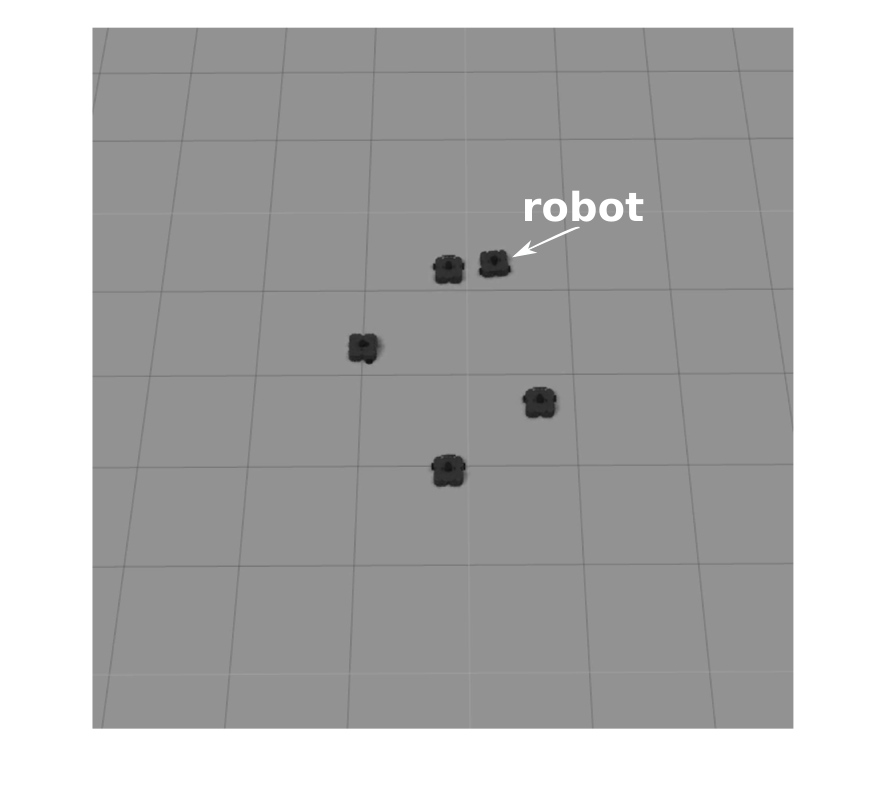}} 
 \subfloat[Stage 4]{\includegraphics[trim={1cm 1cm 1cm 1cm},clip,width=0.25\textwidth]{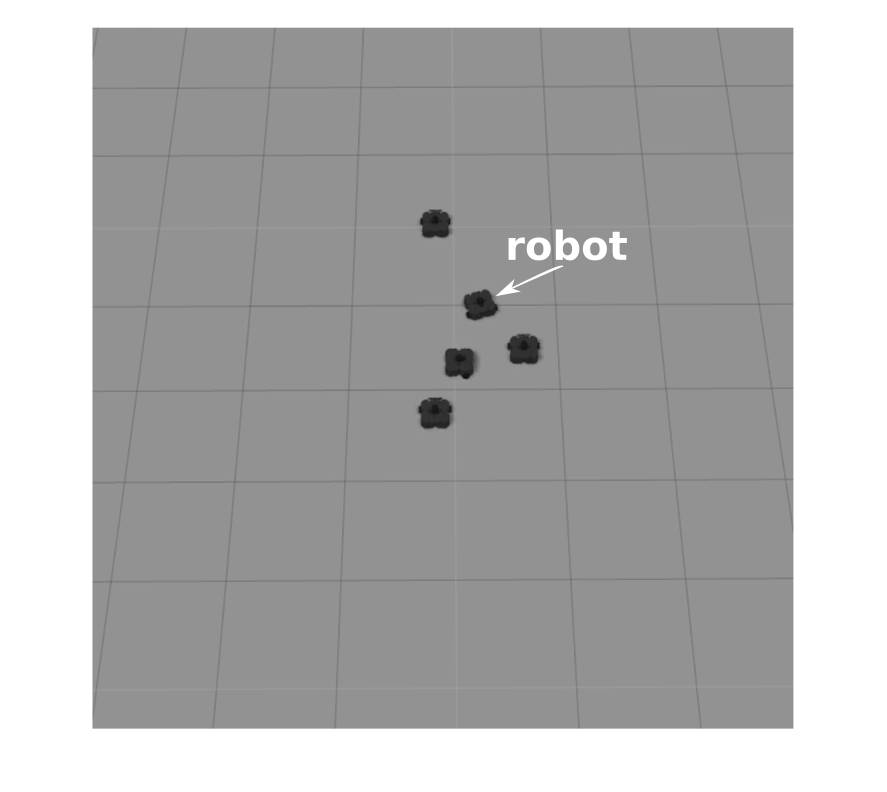}}
 
  \subfloat[Stage 5]{\includegraphics[trim={1cm 1cm 1cm 1cm},clip,width=0.25\textwidth]{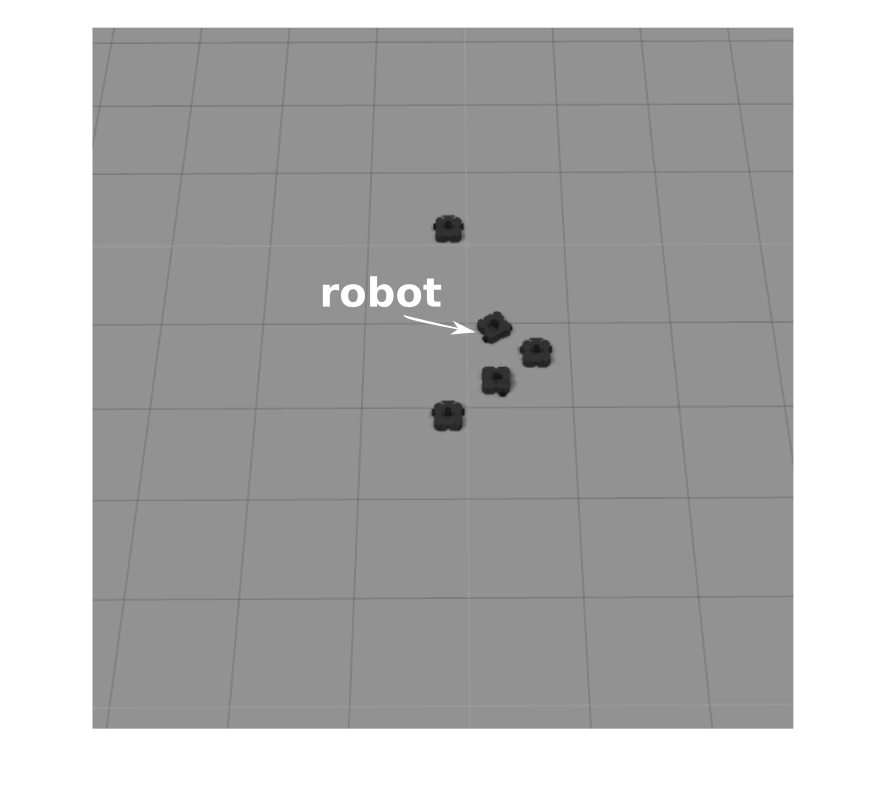}}
  \subfloat[Stage 6]{\includegraphics[trim={1cm 1cm 1cm 1cm},clip,width=0.25\textwidth]{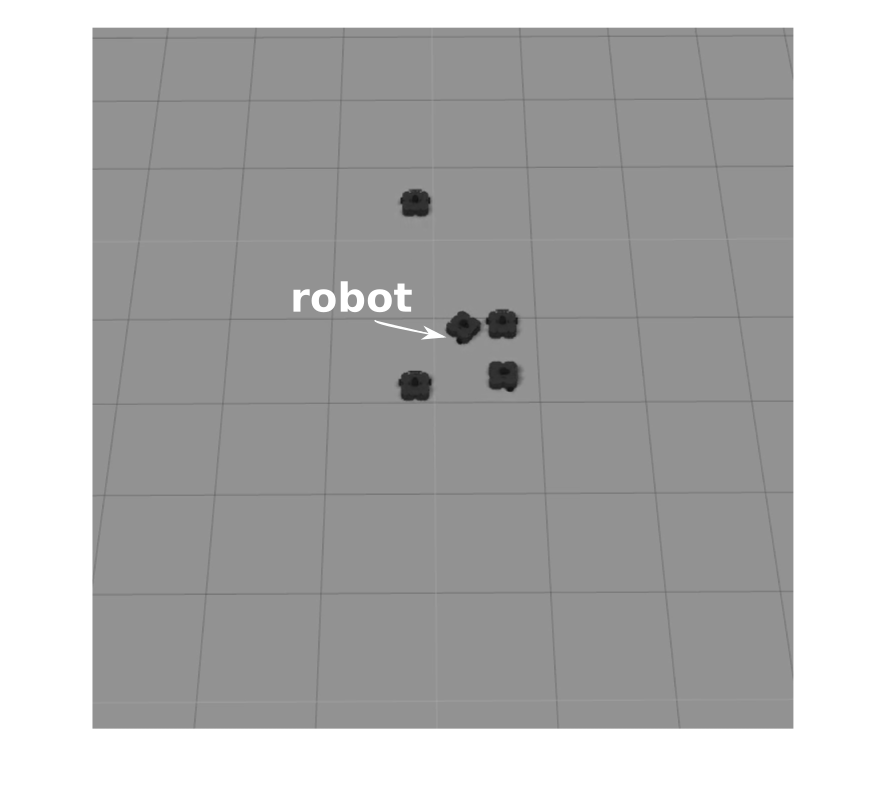}}    
  \subfloat[Stage 7]{\includegraphics[trim={1cm 1cm 1cm 1cm},clip,width=0.25\textwidth]{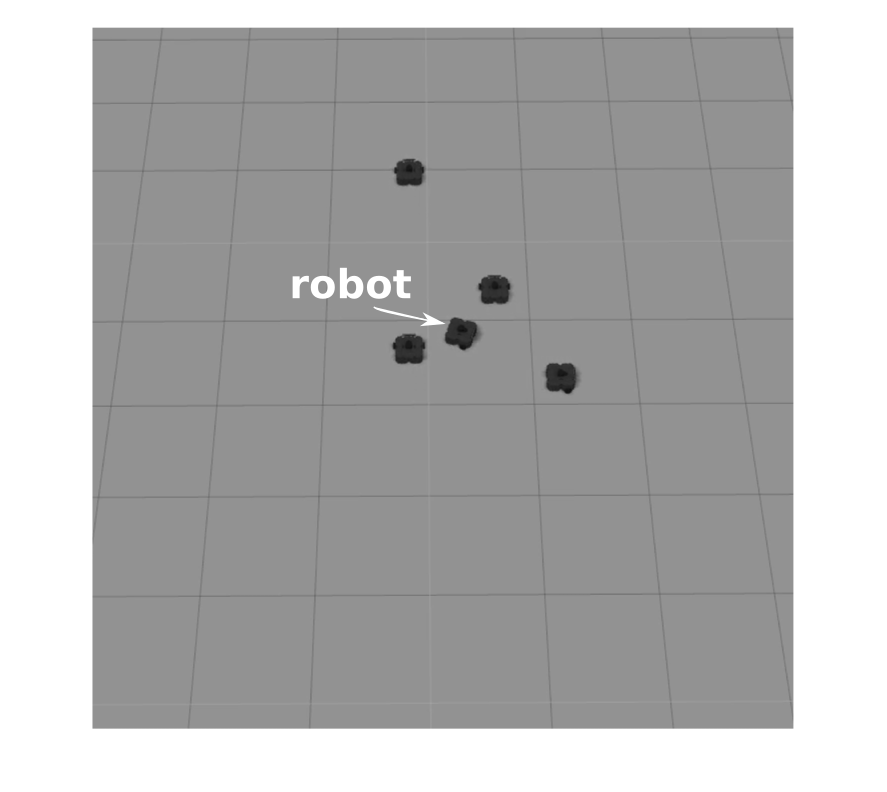}} 
 \subfloat[Stage 8]{\includegraphics[trim={1cm 1cm 1cm 1cm},clip,width=0.25\textwidth]{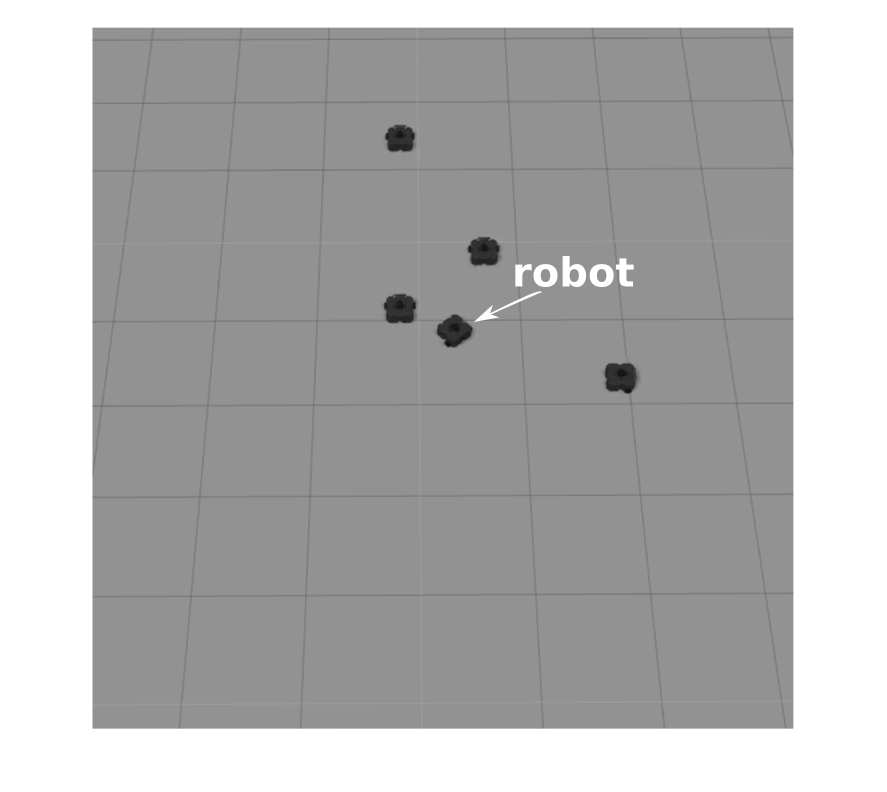}}
        \caption{Top view snapshots of the robot and the obstacles in the Gazebo environment at different stages of the experiment in scenario $D$~\revtext{shown in Fig.~\ref{fig:trajectory}}.}
  \label{fig:snapshot_multi}
\end{figure}
We consider the case of a mobile robot navigating in a 2D environment of $20m \times 40m$. Fig.~\ref{fig:collision_2Drobot} shows the underlying PRM graph (sampled nodes in green, connected by edges) with 90 nodes. In this domains, the robot (radius 0.3$m$), starting from its initial belief state (mean pose denoted by S in the figure) has to reach the node $\B{x}_g$ (G in the figure), minimizing its cost function~(\ref{eq:objective_fn}). The blue/\revtext{magenta} triangles denote the landmarks in the environment and the solid black blobs represent the obstacles in the environment. The red ellipses denote the 3$\sigma$ covariances (only the ($x$,$y$) portion is shown). Unless otherwise mentioned, in all the experiments, $0.99-$safe configurations are solicited and the total planning time is the average time for 25 different runs.

The state $\B{x}_k\doteq(x_k, y_k, \theta_k)$ is the robot pose (position and orientation) at time $k$. The applied control vector $\B{u}_k \doteq (\delta_{rot1}, \delta_{trans}, \delta_{rot2})$ consists of an initial rotation $\delta_{rot1}$, followed by a translation of $\delta_{trans}$ and a final rotation of $\delta_{rot2}$, orienting the robot in the required direction. As a result, the following non-linear dynamics is obtained (\cite{thrun2005book})
 \begin{equation}
\begin{split}
x_{k+1} & = x_{k} + \delta_{trans}  \cos(\theta_{k}+ \delta_{rot1})\\
y_{k+1} & = y_{k} + \delta_{trans}  \sin(\theta_{k}+ \delta_{rot1})\\
\theta_{k+1} & = \theta_{k} + \delta_{rot1}+ \delta_{rot2}
\end{split}
\label{odometry_model}
\end{equation}
For robot localization, we consider a landmark based measurement model that returns the range and bearing. The measurement model with noise is thus obtained as
\begin{equation}
\B{z}_k  = \begin{bmatrix} 
r_k^i    =  \sqrt{(O_k^i(1)-x_k(1))^2+O_k^i(2)-x_k(2))^2}\\ \\
\phi_k^i  =  \arctan(\frac{O_k^i(2)-x_k(2)}{O_k^i(1)-x_k(1)})  - x_{k}(3) \  
\end{bmatrix} +  v_k \
,  \ v_k \sim \mathcal{N}(0,Q_k)
\label{sensor_model}
\end{equation}
\noindent where $r_k^i$ and $\phi_k^i$ are the range and bearing of the $i$-th object $O_k^i$ (at time $k$) relative to the robot frame.

The mean landmark locations are $(0,18), (8,28), (20,18), (18.9,13.5)
$. The landmarks at $(20,18), (18.9,13.5)$ are not precisely known and has an associated uncertainty of $diag(0.02,0.02)$ in each of their locations. We first neglect the uncertainty and plan using the mean landmark locations. The planned trajectory is seen in cyan in Fig.~\ref{fig:collision_2Drobot}(a) and the associated beliefs are seen in red. The overall planning time is $0.0041s(\pm0.0003s)$. We note here that the overall planning time also includes the collision probability computation time. Next, we consider the landmark uncertainty during planning. The planned trajectory and the associated beliefs are seen in Fig.~\ref{fig:collision_2Drobot}(b). We note here that there is a significant change in the planned trajectory. The total planning time in this case is $0.0042s(\pm0.0008s)$. Finally, we plan using the true landmark locations of $(0,18), (8,28), (20,17.5), (18.9,13)
$ which are seen in \revtext{magenta} in Fig.~\ref{fig:collision_2Drobot}(c). The overall planning time is $0.0044s(\pm0.0011s)$. As seen in the figure, the planned trajectory is similar to the case when landmark uncertainty is considered. However, executing the plan synthesized by not considering the landmark uncertainty (scenario in Fig.~\ref{fig:collision_2Drobot}(a)) would lead to collision and larger goal state covariance. This is visualized in Fig.~\ref{fig:execution_sim}. The traces of true robot state across ten simulations while executing the plan synthesized by neglecting object uncertainty (scenario in Fig.~\ref{fig:collision_2Drobot}(a)) is shown in Fig.~\ref{fig:execution_sim}(a). The initial state is sampled from the known initial
belief~\revtext{(plotted as red circle in figure)} and 60$\%$ of the executions lead to collision. Fig.~\ref{fig:execution_sim}(b) shows the traces of true robot state across ten simulations while executing the computed control policy by considering object uncertainty (scenario in Fig.~\ref{fig:collision_2Drobot}(b)). Therefore, not considering the object uncertainty lead to localization errors and thereby synthesize inefficient plans.
\subsection{Single-robot Scenarios}
In this Section, we discuss our collision avoidance approach considering single-robot scenarios in the Gazebo simulator. Dynamic obstacles are simulated using different robots whose motion model is unknown to the considered robot. The robot kinematics is as follows
\begin{equation}
\B{x}_{k+1}  = \begin{bmatrix}
 x_k - \frac{V_k}{\omega_k}\sin(\theta_k) + \frac{V_k}{\omega_k}\sin(\theta_k + \omega_k \Delta t)\\
 y_k + \frac{V_k}{\omega_k}\cos(\theta_k) - \frac{V_k}{\omega_k}\cos(\theta_k + \omega_k \Delta t)\\
 \theta_k +  \omega_k \Delta t
\end{bmatrix} + w_k
\label{eq:kinematics}
\end{equation} 
\noindent where the applied control $\B{u}_k = (V_k, \omega_k)^T$ is made up of the linear and angular velocities and $w_k$ is the noise as defined in Section~\ref{sec:intro}. We define the prior uncertainty in the obstacle location as $diag(0.1m, 0.1m)$ and corrupt the range data returned by LDS with varying noise with variance $0.1\times rand(1)m^2$. By default, $0.99-$safe configurations are solicited and we use a look-ahead horizon of $L=7$. Since the TurtleBot3 robot is used, the collision constraint is $\norm{\B{x}_k -\B{s}_k}^2 \leq (0.22+0.22)^2$. In each experiment, we consider a robot starting from the location $(0, 0)$ and having to reach the goal location of $(3, 0)$, subject to minimizing the objective function in (\ref{eq:objective_fn}).

First, we consider three scenarios ($A,B$ and $C$) where the robot has to avoid head-on collisions with dynamic obstacles. The obstacle linear velocities in each scenario are 0.5$m/s$, 1.0$m/s$ and 2.5$m/s$, respectively. This is however unknown to the robot and  at each time step obstacle states are estimated using the approach detailed in Section~\ref{sec:obs}. The robot and obstacle trajectories for the three scenarios are shown in Fig.~\ref{fig:trajectory}(a)-(c). Note that evading a collision is the main focus of these experiments and hence only the relevant trajectories are plotted. Snapshots of four different stages during each the trajectory execution are shown in Fig.~\ref{fig:snapshot}. The first row corresponds to scenario $A$, the second to scenario $B$ and the third row displays snapshots of scenario $C$. For all the scenarios, stage $1$ shows the initial configuration of the robot and the obstacle. Once the obstacle is detected, a control command for evading the obstacle is computed to move towards a $0.99-$safe configuration. The beginning of execution of such a control command is seen in stage $2$. Stage $3$ shows the snapshot when the obstacle and the robot are very close to each other with the robot evading the obstacle to avoid collision. In stage $4$ it is seen that the robot has successfully avoided collisions. The video accompanying this paper demonstrates these results.
\begin{figure}[]
\centering
 \subfloat[Scenario I]{\includegraphics[trim={1cm 1cm 1cm 1cm},clip,width=0.35\textwidth]{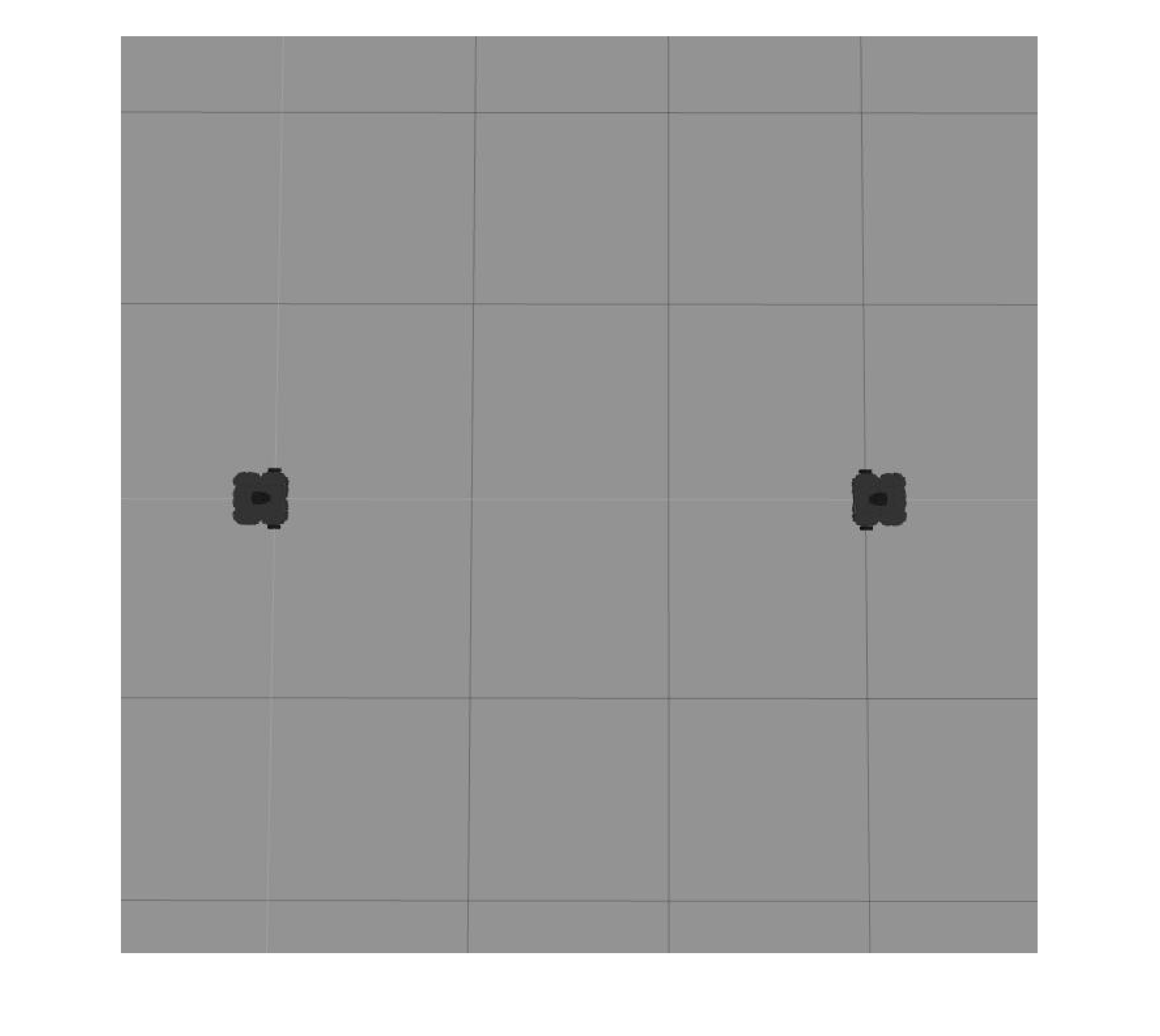}}
  \subfloat[Executed trajectory]{\includegraphics[width=0.4\textwidth]{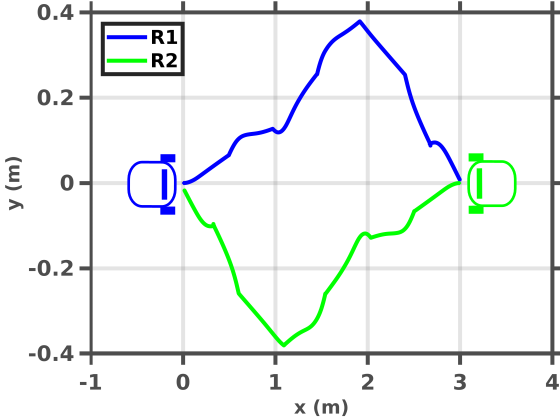}}  
    
  \subfloat[Scenario II]{\includegraphics[trim={1cm 1cm 1cm 1cm},clip,width=0.35\textwidth]{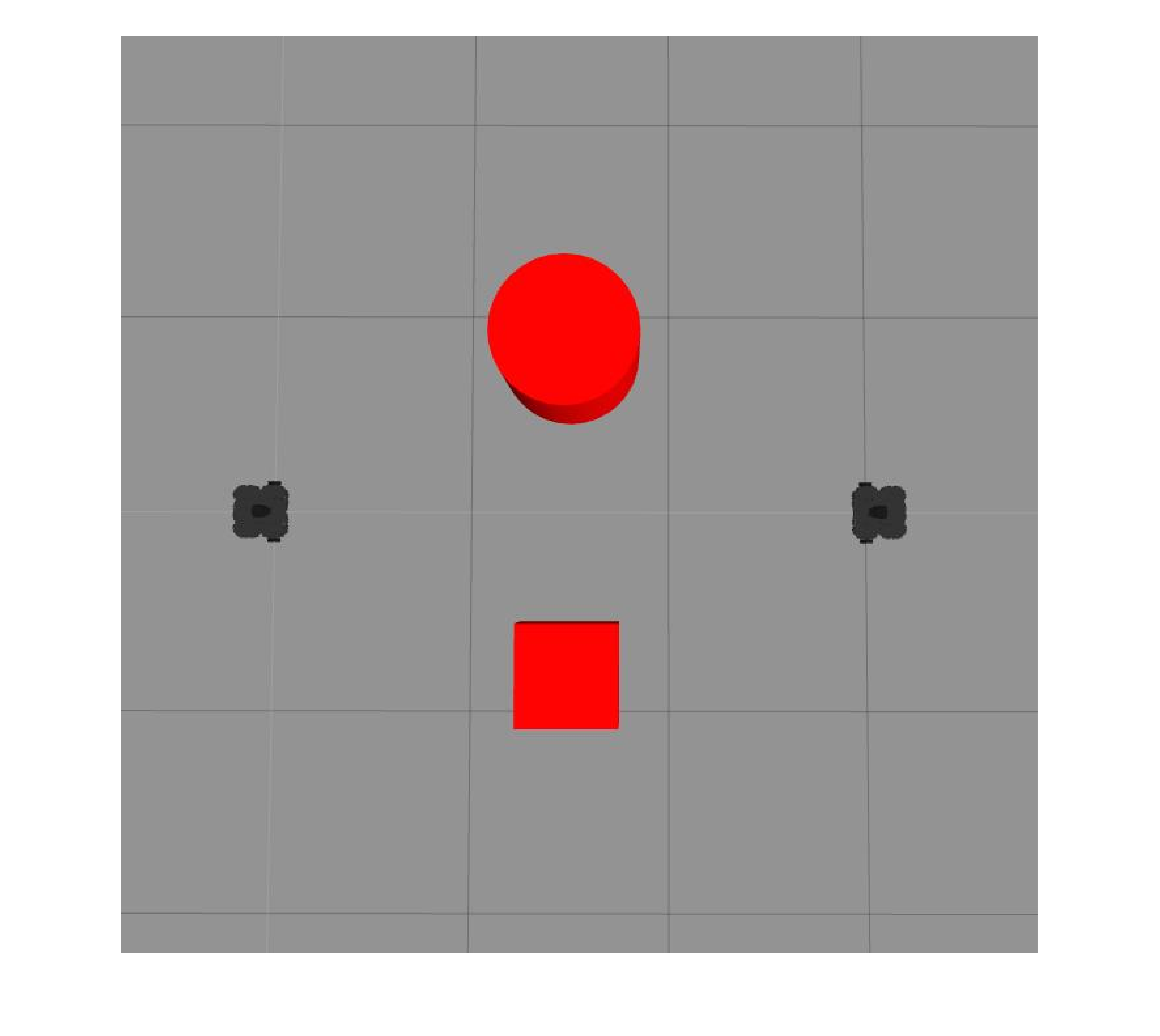}} 
 \subfloat[Executed trajectory]{\includegraphics[width=0.4\textwidth]{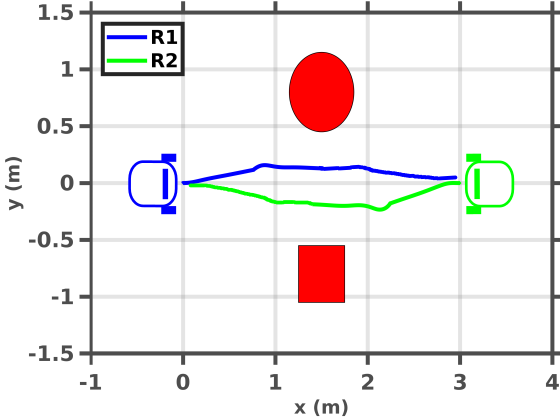}}
 
  \subfloat[Scenario III]{\includegraphics[trim={1cm 1cm 1cm 1cm},clip,width=0.35\textwidth]{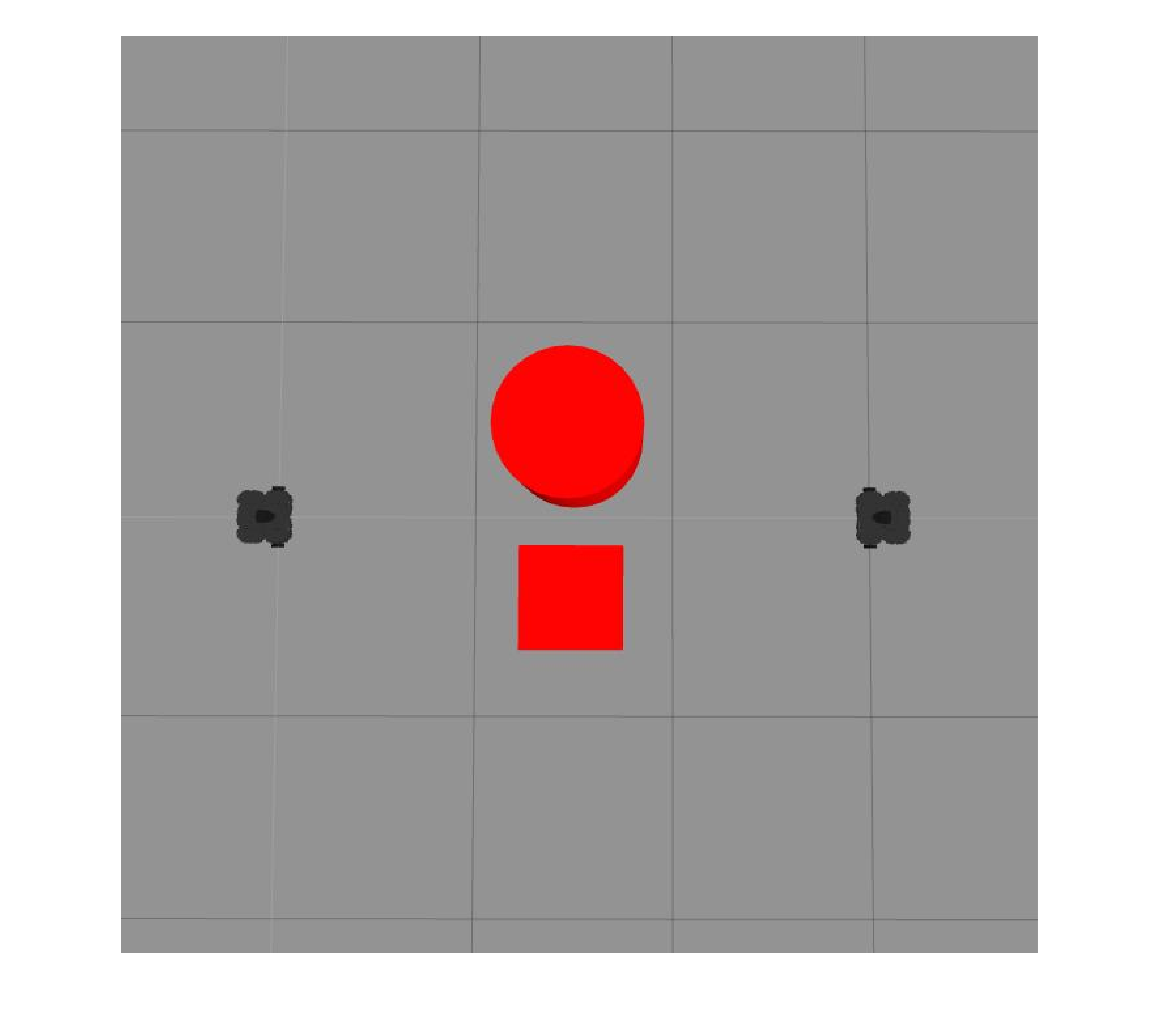}}
  \subfloat[Executed trajectory]{\includegraphics[width=0.4\textwidth]{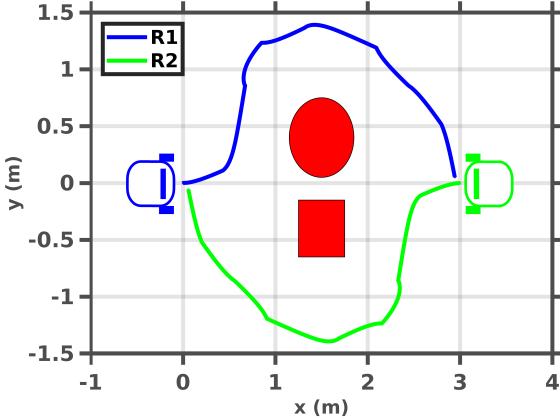}}   
   
  \subfloat[Scenario IV]{\includegraphics[trim={1cm 1cm 1cm 1cm},clip,width=0.35\textwidth]{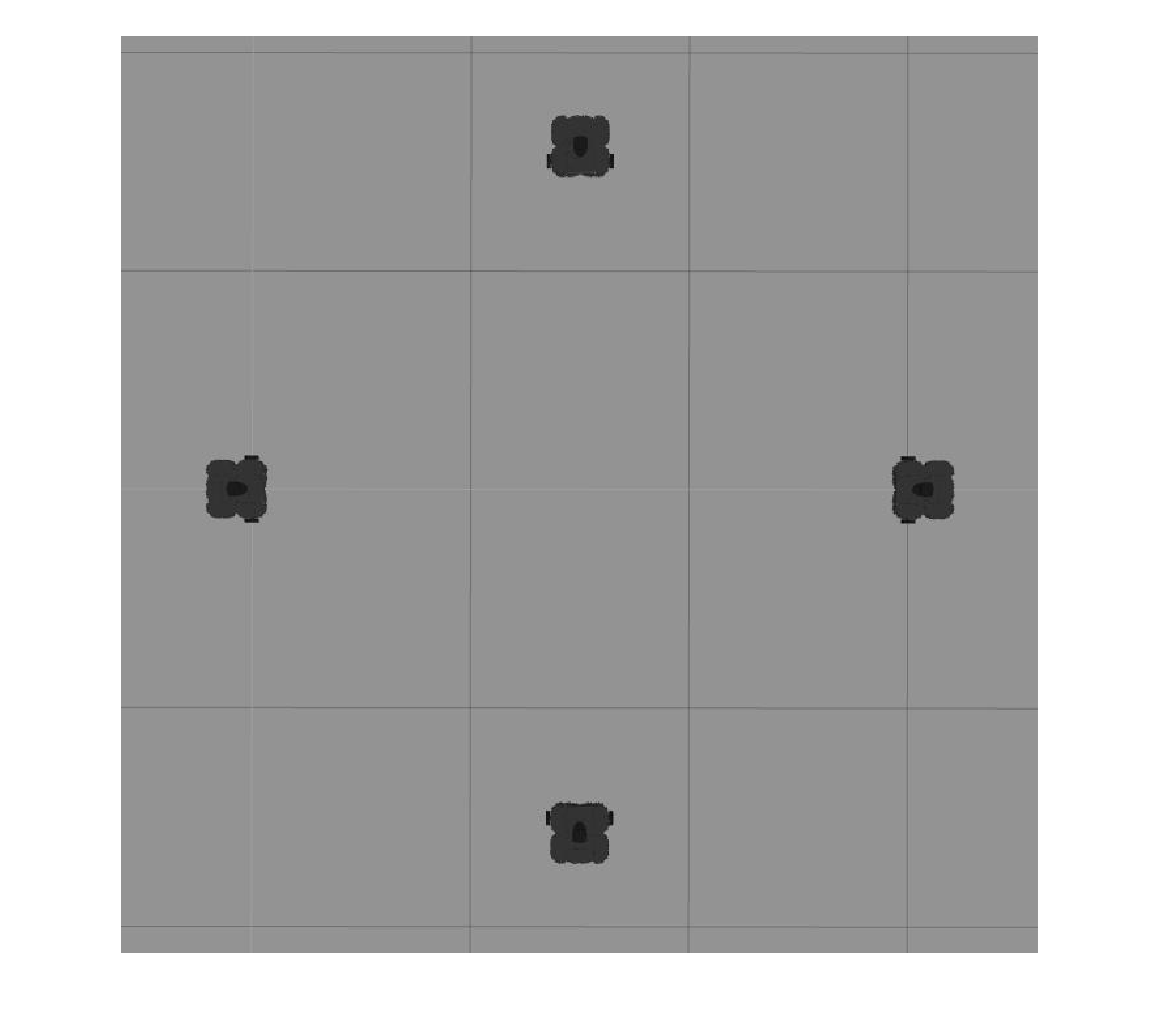}} 
 \subfloat[Executed trajectory]{\includegraphics[width=0.4\textwidth]{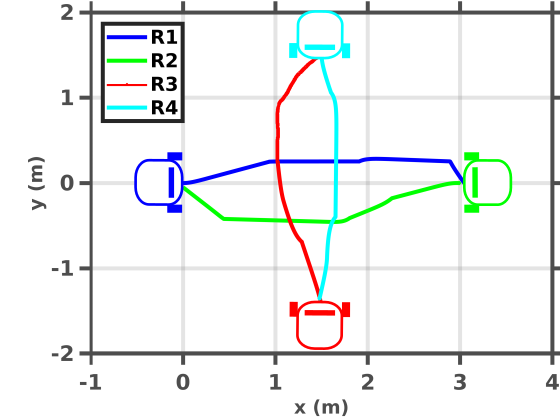}}
        \caption{Different multi-robot scenarios and the corresponding trajectories executed by the robots.}
  \label{fig:multi_scenario}
\end{figure}
For each scenario, the experiment is performed 50 times and the average time for computing the associated collision probability is shown in the first three rows of Table~\ref{table2}. The last row corresponds to Scenario $D$, a multi-obstacle scenario which will be described soon. As a safety metric, the minimum distance between the two robots is also measured and the results are shown in Table~\ref{table2}. For all the scenarios a success rate of 100\% is achieved, that is, in all the 50 experiments, there were no collisions. However, lower look-ahead horizon, that is, $L < 7$ did not give 100\% success rate as most often the obstacles were too close before executing the appropriate control command. Another parameter that affects the success rate is the value of $\epsilon$. For example, a $0.4-$safe configuration always resulted in collision for scenarios $B$ and $C$. 

In scenario $D$, we consider four obstacles, each with different velocities. The robot successfully evades collision with all the four obstacles and the results are shown in the last row of Table~\ref{table2}. The trajectories of the robot and the obstacles can be seen in Fig.~\ref{fig:trajectory}(d). Aerial snapshots at different time instants are shown in Fig.~\ref{fig:snapshot_multi}. Stage $1$ corresponds to the initial configuration of the robot and the obstacles. Stages $2$ and $3$ show the robot moving to evade a head-on collision with  obstacle 2 (obstacle numbers in Fig.~\ref{fig:trajectory}(d)). Stages $4$ through $7$ show different instances while the robot tries to evade the remaining obstacles. Finally, in stage $8$, the robot has successfully avoided potential collisions. The mean computation time for collision probability is 0.3682$s$ and the computation time of the entire framework is 0.4230$s$. The entire framework time includes the time for collision probability computation, uncertainty propagation, and obstacle state estimation.

\begin{table}
\small\sf\centering
 \caption{The minimum distance between the robot and the obstacles and the collision probability computation time for four different scenarios. The minimum distance corresponds to the minimum among all the distances between robots and the obstacles.}
\begin{tabular}{ |c|c|c| } 
 \hline
 Scenario  & Minimum distance (m) & Collision probability  \\
  & & computation time (s)\\
 \hline 
  A & 0.16 & 0.0267 $\pm$ 0.0078 \\ 
  \hline 
 B & 0.31 & 0.0189 $\pm$ 0.0074 \\ 
 \hline
 C & 0.12  & 0.0191 $\pm$ 0.0072 \\
 \hline
 D & 0.20  & 0.0368 $\pm$ 0.0023 \\
 \hline
\end{tabular}
 \label{table2}
\end{table}   

\subsection{Multi-robot Scenarios}
In this Section we demonstrate our approach with multi-robot planning scenarios. In this setting, each robot considers all other robots as dynamic obstacles. However, there is no communication between the robots and the obstacle/robot states are estimated using the approach described in Section~\ref{sec:obs}. 

We first consider different scenarios with two robots. The  initial pose of the robots are $(0,0,0)$ and $(3,0,-\pi)$ and the goal for each robot is to navigate towards the starting location of the other robot. The starting configuration and the executed trajectory of scenario I can be seen in Fig.~\ref{fig:multi_scenario}(a), (b). Scenario II, which includes a cube and a cylinder as static obstacles, is shown in Fig.\ref{fig:multi_scenario}(c), (d). It can be seen that the robots evade collision with each other and the static obstacles and navigate between the obstacles. The locations of the static obstacles are unknown to the robots and they are estimated using the approach discussed in Section~\ref{sec:obs}. However, we assume known data association and we apply the collision constraint derived in Theorem~\ref{theorem2}. The obstacles in scenario II are pulled closer in scenario III (Fig.~\ref{fig:multi_scenario}(e),(f)) to prevent the robots from passing between the obstacles. This is rightly estimated by the robots and they navigate around the obstacles to reach the goal. However it was seen that for $L < 7$, both robots turned to the same side and $20\%$ (10 out of 50) of the time this leads to collision. This is so because, as the robots turn to the side of the cube, the cube occludes one robot from the other. By the time each robot turns around the cube and see the other, they are already too close to avoid collision. In scenario IV (Fig.~\ref{fig:multi_scenario}(g),(h)), we consider four robots, where the robots facing each other are required to swap their positions. The initial poses of each robot are $(0,0,0)$, $(1.5, -1.5,\frac{\pi}{2})$, $(3,0,-\pi)$ and $(1.5, 1.5,-\frac{\pi}{2})$, respectively.

Table~\ref{table3} shows the statistics for the four scenarios discussed above. The minimum distance between the robot and the obstacle and the average computation time for evaluating the collision probability are reported. In scenario IV, it was seen that $\epsilon < 0.99$ leads to collision in $80\%$ of the experiments. For the other scenarios, $\epsilon < 0.9$ successfully evaded collision in all the experiments. 

\begin{table}
\small\sf\centering
\caption{Minimum distance between the robot and the obstacles in four scenarios  and the corresponding collision probability computation time.}
\begin{tabular}{ |c|c|c| } 
 \hline
 Scenario  & Minimum distance (m) & Collision probability  \\
  & & computation time (s)\\
 \hline 
  I & 0.33 & 0.0117 $\pm$ 0.0044 \\ 
  \hline 
 II & 0.08 & 0.0137 $\pm$ 0.0123 \\ 
 \hline
 III & 0.51  & 0.0099 $\pm$ 0.0013 \\
 \hline
 IV & 0.10  & 0.0211 $\pm$ 0.0052 \\
 \hline
\end{tabular}
  \label{table3}
\end{table} 

\section{Discussion}
\label{sec:discussion}
In Section~\ref{sec:comparison}, we have compared our approach to other similar techniques~\cite{lambert2008ICCARV,dutoit2011IEEE,park2018IEEE,zhu2019RAL} and it is seen that our approach outperforms them. In this section we outline few limitations and discuss how to overcome them by proposing suitable extensions. These extensions would enhance the capability and robustness of our approach in challenging scenarios.   

In Section~\ref{sec:obs}, we have modelled the object uncertainty as a Gaussian. The assumption is justified in the case of Gaussian belief states, and works for all practical situations. Yet, in general, the model might not be Gaussian and it has to be determined based on the environment, sensing model and the robot task that needs to be achieved. Finding an appropriate model, especially for non-Gaussian belief states is a work for the future. 

The collision probability approach discussed in this paper is not restrictive to mobile robots and is readily extended to any 3D rigid body robot. For example, a quad-rotor can be approximated using a minimum volume enclosing sphere and therefore our approach can be used directly. Similarly, in the a manipulator robot each link is approximated by minimum volume bounding spheres that tightly enclose the link. For such robots, the collision with an obstacle has to be checked for each bounding volume. For example, let us consider a manipulator robot with $l$ bounding spheres. Then the collision condition for the $i-$th sphere is given by $\mathcal{C}_{\B{x}_k^i,\B{s}_k}$, where $\B{x}_k^i$ is the center of the $i-$th sphere. Furthermore, an alternative and more appropriate  approach is to consider the minimum-volume enclosing ellipsoid for each link~\cite{rimon1997JINT}. For every convex polyhedron, there exists a unique ellipsoid of minimal volume that contains the polyhedron and is called the \textit{L\"{o}wner-John ellipsoid} of the polyhedron~\cite{grotschel1988geometric}. Thus each link can be represented by their corresponding L\"{o}wner-John ellipsoids. However, the collision condition in~(\ref{eq:coll_condition}) is no longer valid. The collision condition should be reformulated using the distance between two ellipsoids. Please note that the representation using L\"{o}wner-John ellipsoid is also extended to the 3D obstacles. 

While formulating the objective function in Section~\ref{sec:obj_function}, we assume that the set of actions from which the robot can plan its future control is known \textit{a priori}. In other words, a finite action set is considered. This justifies the inclusion of the collision cost term $P(\mathcal{C}_{\B{x}_{k+l},\B{s}_{k+l}})$ in~(\ref{eq:objective_fn}). However, our approach is not limited to any specific set of actions or trajectories. The general approach would be to include the set of all possible control actions. The objective function in~(\ref{eq:objective_fn}) is then reformulated as an optimization problem with the collision cost term included as a constraint to keep the collisions within the $1-\epsilon$ bound.

\section{Conclusion}
\label{sec:conclusion}
In this paper, we have presented an approach that incorporates reasoning regarding the landmark uncertainties within the BSP framework.  We consider a Gaussian parametrization of the belief dynamics and derive the corresponding mean and covariance of the belief state when the object uncertainty is considered. We also analyse the effect of adding the object uncertainty for belief estimation and provide the conditions when the effect is negligible. Furthermore, we present a novel approach to compute an exact expression for the collision probability when the robot and obstacle states are uncertain. In contrast, existing works compute an approximation of the actual collision probability. The collision condition is formulated as a quadratic form in random variable and the associated collision probability is the cdf of the quadratic from. We derive the cdf is derived as an infinite series and we prove its convergence and provide an upper bound for the truncation error. We further relax the spherical geometry (of robot and obstacles) assumption by considering the exact convex footprints of the robot and the obstacles and derive the collision constraints for convex polygons. A method to estimate the states of dynamic obstacles and further estimate its future states to enable non-myopic planning is also discussed. Gazebo based simulation using single and multi-robot scenarios with both static and dynamic obstacles demonstrate the real-time online capability of our approach. We further discuss the limitations of our approach and delineate possible directions for future work.

\section*{APPENDIX A: Derivation of~(\ref{eq:object_mu})}
The mean of $b[\B{x}_{k+1}]$ is the value that minimizes $\mathcal{J}_{k+1}$, and therefore it is obtained by equating its first derivative to zero. The first derivative of $\mathcal{J}_{k+1}$ with respect to $\B{x}_{k+1}$ is obtained as
\begin{multline}
\frac{\partial \mathcal{J}_{k+1}}{\partial \B{x}_{k+1}} =  -H_{k+1}^TQ_{k+1}^{-1}\left(\B{z}_{k+1} - h(\bar{\bm{\mu}}_{k+1}) -  H_{k+1}(\B{x}_{k+1}  - \bar{\bm{\mu}}_{k+1})\right) + \\ \Sigma_{O_{k+1}^i}^{-1}\left(\B{x}_{k+1} - \bm{\mu}_{O_{k+1}^i}\right) +  \bar{\Sigma}_{k+1}^{-1} \left(\B{x}_{k+1} - \bar{\bm{\mu}}_{k+1} \right)
\label{eq:first_derivative}
\end{multline}

Since we are evaluating an expression for the mean, we will substitute $\bm{\mu}_{k+1}$ for $\B{x}_{k+1}$. Thus setting the first derivative of $\mathcal{J}_{k+1}$ to zero, we have
\begin{multline}
 H_{k+1}^TQ_{k+1}^{-1}\left(\B{z}_{k+1} - h(\bar{\bm{\mu}}_{k+1})\right)  
 = H_{k+1}^TQ_{k+1}^{-1}H_{k+1}\left(\bm{\mu}_{k+1}  - \bar{\bm{\mu}}_{k+1}\right) + \\ \Sigma_{O_{k+1}^i}^{-1}\left(\bm{\mu}_{k+1} - \bm{\mu}_{O_{k+1}^i}\right) +  \bar{\Sigma}_{k+1}^{-1} \left(\bm{\mu}_{k+1} - \bar{\bm{\mu}}_{k+1} \right)\\
 = H_{k+1}^TQ_{k+1}^{-1}H_{k+1}\left(\bm{\mu}_{k+1}  - \bar{\bm{\mu}}_{k+1}\right) +   \bar{\Sigma}_{k+1}^{-1} \left(\bm{\mu}_{k+1} - \bar{\bm{\mu}}_{k+1} \right) + \\ \Sigma_{O_{k+1}^i}^{-1}\left(\bm{\mu}_{k+1} -\bar{\bm{\mu}}_{k+1} + \bar{\bm{\mu}}_{k+1}- \bm{\mu}_{O_{k+1}^i}\right) \\
  = H_{k+1}^TQ_{k+1}^{-1}H_{k+1}\left(\bm{\mu}_{k+1}  - \bar{\bm{\mu}}_{k+1}\right) +   \bar{\Sigma}_{k+1}^{-1} \left(\bm{\mu}_{k+1} - \bar{\bm{\mu}}_{k+1} \right) + \\ \Sigma_{O_{k+1}^i}^{-1}\left(\bm{\mu}_{k+1} -\bar{\bm{\mu}}_{k+1}\right) +  \Sigma_{O_{k+1}^i}^{-1}\left(\bar{\bm{\mu}}_{k+1}- \bm{\mu}_{O_{k+1}^i}\right)\\
  = \left(H_{k+1}^TQ_{k+1}^{-1}H_{k+1} + \bar{\Sigma}_{k+1}^{-1} + \Sigma_{O_{k+1}^i}^{-1}\right)\left(\bm{\mu}_{k+1}  - \bar{\bm{\mu}}_{k+1}\right) +\\
  +  \Sigma_{O_{k+1}^i}^{-1}\left(\bar{\bm{\mu}}_{k+1}- \bm{\mu}_{O_{k+1}^i}\right)
  \label{eq:mean_derivation}
\end{multline}
\noindent From~(\ref{eq:cov}) we have $\Sigma_{k+1}^{-1} =  H_{k+1}^TQ_{k+1}^{-1}H_{k+1} + \bar{\Sigma}_{k+1}^{-1} + \Sigma_{O_{k+1}^i}^{-1}$. Also using the fact that $K_{k+1}=\Sigma_{k+1} H_{k+1}^TQ_{k+1}^{-1}$,~(\ref{eq:mean_derivation}) simplifies to 
\begin{equation}
K_{k+1}\left(\B{z}_{k+1} - h(\bar{\bm{\mu}}_{k+1})\right)
= \bm{\mu}_{k+1}  - \bar{\bm{\mu}}_{k+1} +  \Sigma_{k+1}\Sigma_{O_{k+1}^i}^{-1}\left(\bar{\bm{\mu}}_{k+1}- \bm{\mu}_{O_{k+1}^i}\right)
\end{equation}
\noindent Rearranging, we get the final expression 
\begin{equation}
\bm{\mu}_{k+1} = \bar{\bm{\mu}}_{k+1} + K_{k+1}\left(\B{z}_{k+1} - h(\bar{\bm{\mu}}_{k+1})\right) +  \Sigma_{k+1}\Sigma_{O_{k+1}^i}^{-1}\left(\bm{\mu}_{O_{k+1}^i} - \bar{\bm{\mu}}_{k+1}\right)
\end{equation}

\section*{APPENDIX B: Derivation of~(\ref{eq:object_sigma})}
In this Appendix we derive the expression for $\Sigma_{k+1}$ in terms of the Kalman gain $K_{k+1}$ and the predicted covariance $\bar{\Sigma}_{k+1}$. For convenience we write down the matrix inversion lemma which states that for any invertible matrices $B$ and $C$ and any matrix $D$ with appropriate dimensions, the following holds true
\begin{equation}
\left(B + DCD^T\right)^{-1} = B^{-1} - B^{-1}D\left(C^{-1} + D^TB^{-1}D\right)^{-1}D^TB^{-1}
\end{equation}

We note here that the Kalman gain $K_{k+1}=\Sigma_{k+1} H_{k+1}^TQ_{k+1}^{-1}$ in~(\ref{eq:object_mu}) is a function of $\Sigma_{k+1}$. Thus we first need to derive an expression for $K_{k+1}$ that does not depend $\Sigma_{k+1}$. To obtain such an expression, we follow the approach for the standard EKF case presented in~\cite{thrun2005book}. We begin by post-multiplying $\Sigma_{k+1} H_{k+1}^TQ_{k+1}^{-1}$ with an identity matrix $I = AA^{-1}$, where
\begin{equation}
A = \left(H_{k+1}\bar{\Sigma}_{k+1}\left(\bar{\Sigma}_{k+1} +\Sigma_{O_{k+1}^i} \right)^{-1}\Sigma_{O_{k+1}^i} H_{k+1}^T + Q_{k+1}\right)
\end{equation}
\noindent To avoid clutter, let us further define $\tilde{\Sigma}_{k+l} = \left(\bar{\Sigma}_{k+1} +\Sigma_{O_{k+1}^i} \right)^{-1}$. The expression for $K_{k+1}$ can then be written as
 \begin{multline}
K_{k+1} = 
\Sigma_{k+1} H_{k+1}^TQ_{k+1}^{-1}\left(H_{k+1}\bar{\Sigma}_{k+1}\tilde{\Sigma}_{k+l}\Sigma_{O_{k+1}^i} H_{k+1}^T + \right. \\ 
\left. Q_{k+1}\right)\left(H_{k+1}\bar{\Sigma}_{k+1}\tilde{\Sigma}_{k+l}\Sigma_{O_{k+1}^i} H_{k+1}^T +Q_{k+1}\right)^{-1}\\
= \Sigma_{k+1} \left(H_{k+1}^TQ_{k+1}^{-1}H_{k+1}\bar{\Sigma}_{k+1}\tilde{\Sigma}_{k+l}\Sigma_{O_{k+1}^i} H_{k+1}^T + \right. \\ 
\left. H_{k+1}^T \right) \left(H_{k+1}\bar{\Sigma}_{k+1}\tilde{\Sigma}_{k+l}\Sigma_{O_{k+1}^i} H_{k+1}^T +Q_{k+1}\right)^{-1}\\
= \Sigma_{k+1} \left(H_{k+1}^TQ_{k+1}^{-1}H_{k+1}\bar{\Sigma}_{k+1}\tilde{\Sigma}_{k+l}\Sigma_{O_{k+1}^i} H_{k+1}^T + \right. \\ 
\left. \left(\bar{\Sigma}_{k+1}\tilde{\Sigma}_{k+l}\Sigma_{O_{k+1}^i}\right)^{-1}\bar{\Sigma}_{k+1}\tilde{\Sigma}_{k+l}\Sigma_{O_{k+1}^i}H_{k+1}^T \right)\\ \left(H_{k+1}\bar{\Sigma}_{k+1}\tilde{\Sigma}_{k+l}\Sigma_{O_{k+1}^i} H_{k+1}^T +Q_{k+1}\right)^{-1}
\label{eq:gain_derivation}
\end{multline}

We will now compute the inverse of the term $\bar{\Sigma}_{k+1}\tilde{\Sigma}_{k+l}\Sigma_{O_{k+1}^i}$. This can be done as follows:
 \begin{multline}
 \left(\bar{\Sigma}_{k+1}\tilde{\Sigma}_{k+l}\Sigma_{O_{k+1}^i}\right)^{-1}
 = \left(\bar{\Sigma}_{k+1}\left(\bar{\Sigma}_{k+1} +\Sigma_{O_{k+1}^i} \right)^{-1}\Sigma_{O_{k+1}^i} \right)^{-1} \\= \Sigma_{O_{k+1}^i}^{-1}\left(\bar{\Sigma}_{k+1} +\Sigma_{O_{k+1}^i} \right)\bar{\Sigma}_{k+1}^{-1} \\
 = \Sigma_{O_{k+1}^i}^{-1} \bar{\Sigma}_{k+1}\bar{\Sigma}_{k+1}^{-1} + \Sigma_{O_{k+1}^i}^{-1}\Sigma_{O_{k+1}^i}\bar{\Sigma}_{k+1}^{-1} 
 = \Sigma_{O_{k+1}^i}^{-1} +\bar{\Sigma}_{k+1}^{-1}
\label{eq:inverse}
\end{multline}
\noindent The expression in (\ref{eq:gain_derivation}) simplifies to 
 \begin{multline}
K_{k+1} = 
\Sigma_{k+1} \left(H_{k+1}^TQ_{k+1}^{-1}H_{k+1}\bar{\Sigma}_{k+1}\tilde{\Sigma}_{k+l}\Sigma_{O_{k+1}^i} H_{k+1}^T + \right. \\ 
\left. \left(\Sigma_{O_{k+1}^i}^{-1} +\bar{\Sigma}_{k+1}^{-1}\right)\bar{\Sigma}_{k+1}\tilde{\Sigma}_{k+l}\Sigma_{O_{k+1}^i}H_{k+1}^T \right)\\ \left(H_{k+1}\bar{\Sigma}_{k+1}\tilde{\Sigma}_{k+l}\Sigma_{O_{k+1}^i} H_{k+1}^T +Q_{k+1}\right)^{-1} \\
= \Sigma_{k+1} \left(H_{k+1}^TQ_{k+1}^{-1}H_{k+1} + \Sigma_{O_{k+1}^i}^{-1} +\bar{\Sigma}_{k+1}^{-1} \right) \\ \bar{\Sigma}_{k+1}\tilde{\Sigma}_{k+l}\Sigma_{O_{k+1}^i}H_{k+1}^T \\ \left(H_{k+1}\bar{\Sigma}_{k+1}\tilde{\Sigma}_{k+l}\Sigma_{O_{k+1}^i} H_{k+1}^T +Q_{k+1}\right)^{-1} \\
=\Sigma_{k+1}\left(\Sigma_{k+1}\right)^{-1}\bar{\Sigma}_{k+1}\tilde{\Sigma}_{k+l}\Sigma_{O_{k+1}^i}H_{k+1}^T \\ \left(H_{k+1}\bar{\Sigma}_{k+1}\tilde{\Sigma}_{k+l}\Sigma_{O_{k+1}^i} H_{k+1}^T +Q_{k+1}\right)^{-1}
\label{eq:gain_derivation1}
\end{multline}
\noindent where we have used the fact that $\Sigma_{k+1}^{-1} = H_{k+1}^TQ_{k+1}^{-1}H_{k+1} + \Sigma_{O_{k+1}^i}^{-1} +\bar{\Sigma}_{k+1}^{-1}$. Thus we obtain
\begin{equation}
K_{k+1}  =\bar{\Sigma}_{k+1}\tilde{\Sigma}_{k+l}\Sigma_{O_{k+1}^i}H_{k+1}^T  \left(H_{k+1}\bar{\Sigma}_{k+1}\tilde{\Sigma}_{k+l}\Sigma_{O_{k+1}^i} H_{k+1}^T +Q_{k+1}\right)^{-1}
 \label{eq:kalman_gain}
\end{equation}

Let us now define $\Xi_{k+1} = \Sigma_{O_{k+1}^i}^{-1} +\bar{\Sigma}_{k+1}^{-1}$. Applying the matrix inversion lemma to the right hand side of~(\ref{eq:cov}), we have
\begin{equation}
\Sigma_{k+1} = \Xi_{k+1} ^{-1}  - \Xi_{k+1}^{-1}H_{k+1}^T\left(Q_{k+1} + H_{k+1}\Xi_{k+1}^{-1}H_{k+1}^T\right)^{-1}H_{k+1}\Xi_{k+1} ^{-1}
\label{eq:sigma_inversion}
\end{equation}
\noindent From~(\ref{eq:inverse}), we have 
\begin{equation}
\Xi_{k+1}^{-1} = \bar{\Sigma}_{k+1}\tilde{\Sigma}_{k+l}\Sigma_{O_{k+1}^i}
\label{eq:s_inverse}
\end{equation}
\noindent We note here that the expression $\Xi_{k+1}^{-1}$ appears four times in~(\ref{eq:sigma_inversion}). Substituting for $\Xi_{k+1}^{-1}$ using~(\ref{eq:s_inverse}) in the second and third expression of $\Xi_{k+1}^{-1}$ in~(\ref{eq:sigma_inversion}), we get
\begin{multline}
\Sigma_{k+1} = \Xi_{k+1}^{-1}  - \bar{\Sigma}_{k+1}\tilde{\Sigma}_{k+l}\Sigma_{O_{k+1}^i}
H_{k+1}^T\\
\left(Q_{k+1} + H_{k+1}\bar{\Sigma}_{k+1}\tilde{\Sigma}_{k+l}\Sigma_{O_{k+1}^i}H_{k+1}^T\right)^{-1}
H_{k+1}\Xi_{k+1}^{-1}
\label{eq:complete_sigma}
\end{multline}
\noindent From~(\ref{eq:kalman_gain}) and~(\ref{eq:s_inverse}), it is easily seen that the expression in~(\ref{eq:complete_sigma}) simplifies to
\begin{multline}
\Sigma_{k+1} = \Xi_{k+1}^{-1} - K_{k+1}H_{k+1}\Xi_{k+1} ^{-1}
= \left(I -  K_{k+1}H_{k+1} \right)\Xi_{k+1}^{-1} \\
= \left(I - K_{k+1}H_{k+1}\right)\bar{\Sigma}_{k+1}\tilde{\Sigma}_{k+l}\Sigma_{O_{k+1}^i} \\
= \left(I - K_{k+1}H_{k+1}\right)\bar{\Sigma}_{k+1}\left(\bar{\Sigma}_{k+1} +\Sigma_{O_{k+1}^i} \right)^{-1}\Sigma_{O_{k+1}^i}
\end{multline}

This completes the derivation. 

\section*{APPENDIX C: Derivation of~(\ref{eq:c_k})}
From~(\ref{eq:ln_M}), we have
\begin{equation}
\ln M(\theta) = d_0 + \sum_{k=1}^{\infty} d_k \frac{\theta^k}{k}
\label{eq:appendix31}
\end{equation}
For differentiable $M(\theta)$, we have
\begin{equation}
\frac{d}{d \theta} \ln M(\theta) = \frac{1}{M(\theta)}\frac{d}{d\theta} M(\theta) = \sum_{k=1}^{\infty} c_k \theta^{k-1}
\label{eq:appendix32}
\end{equation}
\noindent where we have used the definition of $M(\theta)$ given in~(\ref{eq:M_theta_def}). Also note that by construction $M(\theta) > 0$. Re-arranging~(\ref{eq:appendix32}), we obtain
\begin{equation}
M(\theta)\frac{d}{d \theta} \ln M(\theta) = \sum_{k=1}^{\infty} c_k \theta^{k-1}
\end{equation}
\noindent From~(\ref{eq:appendix31}), we have
\begin{equation}
\frac{d}{d \theta} \ln M(\theta) = \sum_{k=1}^{\infty} d_k \theta^{k-1}
\label{eq:appendix33}
\end{equation}
\noindent From~(\ref{eq:appendix32}) and~(\ref{eq:appendix33}), we thus obtain
\begin{equation}
\left(\sum_{k=0}^{\infty} c_k \theta^{k} \right) \left(  \sum_{k=1}^{\infty} d_k \theta^{k-1}\right) =  \sum_{k=1}^{\infty} c_k \theta^{k-1}
\end{equation}
Comparing the coefficient of $\theta^{k-1}$ on both sides of the equation, we get the required expression for $c_k$ as
\begin{equation}
c_k = \frac{1}{k}\sum\limits_{j=0}^{k-1}d_{k-j}c_j 
\end{equation}

\section*{References}
\bibliographystyle{elsarticle-num}
\bibliography{/home/antony/research_genoa/References/References}

\end{document}